\documentclass{article}

\usepackage{arxiv}

\usepackage{amsmath}
\usepackage[utf8]{inputenc} 
\usepackage[T1]{fontenc}    
\usepackage{hyperref}       
\usepackage{url}            
\usepackage{booktabs}       
\usepackage{array}          
\usepackage{amsfonts}       
\usepackage{nicefrac}       
\usepackage{microtype}      
\usepackage{cleveref}       
\usepackage{graphicx}
\usepackage{amsmath}
\usepackage{natbib}
\usepackage{doi}

\usepackage{xcolor}

\title{KernelOnet: An Interpretable Neural Operator Based on Kernel Functions}

\usepackage{authblk}

\author[1]{Yuan Guo}
\author[1]{Hanshu Chen}
\author[1]{Qiang Xi}
\author[3]{Timon Rabczuk}
\author[1,2]{Zhuojia Fu\thanks{Corresponding author. \texttt{paul212063@hhu.edu.cn}}}
\affil[1]{College of Mechanics and Engineering Science, Hohai University, Nanjing 211100, China}
\affil[2]{Key Laboratory of Ministry of Education for Coastal Disaster and Protection, Hohai University, Nanjing 210098, China}
\affil[3]{Institute of Structural Mechanics, Bauhaus-University Weimar, Weimar 99423, Germany}

\renewcommand{\shorttitle}{KernelOnet}

\hypersetup{
pdftitle={KernelOnet: An Interpretable Neural Operator Based on Kernel Functions},
pdfsubject={cs.LG, physics.comp-ph},
pdfauthor={Yuan Guo, Hanshu Chen, Qiang Xi, Timon Rabczuk, Zhuojia Fu},
pdfkeywords={Neural Operator, Kernel Function, DeepONet, fundamental solution},
}

\begin{document}
\maketitle

\begin{abstract}
This paper proposes a new interpretable neural operator framework, termed the Kernel Operator Network (KernelOnet),
which explicitly incorporates kernel functions into the neural operator architecture so that the operator structure is consistent with the kernel-expansion form used in boundary-type kernel-expansion methods.
Unlike traditional neural operators such as DeepONet, which rely on deep networks to learn basis functions implicitly,
KernelOnet replaces the trunk network with explicit kernel functions and provides three complementary kernel construction strategies:
the first is a data-driven learnable kernel, in which a radial basis function is parameterized by a neural network and learned directly from data; for constant-coefficient linear problems, the learned kernel can be regarded as a non-singular fundamental solution;
the second is a physics-informed kernel, which explicitly embeds physical information such as analytic fundamental solutions into the network structure, so that the expansion automatically satisfies the governing equation and can be trained without supervision using boundary conditions alone, without any interior solution data;
the third is a hybrid kernel, which splits the solution according to the linear principal part of the governing equation into a homogeneous part spanned by analytic fundamental solutions and a source part carried by low-rank learned correction kernels, thereby balancing physical priors against data-fitting capability on nonlinear problems for which no analytic fundamental solution exists.
On three benchmark problems and one engineering problem in a shallow-water waveguide, KernelOnet attains high-accuracy solutions; on the examples that can be compared directly with DeepONet it achieves higher accuracy with fewer learnable parameters. Its unsupervised configuration can be trained without any interior solution labels, its per-query inference cost is far below that of per-instance solvers, and it offers an effective route to acoustic propagation in unbounded exterior domains, which general-purpose neural operators struggle to handle.
\end{abstract}


\section{Introduction}

Partial differential equations (PDEs) are ubiquitous in many fields such as physics, engineering and materials science, and their efficient numerical solution has long been one of the core problems of computational mathematics.
Traditional numerical methods, such as finite difference methods \cite{smith1985finitepdf}, finite element methods \cite{reddy1984finiteelement} and various meshless methods, generally require discretizing the solution domain and approximating the solution by solving large-scale linear or nonlinear algebraic systems,
and their computational cost grows sharply with the problem size and dimension; moreover, for tasks with a fixed geometry the system must still be reassembled and solved for every new set of boundary conditions or source terms, which makes it difficult to meet the efficiency requirements of emerging applications such as real-time prediction, inverse problems and uncertainty quantification.

In recent years, the rapid development of deep learning \cite{lecun2015deep} has brought a new paradigm to scientific computing, and neural operators, as a new class of learning frameworks, have attracted widespread attention.
Their goal is to learn directly from data the mapping between function spaces, that is, the operator itself, so that given a new input function such as a boundary condition, a source term or an initial condition, the solution function can be produced directly by a single forward pass without solving the PDE again.
Within the family of neural operators, DeepONet \cite{lu2021learning} constructs the solution operator on the basis of a branch--trunk architecture together with the universal approximation theorem for operators \cite{chen1995universal}, and subsequent work has further introduced physical information to improve accuracy and generalization \cite{wang2021learning}; this physics-informed line of thought has since developed further: on the one hand, the weak form of the PDE is incorporated into operator training in a variational manner, giving rise to the variational physics-informed neural operator (VINO) \cite{eshaghi2025vino}; on the other hand, the learned operator is used for the pretraining and warm-starting of iterative solvers such as finite element methods, so as to accelerate conventional numerical solution procedures \cite{wang2026pretrainfem};
the Fourier neural operator (FNO) \cite{li2021fourier} and its generalizations perform global convolution in the spectral domain and have produced variants such as multipole graph structures \cite{li2020multipole}, learnable deformations on general geometries \cite{li2020learned} and large-scale weather forecasting \cite{pathak2022fourcastnet};
operator learning based on Transformers \cite{cao2021choose,hao2023gnot} and network layers based on Clifford algebras \cite{brandstetter2022clifford} have also appeared;
in addition, tensor-product methods for operators with multiple inputs \cite{jin2022mionet}, multi-head neural operators for multifield and interface-dynamics modelling \cite{eshaghi2026multihead}, Green's function learning for nonlinear boundary value problems \cite{gin2021deepgreen}, and physics-informed neural networks (PINNs) that embed PDE information into the loss function \cite{raissi2019physics}
together with their follow-up work \cite{karniadakis2021physics,cuomo2022scientific,lu2021deepxde} jointly form a flourishing landscape of operator learning and physics-informed machine learning, for which a review can be found in \cite{kovachki2023neural}.
However, most of the above methods rely on deep networks to learn basis functions or operator kernels implicitly, and their structure lacks an explicit connection with the specific governing equation, which raises two problems: first, poor interpretability, since it is difficult to understand how the model works from its internal structure;
second, insufficient physical consistency, since a large amount of high-fidelity training data is usually required, and generalization is limited in the small-sample regime.

In contrast, kernel methods and numerical methods based on kernel expansions have constructed kernel functions explicitly from the very beginning.
In machine learning, the theory of reproducing kernel Hilbert spaces (RKHS) \cite{aronszajn1950theory} provides a rigorous mathematical foundation for kernel functions, and the success of support vector machines \cite{cortes1995support},
Gaussian process regression \cite{williams1996gaussian,rasmussen2006gaussian} and general kernel methods \cite{hofmann2008kernel,scholkopf2002learning} shows that embedding data into a suitable function space through a kernel function yields good generalization performance;
radial basis function (RBF) networks \cite{park1991universal} and regularization network theory \cite{girosi1995regularization} have further revealed the intrinsic connection between kernel methods and neural networks.
In numerical methods, radial basis function interpolation theory \cite{micchelli1986interpolation} provides the convergence foundation for meshless and boundary-type kernel methods; the method of fundamental solutions (MFS) \cite{golberg1995poisson,chen1998numerical,fairweather1998method,fairweather2003scattering,golberg1997method},
the Trefftz method \cite{kita1995trefftz}, the boundary element method (BEM) \cite{brebbia1984boundary} and various meshless methods \cite{belytschko1996meshless,liu2005meshfree,fasshauer2007meshfree,monaghan2005smooth} all employ kernels that contain physical information, such as fundamental solutions, as basis functions in order to achieve high accuracy and high computational efficiency.
Schaback and Wendland \cite{schaback2006kernel} pointed out that the kernel techniques used in machine learning and the kernel functions used in meshless and boundary-type kernel methods are in fact the same mathematical object manifested in two fields.
However, traditional kernel methods and kernel-expansion-based numerical methods usually rely on a manually chosen kernel function and discretization scheme, such as the source-point layout of a boundary collocation method or the choice of the virtual boundary for a boundary integral, and are difficult to adjust automatically for complex nonlinear problems;
and mainstream neural operators learn basis functions implicitly, sacrificing physical consistency and interpretability.
This observation motivates us to combine the two and to propose an operator learning framework that uses explicit kernel functions as the basis functions of a neural network.

To address the above issues, this paper proposes the Kernel Operator Network (KernelOnet), a direct improvement over DeepONet: it replaces the trunk network with explicit kernel functions, so that the operator output structure is consistent with the kernel-expansion form used in boundary-type numerical methods based on kernel expansions, such as the boundary-type collocation method of fundamental solutions (MFS) and the boundary integral boundary element method (BEM), while remaining amenable to analysis within the well-developed theory of kernel methods, such as radial basis function interpolation theory and reproducing kernel Hilbert space theory.
Specifically, this paper provides three complementary kernel construction strategies: (i)~a data-driven learnable kernel (KernelOnet-RBF), whose kernel is a neural-network-parameterized radial basis function learned directly from data by a shallow network, so that for constant-coefficient linear problems the learned kernel can be regarded as a non-singular fundamental solution;
(ii)~a physics-informed kernel (KernelOnet-PIKF), whose kernel is the analytic fundamental solution of the governing equation, so that the expansion automatically satisfies the governing equation and can therefore be trained without supervision using boundary conditions alone, without any interior solution data;
(iii)~a hybrid kernel (KernelOnet-HK), which splits the solution according to the linear principal part of the governing equation into a homogeneous part spanned by analytic fundamental solutions and a source part spanned by $K_c$ low-rank learned correction kernels; the former satisfies the linear principal part exactly while the latter carries the nonlinear source terms that the kernel functions cannot represent, so that when the problem admits no complete fundamental solution, such as a nonlinear equation with nonlinear source terms, a compromise between physical prior and breadth of applicability is still achievable with the help of supervised data.
Numerical experiments on four examples---the Laplace equation on a circular domain, the nonlinear modified Helmholtz equation on a star-shaped domain, the complex Helmholtz equation in an unbounded exterior domain, and the underwater acoustic radiation and propagation induced by spherical-shell vibration in a shallow-water waveguide---show that the proposed KernelOnet achieves high prediction accuracy while significantly improving interpretability, that the unsupervised configuration requires no interior solution labels at all, and that the three variants exhibit complementary advantages on different problems. It should be noted that both the physics-informed kernel and the hybrid kernel presuppose the existence of an analytic fundamental solution of the governing equation or of its linear principal part, which is the condition under which the framework takes effect.

The main contributions of this paper are summarized as follows:
(1)~the KernelOnet framework is proposed, which embeds kernel functions explicitly into a neural operator to replace the implicit trunk network of DeepONet, establishing an intrinsic connection between neural operators and boundary-type numerical methods based on kernel expansions;
(2)~three complementary kernel construction strategies are given, namely the data-driven learnable kernel, the physics-informed kernel and the hybrid kernel, and their physical meaning and parameter efficiency are analysed;
(3)~on three benchmark examples, the advantages of KernelOnet in accuracy, interpretability and unsupervised training are systematically verified, and in particular an effective solution route is provided for unbounded-domain problems that general-purpose neural operators struggle to solve;
(4)~with a view to engineering applications, KernelOnet is applied to the underwater acoustic radiation and propagation problem induced by spherical-shell vibration in a shallow-water waveguide, where the waveguide Pekeris kernel and the normal-mode kernel are used to construct the kernel functions, so that the near field and the far field are both solved without supervision and good robustness is exhibited under various sound speed profiles.

The remainder of this paper is organized as follows. Section~2 introduces the methodology and theoretical derivation of KernelOnet in detail: it first gives the problem definition of operator learning and its differences from conventional neural networks, then presents the two baseline architectures DeepONet and PI-DeepONet together with their loss constructions, and afterwards introduces kernel functions and kernel-expansion-based boundary-type solution methods as the theoretical bridge connecting neural operators with PDE solution methods; on this basis, three complementary kernel construction strategies are proposed and compared, and finally the convergence and accuracy of each variant are analysed from the perspective of kernel theory. Section~3 systematically verifies the proposed method on three benchmark examples and one practical engineering example. Section~4 concludes the paper and outlines future work.

\section{Methodology}

\subsection{Operator Learning}

Let $\mathcal{A}$ and $\mathcal{U}$ denote the input function space and the output solution function space, respectively. The goal of operator learning is to approximate the solution operator determined by a PDE,
\begin{equation}
\mathcal{G}:\mathcal{A}\rightarrow\mathcal{U},\qquad a\mapsto u,
\end{equation}
where the input function $a$ (for example a boundary condition, a source term or an initial condition) and the output solution $u$ satisfy the governing equation $\mathcal{L}u=f$ together with the corresponding well-posedness conditions (boundary and initial conditions). Unlike the traditional "solve problem by problem" paradigm, operator learning simultaneously exploits multiple input--output function pairs
\begin{equation}
\mathcal{D}=\left\{\left(a^{(i)},u^{(i)}\right)\right\}_{i=1}^{N}
\end{equation}
to learn a parametric approximation $\mathcal{G}_\theta\approx\mathcal{G}$ of the operator $\mathcal{G}$, so that during inference it directly yields the solution $u=\mathcal{G}_\theta(a)$ for an unseen input function $a$ without solving the PDE again. Representative operator learning methods include DeepONet \cite{lu2021learning} and the Fourier neural operator (FNO) \cite{li2021fourier}, among others; a review can be found in \cite{kovachki2023neural}.
Compared with traditional neural network methods, operator learning differs in two essential respects.
    First, conventional neural networks learn a mapping between fixed-dimensional vectors, and both the input and output dimensions are fixed at training time, so that changing the grid or the resolution requires retraining; in operator learning, the inputs and outputs are functions, and the discrete values at a finite number of nodes are merely sampled realizations of those functions, so that in principle the dependence on a fixed discrete layout can be removed. Whether genuine resolution independence is achieved, however, depends on the specific network architecture: the Fourier neural operator \cite{li2021fourier} and related methods parameterize functions in the frequency domain and can be evaluated directly on inputs and outputs of different resolutions, and therefore possess resolution invariance; the trunk network of DeepONet can likewise be evaluated at arbitrary output coordinates, but its branch network takes its input from a fixed set of sensors, so that changing the input sampling layout requires interpolation or readjustment of the model. To address this limitation of the branch network, the resolution-independent neural operator proposed by Bahmani et al. \cite{bahmani2025resolution} learns a dictionary of continuous basis functions by means of an implicit neural representation and projects input functions with differing sensor layouts onto a fixed-dimensional embedding space, thereby obtaining input-side resolution independence without modifying the DeepONet architecture.
Second, operator learning provides a theoretical guarantee at the level of universal approximation: under certain conditions, operators constructed by neural networks can approximate nonlinear continuous operators to arbitrary accuracy \cite{chen1995universal}, which justifies the feasibility of approximating solution operators with a finite number of parameters.

\subsection{DeepONet and PI-DeepONet}

DeepONet builds on the classical results of Chen and Chen \cite{chen1995universal}, Hornik et al. \cite{hornik1989multilayer} and Cybenko \cite{cybenko1989approximation} on the universal approximation of neural networks: for any continuous operator $\mathcal{G}$ and any compact set $K$, there exists a finite-term expansion of the form
\begin{equation}
\mathcal{G}(a)(\mathbf{x})\approx\sum_{k=1}^{p}\beta_k(a)\,\tau_k(\mathbf{x})
\end{equation}
that approximates $\mathcal{G}(a)(\mathbf{x})$ uniformly on $K$, where the coefficients $\beta_k(a)$ depend only on the input function $a$ and the basis functions $\tau_k(\mathbf{x})$ depend only on the spatial coordinate $\mathbf{x}$.

\begin{figure}[htbp]
\centering
\includegraphics[width=0.95\textwidth]{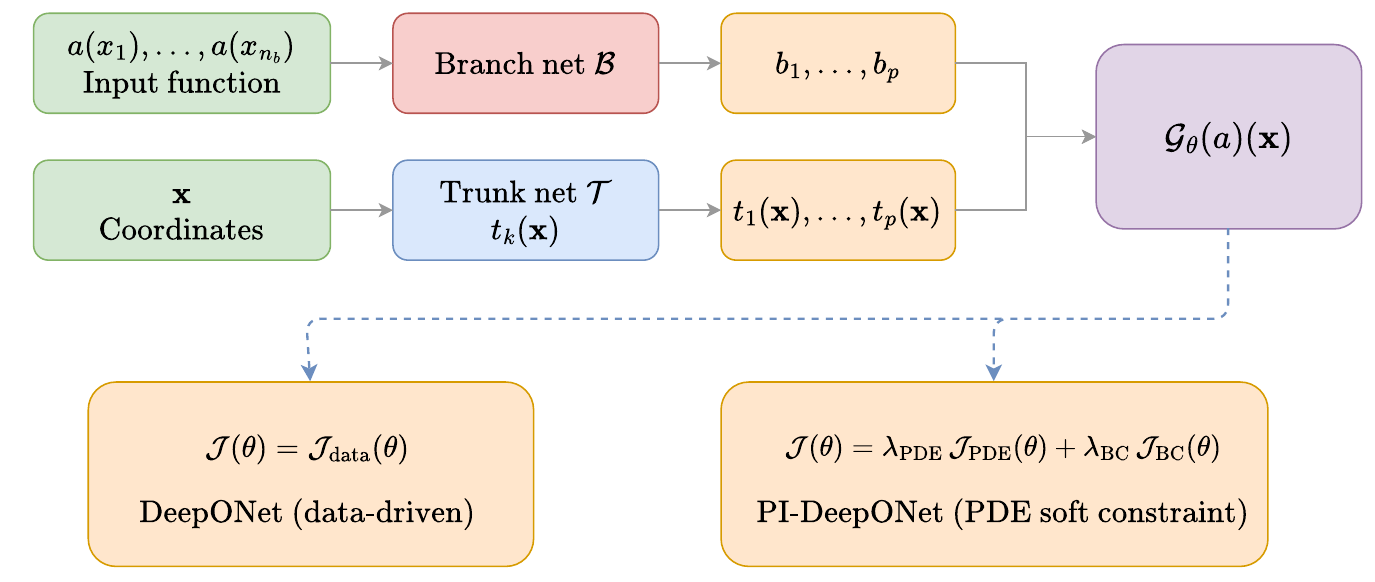}
\caption{Architectures of DeepONet and PI-DeepONet. DeepONet learns the basis functions $t_k$ implicitly through a fully connected trunk network and optimizes the data loss in a supervised manner; on this basis, PI-DeepONet adds the PDE residual and the boundary condition residual to the loss function as soft constraints, so that in the unsupervised case the loss contains only these two terms, balanced by the weights $\lambda_{\mathrm{PDE}}$ and $\lambda_{\mathrm{BC}}$, while the data loss term can be added when supervised data are available to further improve performance.}
\label{fig:deeponet}
\end{figure}

DeepONet uses two neural networks to approximate the coefficients and the basis functions of this expansion separately: the branch network $\mathcal{B}:\mathbb{R}^{n_b}\rightarrow\mathbb{R}^{p}$ maps the discrete values of the input function at a set of sensors to the coefficients $\beta_k$, and the trunk network $\mathcal{T}:\mathbb{R}^{d}\rightarrow\mathbb{R}^{p}$ maps the spatial coordinates to the basis functions $\tau_k$; the final output is the inner product of the two plus a bias:
\begin{equation}
\mathcal{G}_\theta(a)(\mathbf{x})=\sum_{k=1}^{p}b_k(a)\,t_k(\mathbf{x})+b_0,
\end{equation}
where $b_0$ is a learnable bias term introduced in this paper and $p$ is the number of basis functions (that is, the output dimension of the trunk network).
    In practice the trunk output is usually passed through a $\tanh$ activation to ensure boundedness.
    Although DeepONet possesses universal approximation capability, in the fully connected trunk network adopted in this paper the basis functions $t_k(\mathbf{x})$ are learned entirely implicitly from data, so that neither the physical structure of the governing equation is explicitly exploited nor is a direct physical interpretation available; this is the starting point of the improvement proposed in this paper.
    The architecture of DeepONet is shown in Figure~\ref{fig:deeponet}.

DeepONet can be trained in a supervised manner; in this paper, the discrete values of the input function at a finite number of boundary points serve as the input of the branch network, and the discrete values of the output solution at interior evaluation points serve as the supervision signal. Let the training data be
\begin{equation}
\mathcal{D}=\left\{\left(\mathbf{y}_b^{(i)},\mathbf{u}^{(i)}\right)\right\}_{i=1}^{N},\qquad \mathbf{y}_b^{(i)}\in\mathbb{R}^{n_b},\quad \mathbf{u}^{(i)}\in\mathbb{R}^{n_t},
\end{equation}
where $n_b$ is the number of boundary points and $n_t$ is the number of interior evaluation points. The model parameters $\theta$ are obtained by minimizing the data loss, that is, the pointwise squared error between the output and the reference solution over all interior evaluation points:
\begin{equation}
\mathcal{J}(\theta)=\mathcal{J}_{\mathrm{data}}(\theta)=\frac{1}{N\,n_t}\sum_{i=1}^{N}\sum_{k=1}^{n_t}\left(\mathcal{G}_\theta\left(\mathbf{y}_b^{(i)}\right)\left(\mathbf{x}^{(k)}\right)-u_k^{(i)}\right)^{2},
\end{equation}
where $\mathbf{x}^{(k)}$ is the $k$-th interior evaluation point and $\mathcal{G}_\theta\left(\mathbf{y}_b^{(i)}\right)\left(\mathbf{x}^{(k)}\right)$ is the value of the operator output at that point.

To introduce physical information into operator learning, the physics-informed DeepONet (PI-DeepONet) \cite{wang2021learning} was proposed on the basis of DeepONet: instead of relying solely on the purely data-driven loss above, it adds the residual of the governing equation to the loss function as a soft constraint. Specifically, in the unsupervised setting where no interior solution labels are used, its loss function consists of two weighted terms:
\begin{equation}
\mathcal{J}(\theta)=\lambda_{\mathrm{PDE}}\,\mathcal{J}_{\mathrm{PDE}}(\theta)+\lambda_{\mathrm{BC}}\,\mathcal{J}_{\mathrm{BC}}(\theta),
\end{equation}
where $\mathcal{J}_{\mathrm{PDE}}$ is the PDE residual loss evaluated at interior collocation points, $\mathcal{J}_{\mathrm{BC}}$ is the boundary condition residual loss, and $\lambda_{\mathrm{PDE}}$ and $\lambda_{\mathrm{BC}}$ are the corresponding weight coefficients.
    When supervised data are available, a data loss term $\mathcal{J}_{\mathrm{data}}$, that is, the mean squared error loss of DeepONet, can be added to the above expression to further improve accuracy and training stability.
    The PDE residual loss is obtained by differentiating the network output to high order, making use of automatic differentiation; for the Laplace equation, for instance, $\mathcal{J}_{\mathrm{PDE}}=\frac{1}{N}\sum_{i}\sum_{k}\left(\nabla^2\mathcal{G}_\theta(a^{(i)})(\mathbf{x}_k)\right)^2$.
    This allows the network to be trained with the help of the PDE constraint when interior solution labels are scarce, thereby alleviating to some extent the dependence on large amounts of labelled data.

However, PI-DeepONet incorporates physical information into the loss function as a soft constraint, which entails several inherent limitations. First, the weights $\lambda$ between the PDE residual, the boundary residual and the optional data term must be tuned manually; the optimal weights differ substantially from problem to problem, and imbalanced weights lead to unstable training.
    Second, the PDE residual loss requires repeated computation of high-order derivatives of the network output, which is computationally expensive.
    Third, even when training converges, the network output only approximately satisfies the governing equation at the collocation points, so that physical consistency is not rigorously guaranteed. These limitations show that the soft-constraint approach can hardly reconcile training stability with rigorous physical consistency.

\subsection{\texorpdfstring{Kernel Functions and Kernel-Expansion Methods for PDEs}{Kernel Functions and Kernel-Expansion Methods for PDEs}}

A kernel function is a bivariate function $K(\mathbf{x},\mathbf{y})$ defined on a product space. In function approximation, given a set of centres $\left\{\mathbf{x}_s^{(j)}\right\}_{j=1}^{N_c}$, a kernel function can be used directly as a basis function to expand a target function $u(\mathbf{x})$ as a linear combination of kernels,
\begin{equation}
u(\mathbf{x})\approx\sum_{j=1}^{N_c}\beta_j\,K\left(\mathbf{x},\mathbf{x}_s^{(j)}\right),
\end{equation}
where the coefficients $\beta_j$ are determined by data or by the equation.

As noted above, kernel methods have been widely used in machine learning, with a theoretical foundation given by the theory of reproducing kernel Hilbert spaces (RKHS) \cite{aronszajn1950theory}: when a kernel function is symmetric and positive definite, it induces a complete inner product space and becomes the reproducing kernel of that space; the relevant properties will be discussed in Section~2.5.

This section focuses on another use of kernel functions: as basis functions for the solutions of partial differential equations. Since the solution of a PDE is itself a function, if its approximation space is spanned by kernel functions, solving the PDE is reduced to determining the coefficients of the kernel expansion; when the kernel functions and the centres are suitably configured, this expansion can approximate the target function in the corresponding function space. The specific form of the kernel function determines the approximation capability of the approximation space, so that choosing a suitable kernel for a given problem is essential. In this paper, general kernel functions are denoted by $\varphi$ and physics-informed kernels such as fundamental solutions by $\Phi$. The remainder of this section first presents these two classes of kernel functions (Section~2.3.1) and then summarizes the kernel-expansion solution methods derived from them (Section~2.3.2).

\subsubsection{\texorpdfstring{General and Physics-Informed Kernels}{General and Physics-Informed Kernels}}

Depending on whether they contain physical information of the governing equation, kernel functions can be divided into two broad classes: general kernel functions and physics-informed kernel functions.
    A general kernel function does not depend on a specific equation but is determined only by geometric and smoothness requirements; the most typical example is the radial basis function (RBF) \cite{buhmann2003radial,wendland2005scattered}.
    An RBF depends only on the radial distance between two points, i.e. $K(\mathbf{x},\mathbf{y})=\varphi\left(\left|\mathbf{x}-\mathbf{y}\right|\right)$, and its rotational invariance makes it naturally suited to isotropic problems.
    Common RBFs include the Gaussian, multiquadric and thin-plate spline kernels, which are positive definite or conditionally positive definite and possess universal approximation capability \cite{micchelli1986interpolation,park1991universal,girosi1995regularization}; precisely because they do not satisfy any governing equation in advance, their form can be chosen with great flexibility, which makes them suitable for complex geometries and nonlinear problems.

    In contrast to general kernel functions, a physics-informed kernel function (PIKF) is a kernel function that contains, wholly or in part, information about the governing equation (PDE); it may be chosen as a fundamental solution, a Green's function, a harmonic function, a radial Trefftz function or even the solution of some linearly simplified PDE \cite{fu2024pikfnn,kita1995trefftz}.
    The most basic of these is the fundamental solution of the governing equation.
    Let $\mathcal{L}$ be a linear partial differential operator. If there exists a function $\Phi$ satisfying
\begin{equation}
\mathcal{L}\Phi=-\delta(\mathbf{x}-\mathbf{x}_s),
\end{equation}
where $\delta$ is the Dirac distribution, and $\mathbf{x}_s$ is the source point, then $\Phi$ is called the fundamental solution of the operator $\mathcal{L}$ at $\mathbf{x}_s$\cite{brebbia1984boundary}; unlike a Green's function, which further satisfies specific homogeneous boundary conditions, a fundamental solution carries no boundary conditions whatsoever.
    Taking the two-dimensional Laplace operator $\mathcal{L}=\nabla^2$ as an example, for a radial function the Laplacian in polar coordinates reduces to $\nabla^2=\dfrac{\partial^2}{\partial r^2}+\dfrac{1}{r}\dfrac{\partial}{\partial r}$;
    for a rotationally symmetric function $\Phi(r)$ that depends only on the radial distance, the equation $\nabla^2\Phi=-\delta(\mathbf{x}-\mathbf{x}_s)$ reduces, for $r\neq0$, to the ordinary differential equation
\begin{equation}
\Phi''(r)+\frac{1}{r}\Phi'(r)=0\quad\Longrightarrow\quad \Phi(r)=C_1\ln r+C_2.
\end{equation}
The constants of integration are determined by the flux condition at the source point, which gives the fundamental solution of the two-dimensional Laplace operator,
\begin{equation}
\Phi(r)=-\frac{1}{2\pi}\ln r,\qquad r=\left|\mathbf{x}-\mathbf{x}_s\right|,
\end{equation}
satisfying $\nabla^2\Phi=-\delta(\mathbf{x}-\mathbf{x}_s)$. Similarly, the fundamental solution of the Helmholtz operator $\mathcal{L}=\nabla^2+k^2$ is given by the zeroth-order Hankel function of the first kind:
\begin{equation}
\Phi(r)=\frac{i}{4}H_0^{(1)}(kr),
\end{equation}
which also automatically satisfies the Sommerfeld radiation condition at infinity and is therefore particularly suitable for unbounded exterior domain problems. Unlike a general kernel function, a PIKF embeds the physical information of the governing equation explicitly into the kernel function, so that the kernel automatically satisfies, or partially satisfies, the governing equation and thus naturally preserves physical consistency in function approximation.

\subsubsection{\texorpdfstring{Kernel-Expansion Solution Methods for PDEs}{Kernel-Expansion Solution Methods for PDEs}}

The two classes of kernel functions above correspond precisely to two kernel-expansion routes: general kernel functions to the domain-type route and physics-informed kernel functions to the boundary-type route. What these methods have in common is that the solution is written as a linear combination of kernel functions at a number of centres (source points); they differ in three respects---whether the kernel function satisfies the governing equation in advance, whether the centres lie in the whole domain or on the boundary, and whether the boundary conditions are enforced pointwise in a strong form or in a weak form through a boundary integral equation. Along these three dimensions, three basic classes of methods can be distinguished: (i)~domain-type collocation, which uses general kernel functions as basis functions with centres distributed throughout the domain; (ii)~boundary-type collocation, which uses physics-informed kernel functions as basis functions with centres (source points) placed on the boundary or on a virtual boundary close to it; and (iii)~boundary integral methods, which use fundamental solutions as integral kernels to construct single- or double-layer potentials on the boundary and replace the collocation coefficients by boundary densities. None of these three classes requires a volume mesh; in particular, (ii) and (iii) require only a boundary discretization, thereby alleviating the limitations imposed by mesh generation on complex geometries and large-scale problems. Beyond this framework, when the equation contains nonlinear terms and the operator no longer admits a usable fundamental solution, one may take the fundamental solution of the linear principal part as the kernel and supplement it with linearization or correction kernels, which constitutes a fourth class of extension (see item~(4)).

\paragraph{(1) Domain-type collocation}

Domain-type collocation places collocation points both in the interior of the solution domain and on its boundary, the classical representative being Kansa's method \cite{kansa1990multiquadrics}.
    This method uses general radial basis functions $\varphi$ as basis functions and expands the solution as a linear combination over $N$ centres $\{\mathbf{x}_j\}_{j=1}^{N}$ distributed throughout the domain,
\begin{equation}
u(\mathbf{x})\approx\sum_{j=1}^{N}\alpha_j\,\varphi\left(\left|\mathbf{x}-\mathbf{x}_j\right|\right),
\end{equation}
    and then requires the interior collocation points to satisfy the governing equation $\mathcal{L}u=f$ and the boundary collocation points to satisfy the boundary conditions; the linear system formed by these collocation equations determines the unknown coefficients $\alpha_j$.
    Since RBFs need not satisfy the governing equation in advance, domain-type collocation can handle complex geometries and nonlinear problems flexibly;
    the price is that the collocation points must cover the entire domain and that the resulting algebraic system is usually dense, so that the computational cost rises markedly as the number of degrees of freedom increases.
    Moreover, the collocation matrix of Kansa's method is in general non-symmetric and may be singular or severely ill-conditioned, its well-posedness depending on the collocation layout and on the choice of the shape parameter of the basis functions \cite{fasshauer2007meshfree}.

\paragraph{(2) Boundary-type collocation}

Unlike the domain-type methods, a boundary-type collocation method places collocation points only on the boundary of the solution domain and selects physics-informed kernel functions that automatically satisfy the governing equation as its basis functions.
    Since the governing equation is already satisfied exactly inside the domain by the basis functions, it suffices to enforce the boundary conditions at the boundary collocation points, and this feature reduces the dimension of the problem by one---a three-dimensional problem becoming two-dimensional, for instance---so that the number of collocation points required is greatly reduced.
    Its classical representative is the method of fundamental solutions (MFS) \cite{golberg1997method,fairweather1998method}, which uses the fundamental solution of the governing equation as the basis function and enjoys advantages such as exponential convergence and high accuracy \cite{golberg1995poisson,chen1998numerical,fairweather2003scattering}.
    To avoid the singularity of the fundamental solution at the source point, the MFS places the source points on a virtual boundary outside the physical domain.
    Let $\left\{\mathbf{x}_s^{(j)}\right\}_{j=1}^{N_s}$ be $N_s$ source points distributed outside the solution domain; then the solution can be approximated as
\begin{equation}
u(\mathbf{x})\approx\sum_{j=1}^{N_s}\alpha_j\,\Phi\left(\left|\mathbf{x}-\mathbf{x}_s^{(j)}\right|\right),
\end{equation}
where the coefficients $\alpha_j$ are determined by the boundary conditions; by the denseness of the solution space of elliptic equations (a Runge-type approximation theorem) \cite{fairweather1998method}, this kernel expansion approaches the true solution as the number of source points $N_s$ increases, when the source points are suitably configured, exhibiting spectral convergence for smooth problems.

A key difficulty of boundary-type collocation is the source-point layout: the source points must avoid the boundary in order to circumvent the singularity of the fundamental solution, yet their positions directly affect the condition number of the coefficient matrix and the accuracy. The MFS avoids the singularity by moving the source points outward onto a virtual boundary outside the solution domain, but the position of this virtual boundary is usually chosen by experience. The singular boundary method (SBM) \cite{gu2011singular,fu2020hybridfemsbm} instead places the source points directly on the physical boundary nodes and subtracts the singular part of the fundamental solution by means of source intensity factors, so that no virtual boundary is needed; the price is that a source intensity factor has to be estimated for every boundary node. Both classes of methods show that, although an expansion based on fundamental solutions is physically consistent and highly accurate, the distance between the source points and the boundary is a hyperparameter that must be handled with care.

This sensitivity to the hyperparameter stems from the analyticity of the fundamental-solution kernel: the source points of the MFS lie outside the domain, so that its kernel is analytic and smooth on the boundary and the corresponding boundary operator is a smoothing (infinitely differentiable) operator, whose singular values decay geometrically and whose condition number grows exponentially with the number of source points $N_s$. In other words, the MFS trades spectral accuracy for severe ill-conditioning, and the choice of the source distance is essentially a compromise between approximation accuracy and matrix conditioning.

\paragraph{(3) Boundary integral methods}

Besides serving as basis functions for collocation, fundamental solutions can also act as the integral kernel of boundary integral methods, and are used to construct boundary integral equations (BIEs). The starting point is to represent the solution inside the domain as a potential integral over the boundary; taking the single-layer potential constructed from a general fundamental solution $\Phi$ as an example,

\begin{equation}
u(\mathbf{x})=\int_{\Gamma}\Phi\left(\left|\mathbf{x}-\mathbf{y}\right|\right)\sigma(\mathbf{y})\,\mathrm{d}\Gamma_{\mathbf{y}},
\end{equation}
where $\sigma$ is the unknown boundary density. After discretizing the boundary into elements and approximating the integral numerically, the above expression can formally be written as a linear combination of basis functions,

\begin{equation}
u(\mathbf{x})\approx\sum_{j=1}^{N_e}\sigma_j\int_{\Gamma_j}\Phi\left(\left|\mathbf{x}-\mathbf{y}\right|\right)\mathrm{d}\Gamma_{\mathbf{y}},
\end{equation}
where $\sigma_j$ is the constant density on the $j$-th element and the basis functions are the integrals of the fundamental solution over that element (the self-influence element contains an integrable weak singularity). This structure is precisely the foundation of the boundary element method (BEM) \cite{brebbia1984boundary}: it shares the same fundamental-solution kernel with the MFS, the difference being that a weak-form integral equation replaces strong-form pointwise collocation. Because the integration is performed on the true boundary, the BEM needs to introduce no virtual source distance; although the self-influence element integral contains a weak (logarithmic) singularity, this singularity is integrable and can be handled exactly by analytic integration or by singularity subtraction.

From the point of view of operator properties, the kernel of the BEM is only weakly singular on the boundary and the operator smooths only to first order, so that its singular values mostly decay algebraically and its condition number grows polynomially with the number of elements $N_e$---usually milder than the exponential ill-conditioning of the MFS---but near corners the boundary density becomes singular and graded meshes or special elements must be used. For comparison, the virtual boundary method (VBM) \cite{sun1997virtual} places the boundary integral on a virtual boundary enclosing the solution domain, so that no singular integral has to be handled because the integration points are far from the true boundary; its price is the same as that of the MFS---the position of the virtual boundary must be chosen artificially and directly affects the computational accuracy and the condition number of the coefficient matrix.

\paragraph{(4) Extension to nonlinear problems}

When the governing equation contains nonlinear terms, the principle of superposition no longer holds and the operator has no usable fundamental solution in the usual sense, so that the kernel expansion formed by a linear superposition of fundamental solutions no longer applies. The typical way to handle such problems is to "approximate the nonlinear by the linear": the nonlinear operator is split into a linear principal part $\mathcal{L}_0$ and a nonlinear remainder, or the nonlinear term is linearized by a Newton or Picard iteration at the current approximate solution, reducing the original problem to a sequence of linear subproblems, each of which is then solved by the kernel expansion described above \cite{brebbia1984boundary,wrobel1987dual}.

After linearization, the nonlinear remainder is moved to the right-hand side of the equation and acts as an equivalent source term, so that each step still requires the solution of a linear equation of the form $\mathcal{L}_0u=g$. Its solution can be written as a homogeneous part spanned by the fundamental solution of the linear principal part plus a particular-solution kernel for the equivalent source term,
\begin{equation}
u(\mathbf{x})\approx\sum_{j=1}^{N_s}\alpha_j\,\Phi_0\left(\left|\mathbf{x}-\mathbf{x}_j\right|\right)+\sum_{k=1}^{K_c}\gamma_k\,\Psi_0\left(\left|\mathbf{x}-\mathbf{x}_k\right|\right),
\end{equation}
where $\Phi_0$ is the fundamental solution of the linear principal-part operator $\mathcal{L}_0$ and $\Psi_0$ is the particular-solution kernel of the equivalent source term, while $\alpha_j$ and $\gamma_k$ are the expansion coefficients, the latter being determined by a functional approximation of the equivalent source term. In the Newton or Picard iteration it suffices to substitute the current residual into the equivalent source term at each step, so that the same set of kernel expansions can be reused: the homogeneous part satisfies $\mathcal{L}_0\Phi_0=\delta$ exactly, while all the nonlinear information is carried by the particular-solution kernel and updated as the iteration proceeds. This construction---"fundamental solution of the linear principal part plus nonlinear correction", of which the dual reciprocity method is a classical realization \cite{nardini1983new,wrobel1987dual,partridge1991dual}---gives kernel-expansion methods the ability to handle nonlinear problems while preserving physical consistency, and it is also the common starting point of all kinds of nonlinear kernel methods, whether domain-type collocation or boundary-type expansion.

In summary, kernel functions occupy a central position in kernel-expansion-based PDE solution methods: domain-type collocation uses general radial basis functions as kernels and is flexible in form but requires a domain-wide discretization; boundary-type collocation and boundary integral methods take the fundamental solution (PIKF) of the governing equation as their core and embed the physical information of "automatically satisfying the governing equation" explicitly into the basis functions, thereby obtaining higher accuracy with fewer degrees of freedom; and for nonlinear problems one takes the fundamental solution of the linear principal part and supplements it with linearization or correction kernels. It can thus be seen that domain-type collocation, boundary-type collocation and boundary integral methods share one and the same library of kernel functions and differ only in the choice of kernel function and in the way the boundary conditions are imposed, and are therefore unified at the level of the kernel expansion.

\subsection{Kernel Operator Network}

\begin{figure}[htbp]
\centering
\includegraphics[width=0.98\textwidth]{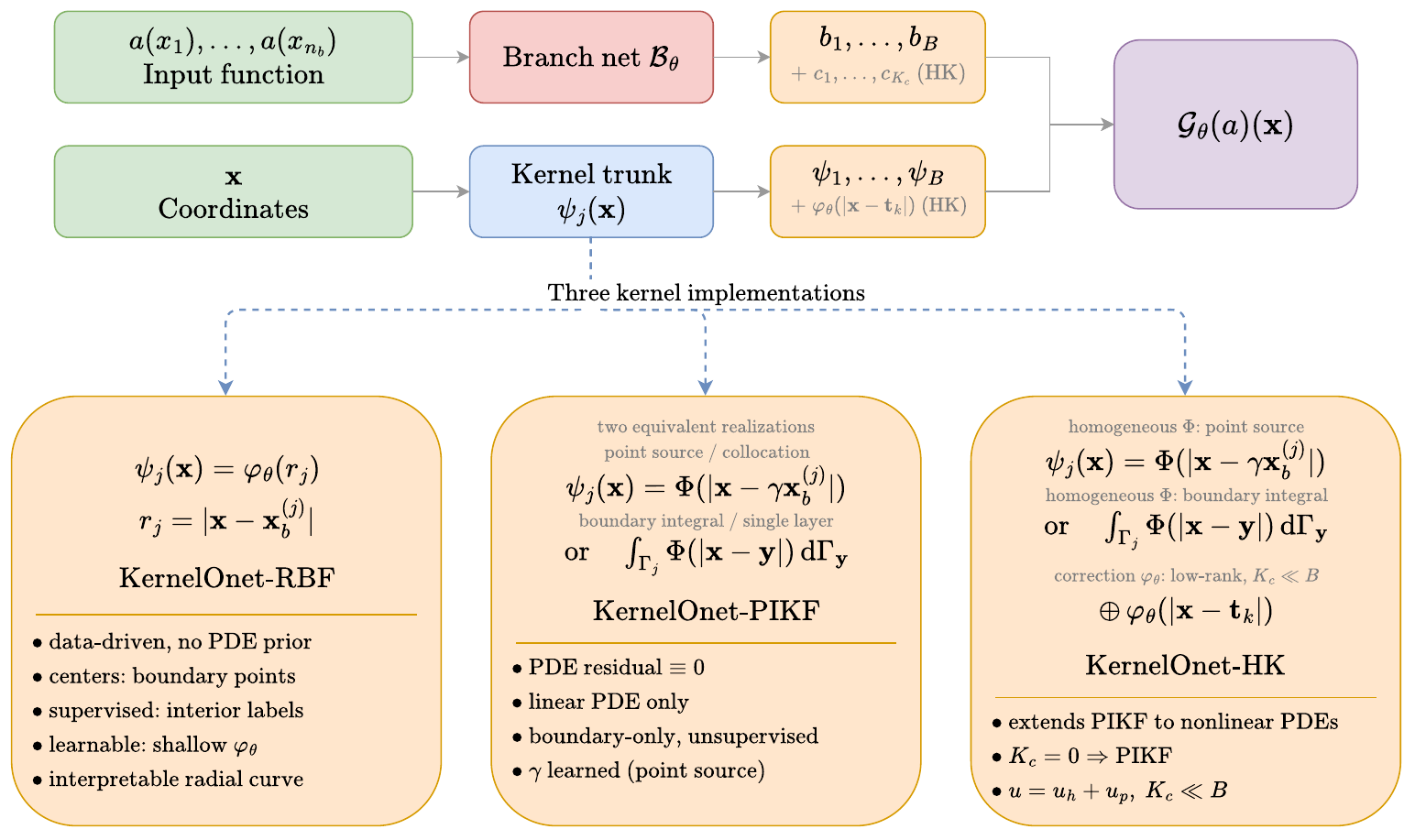}
\caption{Architecture of KernelOnet. The three variants share the same branch network and the same operator expansion structure $u(\mathbf{x})=\sum_{j=1}^{B}b_j(a)\,\psi_j(\mathbf{x})$, differing only in the kernel trunk $\psi_j(\mathbf{x})$: KernelOnet-RBF uses a neural-network-parameterized radial basis function; KernelOnet-PIKF takes the analytic fundamental solution as its kernel (the collocation form uses a $\gamma$-shifted fundamental solution, the boundary integral form a single-layer potential kernel on the true boundary) and its expansion satisfies the governing equation exactly, so that it can be trained without supervision; KernelOnet-HK mixes analytic fundamental-solution kernels with low-rank learned correction kernels, $\psi=\Phi(\gamma\,\mathbf{x}_b)\oplus \varphi_\theta(\mathbf{x}-\mathbf{t}_k)$, absorbing the nonlinear source terms into a correction branch with $K_c\ll B$ and requiring supervised training.}
\label{fig:kernelonet}
\end{figure}

The kernel-expansion perspective of Section~2.3 has a direct implication for the design of neural operators: the trunk network of DeepONet \cite{lu2021learning} can be regarded as learning basis functions implicitly in a data-driven sense, whereas the Kernel Operator Network (KernelOnet) proposed in this paper makes the kernel functions explicit, so that the network structure and the kernel expansion are mathematically consistent---the coefficients are given by the branch network and the kernel functions by the trunk network. Based on this observation, the general form of KernelOnet is
\begin{equation}
\mathcal{G}_\theta(a)(\mathbf{x})=\sum_{j=1}^{B}b_j(a)\,\psi_j(\mathbf{x}),
\end{equation}
where the coefficients are given by the branch network $\mathbf{b}(a)=\mathcal{B}_\theta(a)$ and the family of basis functions $\psi_j(\mathbf{x})$ is no longer learned implicitly by a free network but is explicitly constrained to be a kernel function.
    Figure~\ref{fig:kernelonet} shows the overall architecture of KernelOnet: the three variants share the same branch network and the same operator expansion structure and differ only in the construction of the kernel trunk $\psi_j(\mathbf{x})$---KernelOnet-RBF parameterizes a radial basis function by a neural network and learns the kernel shape from data;
    KernelOnet-PIKF takes the analytic fundamental solution as its kernel, and admits two equivalent forms, collocation and boundary integral (the former uses a $\gamma$-shifted fundamental solution, the latter a single-layer potential integral over the true boundary), so that the expansion satisfies the governing equation exactly;
    KernelOnet-HK, on the other hand, mixes analytic fundamental-solution kernels with low-rank learned correction kernels, absorbing the nonlinear source terms into a correction branch with $K_c\ll B$.
    The three variants are introduced in the following three subsections.

\subsubsection{KernelOnet-RBF: Data-Driven Learnable Kernel}

KernelOnet-RBF learns the kernel function directly in a data-driven manner, using a neural-network-parameterized radial basis function as the kernel:
\begin{equation}
\psi_j(\mathbf{x})=\varphi_\theta\left(r_j\right),\qquad r_j=\left|\mathbf{x}-\mathbf{x}_b^{(j)}\right|,
\end{equation}
where $\varphi_\theta$ is a univariate shallow network taking only the radial distance $r$ as input, and is learned jointly with the branch network during training. The kernel centres are taken directly as the boundary discretization points $\left\{\mathbf{x}_b^{(j)}\right\}_{j=1}^{B}$, so that every output coefficient of the branch network corresponds exactly to the weight of the basis function at one boundary point, with a clear physical meaning. This kernel function does not assume a specific form in advance but learns the radial shape adaptively from data. For constant-coefficient linear equations, by the approximation theory of Section~2.5 the solution can be spanned by an expansion of fundamental solutions at the boundary centres, and the fundamental solution is the optimal choice for a single radial kernel; in this case the learned kernel function can be regarded as a regularized approximation of the fundamental solution. It must be emphasized that this interpretation holds only in that case: once the equation contains nonlinear source terms or variable coefficients, the solution can no longer be spanned by an expansion of a single fundamental solution, and the kernel function has to fit both the homogeneous field and the source term, so that it can no longer be interpreted as a fundamental solution. At the same time, its inductive bias of depending only on the radial distance guarantees rotational invariance, consistent with the symmetry of isotropic fundamental solutions in constant-coefficient linear problems. KernelOnet-RBF is trained with the data loss of Section~2.2 and therefore belongs to supervised learning; its range of applicability is also the broadest of the three variants, since it can naturally handle strongly nonlinear problems, variable coefficients and complex geometries for which no analytic fundamental solution exists.
    The learned kernel function has the potential to transfer across problems and can still offer a certain degree of interpretability in the form of a readable radial curve.
    The price is that the physical guarantee of "the expansion automatically satisfying the governing equation" is lost entirely: the form to which the kernel function converges is determined completely by data, and physical consistency is only soft and indirect.

\subsubsection{KernelOnet-PIKF: Physics-Informed Kernel}

KernelOnet-PIKF embeds physical information explicitly into the kernel function.
    It rests on the concept of the physics-informed kernel function (PIKF) introduced in Section~2.3: taking a PIKF as the basis function of the trunk network embeds PDE information directly into the network structure, without the need to add PDE constraints to the loss function as PINNs do; this idea has been validated by the physics-informed kernel function neural network (PIKFNN) proposed by Fu et al. \cite{fu2024pikfnn}.
    KernelOnet-PIKF takes the analytic fundamental solution of the governing equation as its kernel and constructs the following collocation expansion:
\begin{equation}
\mathcal{G}_\theta(a)(\mathbf{x})=\sum_{j=1}^{B}b_j(a)\,\Phi\left(\left|\mathbf{x}-\gamma\,\mathbf{x}_b^{(j)}\right|\right),
\label{eq:pikf_colloc}
\end{equation}
where $\Phi$ is the analytic fundamental solution of the governing equation, $\gamma$ is a learnable radial scaling scalar and $b_j(a)$ is given by the branch network; the source points are the boundary discretization points $\mathbf{x}_b^{(j)}$ scaled by $\gamma$. Since $\Phi$ satisfies $\mathcal{L}\Phi=-\delta(\mathbf{x}-\mathbf{x}_s)$ (where $\mathcal{L}$ is the very governing equation to be approximated) and the source points lie outside the solution domain, each basis function satisfies the homogeneous governing equation inside the domain, so that for arbitrary values of the coefficients the expansion automatically satisfies the governing equation inside the domain and the PDE residual vanishes identically. Consequently KernelOnet-PIKF requires no interior solution data and no computation of a PDE residual term for training; it suffices to enforce the boundary conditions on the boundary, which realizes unsupervised operator learning in the true sense.

Within the same boundary-type framework, KernelOnet-PIKF can also be written as a kernel expansion in boundary integral (single-layer potential) form, that is, with the fundamental-solution kernel distributed on the true boundary:

\begin{equation}
\mathcal{G}_\theta(a)(\mathbf{x})=\int_{\Gamma}\Phi\left(\left|\mathbf{x}-\mathbf{y}\right|\right)b(a)(\mathbf{y})\,\mathrm{d}\Gamma_{\mathbf{y}}\;\approx\;\sum_{j=1}^{B}b_j(a)\int_{\Gamma_j}\Phi\left(\left|\mathbf{x}-\mathbf{y}\right|\right)\mathrm{d}\Gamma_{\mathbf{y}},
\label{eq:pikf_bie}
\end{equation}
where $b(a)$ is the boundary density function output by the branch network and $\Gamma_j$ is the $j$-th boundary patch. Both expressions take the fundamental solution $\Phi$ as their kernel and therefore belong to the same PIKF framework, corresponding respectively to the boundary collocation and the boundary integral implementation. The two are not the same limit: in the integral form the source distribution lies on the true boundary $\Gamma$, and although the self-influence element integral contains a weak singularity (logarithmic in two dimensions) it is integrable, so that the value at a boundary collocation point is finite and no source point needs to be moved; the collocation form, by contrast, is a point source of zero measure, and when the source point falls on the boundary the kernel is unbounded there so that the collocation equation cannot hold directly, and the source points must be moved off the boundary. In other words, the point-source form corresponds to a kernel expansion on a virtual boundary while the integral form corresponds to a kernel expansion on the true boundary.

Thus KernelOnet-PIKF structurally unifies the boundary-type kernel methods of Section~2.3: its collocation form agrees with the collocation forms of the MFS \cite{golberg1997method}, the Trefftz method \cite{kita1995trefftz} and the SBM \cite{gu2011singular}, while its boundary integral form shares the same integral kernel as the single-layer potential of the BEM \cite{brebbia1984boundary}. The difference is that the expansion coefficients are no longer obtained by solving simultaneously for the source-point layout and the boundary conditions but are given directly by the branch network, and that in the collocation form the source points give a virtual surface automatically through the scaling of the boundary points, which removes the need for the virtual boundary selection of the MFS/VBM and for the source intensity factor estimation of the SBM; the boundary integral form, meanwhile, performs panel integration directly on the true boundary.

The training loss contains only the boundary residual:
\begin{equation}
\mathcal{J}(\theta)=\frac{1}{N\,B}\sum_{i=1}^{N}\sum_{k=1}^{B}\left|\mathcal{G}_\theta\left(\mathbf{y}_b^{(i)}\right)\left(\mathbf{x}_b^{(k)}\right)-y_{b,k}^{(i)}\right|^2,
\end{equation}
where $\mathbf{y}_b^{(i)}=\left(y_{b,1}^{(i)},\ldots,y_{b,B}^{(i)}\right)$ is the boundary condition of the $i$-th sample, so that the supervision signal is the input itself. Unlike physics-informed neural networks (PINNs) \cite{raissi2019physics} and the physics-informed DeepONet (PI-DeepONet) \cite{wang2021learning}, which add the PDE residual to the loss as a soft constraint and satisfy the governing equation only approximately through optimization, KernelOnet-PIKF hard-codes the physical information into the kernel structure, which both avoids the computational cost of repeatedly evaluating high-order PDE residuals and eliminates the problem of balancing the weights between the PDE residual and the boundary residual.

The above expansion and training scheme can be extended directly to the complex case: taking the complex Helmholtz equation in an unbounded exterior domain as an example, one chooses the complex fundamental solution $\Phi(r)=\frac{i}{4}H_0^{(1)}(kr)$ as the kernel, and both the kernel function and the expansion coefficients take complex values, so that the form of the expansion and of the loss remains unchanged, while the real and imaginary parts of the output solution are given by the cross inner products of the real and imaginary parts of the coefficients with those of the kernel function. Since the complex fundamental solution itself satisfies the governing equation and the Sommerfeld radiation condition exactly, the above properties continue to hold in the complex case, as will be verified in Example~3.

Moreover, because the kernel function is taken directly as the analytic fundamental solution, KernelOnet-PIKF has the strongest interpretability of the three variants, since after training the physics learned by the network can be read off pointwise; the price is a strict precondition: the whole governing equation must admit an analytic fundamental solution and the solution must satisfy the principle of superposition, so that it applies only to linear problems that themselves possess fundamental solutions, such as Laplace, Helmholtz, modified Helmholtz, biharmonic and convection--diffusion problems. The fundamental solutions of common operators are given in Appendix~\ref{app:fund}. Among them, the fundamental solution automatically satisfies the radiation condition at infinity, which makes it particularly suitable for unbounded exterior domains and wave propagation problems; for nonlinear problems whose governing equation admits no fundamental solution, KernelOnet-PIKF no longer applies and KernelOnet-HK, introduced in the next subsection, should be used instead.

When the family of kernel functions is numerically ill-conditioned under the given discretization of source and evaluation points, so that using it directly as a kernel would pollute the gradients, the kernel basis must be orthogonalized as a preprocessing step. In this paper, the kernel function matrix $G(\mathbf{x})$ is orthogonalized by a singular value decomposition (SVD) of the kernel basis, taking the first $q$ principal singular directions to form a well-conditioned orthogonal kernel basis $U_q(\mathbf{x})$ (the criterion for selecting the retained directions is given in Appendix~\ref{app:svd}: usually the principal singular directions of the boundary kernel matrix are taken, but when the magnitude of the evaluation block differs greatly from that of the boundary block the retained directions must instead be selected from the input subspace, which is the case for the far field of Example~4), so that the operator output in both the collocation and the boundary integral form can be written uniformly as $\mathcal{G}_\theta(a)(\mathbf{x})=U_q(\mathbf{x})\,\mathbf{a}$; this preprocessing does not destroy the physical consistency of the expansion. Its construction details, truncation error analysis, continuous basis extension and coupling with source-point learning and offline preprocessing are given in Appendix~\ref{app:svd}.

\subsubsection{KernelOnet-HK: Hybrid Kernel}

When the governing equation contains nonlinear terms and its operator admits no usable fundamental solution, the principle of superposition no longer holds and the whole equation cannot be represented by a linear superposition of a single fundamental solution. If KernelOnet-PIKF of Section~2.4.2 were still used in this case, with the fundamental solution of its linear principal part as the kernel, the expansion would satisfy the homogeneous equation $\mathcal{L}u_h=0$ identically and be unable to represent the solution driven by the nonlinear remainder, thus losing its approximation capability.
    For this reason, this section proposes a third construction strategy for KernelOnet---the hybrid kernel (HK)---whose central idea is to decompose the solution according to the linear principal part of the governing equation: the solution is written as the sum of a homogeneous branch and a correction branch,
\begin{equation}
u=u_{h}+u_{p},\qquad
\mathcal{L}\,u_{h}=0,\qquad
\mathcal{L}\,u_{p}=-N[u],
\label{eq:hk_decomp}
\end{equation}
where $\mathcal{L}$ is the linear principal part of the governing equation and $N[u]$ is the nonlinear remainder. Here $u_{h}$ is the homogeneous branch, lying in the null space of $\mathcal{L}$ and representable by an expansion of the fundamental solution $\Phi$ of $\mathcal{L}$; the nonlinearity is concentrated entirely in the correction branch $u_{p}$, which is determined by the source term $-N[u]$ and in general no longer satisfies $\mathcal{L}u_p=0$, so that it must be represented by a family of basis functions that do not satisfy this homogeneous equation.
    Accordingly, KernelOnet-HK takes the kernel function to be a mixture of an analytic fundamental-solution kernel and a low-rank learned correction kernel. As in Section~2.4.2, the homogeneous branch admits two equivalent forms, collocation (a $\gamma$-shifted fundamental solution) and boundary integral (a single-layer potential kernel on the true boundary); the collocation form is taken as an example below:
\begin{equation}
\mathcal{G}_\theta(a)(\mathbf{x})
=\underbrace{\sum_{j=1}^{B}b_j(a)\,\Phi\!\left(\left|\mathbf{x}-\gamma\,\mathbf{x}_b^{(j)}\right|\right)}_{\text{analytic fundamental-solution kernel (satisfies }\mathcal{L}u_h=0\text{, a physical prior)}}
+\underbrace{\sum_{k=1}^{K_c}c_k(a)\,\varphi_\theta\!\left(\left|\mathbf{x}-\mathbf{t}_k\right|\right)}_{\text{learned source-term correction kernel (low-rank, }K_c\ll B\text{)}},
\label{eq:hk_expand}
\end{equation}
where $\Phi$ is the analytic fundamental solution of the linear principal part $\mathcal{L}$ (the definition of $\gamma$ is given in Section~2.4.2); $\varphi_\theta$ is a shallow network taking only the radial distance as input, belonging to the same family as the kernel function of KernelOnet-RBF; $\left\{\mathbf{t}_k\right\}_{k=1}^{K_c}$ are $K_c$ fixed centres of the correction kernel, sampled inside the solution domain by Latin hypercube sampling (LHS) (for unbounded exterior problems the centres are taken in a bounded region inside the obstacle); and $b_j(a)$ and $c_k(a)$ are output at once by the same branch network, whose output dimension is $B+K_c$.
    Since the parameters $\theta$ of $\varphi_\theta$ are shared across all samples and only the coefficients $c_k(a)$ vary with the input, for any given input the correction term can only lie in the
\begin{equation}
\mathcal{V}_{K_c}=\mathrm{span}\left\{\varphi_\theta\!\left(\left|\mathbf{x}-\mathbf{t}_1\right|\right),\ldots,\varphi_\theta\!\left(\left|\mathbf{x}-\mathbf{t}_{K_c}\right|\right)\right\}
\label{eq:hk_span}
\end{equation}
$K_c$-dimensional subspace. Thus $K_c\ll B$ limits the dimensionality occupied by the data-driven part, so that the analytic fundamental-solution kernel continues to dominate the expansion and the model does not degenerate into the purely data-driven KernelOnet-RBF, while the correction branch is guaranteed to have a clear physical meaning rather than being a black box.

    The hybrid kernel also connects the two endpoints of Sections~2.4.1 and 2.4.2 through a single tunable dimension $K_c$: when $K_c=0$ the correction branch disappears and the model degenerates to KernelOnet-PIKF; when $K_c=B$, and the centres and basis-function form of the correction kernels agree with those of KernelOnet-RBF and the weights of the first branch tend to zero, it degenerates to KernelOnet-RBF. The three variants are therefore not independent of one another but form a continuous spectrum from "strong physical prior" to "fully data-driven": PIKF and RBF are the two endpoints and HK is the intermediate form continuously tuned between them by $K_c$.It should be noted that the RBF limit is idealized: it requires the centres and the basis-function form of the correction kernel to agree with those of KernelOnet-RBF.

    Since the correction branch destroys the property that the expansion satisfies the governing equation exactly, training KernelOnet-HK requires interior solution data for supervised learning, and its data loss is
\begin{equation}
\mathcal{J}_{\mathrm{data}}(\theta)=\frac{1}{N\,n_t}\sum_{i=1}^{N}\sum_{k=1}^{n_t}\left(u_{\mathrm{pred},k}^{(i)}-u_{\mathrm{true},k}^{(i)}\right)^{2}.
\label{eq:hk_loss}
\end{equation}
The physical prior enters the model structure only through the analytic kernel basis (as a hard constraint) and the training loss contains no PDE residual term; compared with purely data-driven methods that require large numbers of interior solution labels, its training does not rely on interior labels at all.
     It should be emphasized that the treatment in this paper introduces no inference-time iteration: the value of operator learning lies in amortized inference---once training is complete, a single forward pass yields the solution for a new input function. This construction is consistent with the idea described in Section~2.3.2 of "taking the fundamental solution of the linear principal part as the homogeneous kernel and absorbing the nonlinearity into a particular-solution kernel for the source term", the difference being that here the particular-solution kernel is obtained from data through the low-rank learned branch.
    The hybrid kernel also brings interpretability: after training, the coefficient norms $\|b(a)\|$ and $\|c(a)\|$ of the homogeneous and correction branches can be read off separately, so as to determine quantitatively the proportions of the "homogeneous component" ($\mathcal{L}u_h=0$) and the "source component" in the solution; the correction kernel centres $\left\{\mathbf{t}_k\right\}$ and the learned radial shape $\varphi_\theta$ also provide physical insight into the spatial distribution of the nonlinear source term.
    When the governing equation itself admits a fundamental solution, $N[u]\equiv0$ in Eq.~\eqref{eq:hk_decomp}, and one may take $u_p\equiv0$, so that KernelOnet-HK degenerates to KernelOnet-PIKF and linear problems are still handled by PIKF in an unsupervised manner.

\subsubsection{Comparison and Summary}

Table~\ref{tab:method_compare} summarizes the similarities and differences among DeepONet, PI-DeepONet and the three KernelOnet variants from seven aspects: the form of the basis functions, the way physical information is embedded, PDE consistency, the supervision information, the learnable kernel parameters, and the applicability to nonlinear and unbounded-domain problems.
    As regards the form of the basis functions, DeepONet and PI-DeepONet learn the basis functions implicitly through a network, whereas KernelOnet makes the basis functions explicit as kernel functions---a neural-network-parameterized radial basis function for KernelOnet-RBF, an analytic fundamental-solution kernel (with two equivalent forms, collocation and boundary integral) for KernelOnet-PIKF, and a mixture of the analytic fundamental-solution kernel with low-rank learned correction kernels for KernelOnet-HK (Eq.~\eqref{eq:hk_expand}).
    As regards the way physical information is embedded and PDE consistency, DeepONet makes no use of physical information at all and cannot satisfy PDE consistency, PI-DeepONet adds the PDE residual to the loss as a soft constraint and satisfies the equation only approximately at the collocation points, the kernel function of KernelOnet-RBF does not satisfy the governing equation in advance, KernelOnet-PIKF hard-codes the physical information into the kernel function so that the expansion satisfies the governing equation exactly, and in KernelOnet-HK the homogeneous branch satisfies the linear principal part exactly while the correction branch carries the nonlinear source term, so that the expansion as a whole is satisfied only approximately.
    As regards the supervision information, DeepONet, KernelOnet-RBF and KernelOnet-HK all require interior solution labels, PI-DeepONet requires boundary conditions together with the PDE residual, and KernelOnet-PIKF requires only boundary conditions and can be trained without supervision.
    As regards the learnable kernel parameters, the basis functions of the two DeepONet variants are learned by the network as a whole and require no dedicated kernel parameters (PI-DeepONet requires tuning of the loss weights $\lambda$), KernelOnet-RBF learns a shallow kernel function network, KernelOnet-PIKF learns only one scalar $\gamma$ in its collocation form (the boundary integral form contains no learnable kernel parameter), and KernelOnet-HK learns a low-rank correction kernel network $\varphi_\theta$ (whose $K_c$ correction kernel centres are fixed sampling points, with $K_c\ll B$)---the strong inductive bias brought by explicit kernel functions reduces the number of learnable parameters and yields higher accuracy and better interpretability for the same amount of data.
    As regards the range of applicability, for nonlinear problems KernelOnet-PIKF does not apply, KernelOnet-HK applies only when the linear principal part admits an analytic fundamental solution, and KernelOnet-RBF can cover more general problems such as variable coefficients and strong nonlinearities; for unbounded-domain problems, only KernelOnet-PIKF and KernelOnet-HK, which take analytic fundamental solutions as kernels, are applicable.

In summary, the three kernel construction strategies proposed in this section form a continuous spectrum from a strong physical prior to fully data-driven learning, complementing rather than replacing one another. This trade-off between "strength of the physical prior" and "breadth of applicability" is precisely the design motivation behind the three variants in Figure~\ref{fig:kernelonet} sharing one architecture and differing only in the kernel trunk, and it will be systematically verified in the four examples of Section~3.

Each of the three variants continues one of the existing kernel-expansion routes of Section~2.3: KernelOnet-RBF continues the domain-type collocation route and can be regarded as the counterpart in operator learning of Kansa's method with general radial basis functions as kernels; KernelOnet-PIKF continues the boundary-type route, its collocation form being of the same origin as the method of fundamental solutions, the Trefftz method and the singular boundary method, while its boundary integral form shares the same integral kernel as the single-layer potential of the boundary element method; and KernelOnet-HK continues the nonlinear extension route, its construction of "fundamental solution of the linear principal part plus source-term correction" being consistent with Newton collocation and the dual reciprocity method \cite{nardini1983new,partridge1991dual}. The three thus bring all three technical routes of traditional kernel-expansion methods into the operator learning framework.

\begin{table}[htbp]
\centering
\small
\caption{Comparison of DeepONet, PI-DeepONet, KernelOnet-RBF, KernelOnet-PIKF and KernelOnet-HK.}
\label{tab:method_compare}
\setlength{\tabcolsep}{4pt}
\begin{tabular}{>{\raggedright\arraybackslash\hyphenpenalty=10000}p{2.0cm}>{\centering\arraybackslash}p{2.3cm}>{\centering\arraybackslash}p{2.3cm}>{\centering\arraybackslash}p{2.4cm}>{\centering\arraybackslash}p{2.7cm}>{\centering\arraybackslash}p{2.5cm}}
\toprule
 & \textbf{DeepONet} & \textbf{PI-DeepONet} & \textbf{KernelOnet-RBF} & \textbf{KernelOnet-PIKF} & \textbf{KernelOnet-HK} \\
\midrule
Form of basis functions & Implicitly learned basis functions & Implicitly learned basis functions & NN-parameterized radial basis function $\psi(r)=\varphi_\theta(r)$ & Analytic fundamental-solution kernel: collocation / boundary integral & Analytic fundamental solution $+$ low-rank correction kernel $\Phi\oplus \varphi_\theta$ \\
Embedding of physical information & None & Soft-constrained PDE residual & Data-driven kernel function & Hard-coded into the kernel function & Hard-coded $+$ low-rank source-term correction \\
PDE consistency & Not satisfied & Approximately satisfied, collocation only & Not satisfied in advance & Exactly satisfied, $\mathcal{L}u=0$ in the domain & Approximately satisfied, homogeneous branch exact and correction branch carrying the source term \\
Supervision information & Interior solution labels & Boundary conditions $+$ PDE residual & Interior solution labels & Boundary conditions only, unsupervised & Interior solution labels \\
Learnable kernel parameters & None & None & Shallow kernel function network & Collocation: 1 scalar $\gamma$; integral: none & Low-rank correction network $\varphi_\theta$ \\
Nonlinear problems & Applicable & Applicable & Applicable & Not applicable & Applicable \\
Unbounded-domain problems & Not applicable & Not applicable & Not applicable & Applicable & Applicable \\
\bottomrule
\end{tabular}
\end{table}

\subsection{Theoretical Analysis: Convergence and Accuracy}

This section analyses the convergence and accuracy of the three KernelOnet variants at the level of approximation theory, starting from kernel function theory, in order to clarify the mathematical basis of the trade-off between "strength of the physical prior" and "breadth of applicability" described in Section~2.4. For the sake of uniformity, the operator expansions of the three variants are all regarded as approximations of the solution in an approximation space spanned by kernel functions: given $B$ centres (source points) $\left\{\mathbf{x}_s^{(j)}\right\}_{j=1}^{B}$, define the approximation space
\begin{equation}
\mathcal{V}_{B}=\mathrm{span}\left\{\psi_1,\psi_2,\ldots,\psi_B\right\},
\end{equation}
    so that the prediction $\mathcal{G}_\theta(a)$ of KernelOnet is an element of $\mathcal{V}_{B}$ whose coefficients are given by the branch network.
    For KernelOnet-HK the approximation space additionally contains $K_c$ low-rank correction kernels, that is, the direct sum $\mathcal{V}_{B}\oplus\mathcal{V}_{K_c}$ (with $K_c\ll B$), and its error is correspondingly divided into a homogeneous-branch part and a source-branch part, corresponding respectively to $u_h$ and $u_p$ in Eq.~\eqref{eq:hk_decomp}.
    Different from the above division according to the components of the solution (homogeneous branch and source branch), the total error can also be decomposed by origin into two parts: first, the "approximation error" of the approximation space $\mathcal{V}_{B}$ with respect to the true solution, that is, whether the shape of the kernel function "fits" the true solution; and second, the "estimation error" introduced by the determination of the coefficients and by training.
    The differences among the three variants are essentially differences in the former, that is, in the choice of the kernel function space.

\paragraph{(1) Error bound in the reproducing kernel space framework---the accuracy basis of KernelOnet-RBF} KernelOnet-RBF uses the radial basis function $\psi(r)=\varphi_\theta(r)$ as its kernel.
    In reproducing kernel Hilbert space (RKHS) theory \cite{aronszajn1950theory,schaback2006kernel}, a radial basis function is rigorously regarded as the reproducing kernel of some function space (its native space), from which the standard error bound for interpolation approximation follows. It should be pointed out that this bound presupposes that the kernel function is strictly or conditionally positive definite; the $\varphi_\theta$ learned by KernelOnet-RBF does not automatically satisfy this condition, so that the bound holds only when it is (approximately) positive definite or is made positive definite by regularization.
    Let $\Omega$ be the solution domain, let $\varphi$ be a strictly or conditionally positive definite radial basis function with native space $\mathcal{N}_{\varphi}(\Omega)$, and let $X=\{\mathbf{x}_s^{(j)}\}$ be the discrete set of centres; then for any $u\in\mathcal{N}_{\varphi}(\Omega)$, the RBF interpolation approximant $s$ satisfies the pointwise bound
\begin{equation}
|u(\mathbf{x})-s(\mathbf{x})|\le P_{\varphi,X}(\mathbf{x})\,|u|_{\mathcal{N}_{\varphi}},
\end{equation}
where $P_{\varphi,X}(\mathbf{x})$ is the power function associated with the kernel function and the centre distribution and $|u|_{\mathcal{N}_{\varphi}}$ is the semi-norm of the native space \cite{wendland2005scattered,buhmann2003radial}.
    The dependence of the error on the density of the centre distribution is characterized by the fill distance $h_{X,\Omega}=\sup_{\mathbf{x}\in\Omega}\min_{j}|\mathbf{x}-\mathbf{x}_s^{(j)}|$: as the centres become denser, $h_{X,\Omega}\to0$ and the approximation error converges at an order determined by the smoothness of the kernel function---algebraic convergence for finitely smooth kernels, and spectral convergence for smooth kernels such as the Gaussian and multiquadric kernels.
    At the same time, Schaback and Wendland \cite{schaback2006kernel} revealed the well-known "uncertainty principle" of kernel methods: there is an intrinsic trade-off between the error and the condition number, the two cannot be improved simultaneously, and good accuracy is inevitably accompanied by severe ill-conditioning.
    This result explains theoretically the characteristics of KernelOnet-RBF---its kernel function needs finite truncation and regularization to maintain numerical well-posedness, while under data-driven learning it adaptively learns a kernel shape that "fits" the true solution, thereby gaining approximation room within a broader family of kernel shapes.

\paragraph{(2) Spectral convergence of the fundamental-solution expansion---the accuracy basis of KernelOnet-PIKF} KernelOnet-PIKF uses the fundamental solution $\Phi$ of the governing equation as its kernel.
    Since the source points lie outside the domain, the fundamental solution satisfies the homogeneous governing equation inside the domain, so that the expansion $\mathcal{G}_\theta(a)$ automatically satisfies the governing equation and its approximation error is determined entirely by the residual of the boundary condition fitting; this is mathematically the same structure as the method of fundamental solutions (MFS) \cite{golberg1997method,fairweather1998method} and the Trefftz method \cite{kita1995trefftz}.
    For such boundary-type methods, when the solution is analytic or sufficiently smooth in the domain, the family of fundamental solutions is inherently of high resolution and the approximation error exhibits "spectral convergence" as the number of basis functions $B$ increases, approaching exponential or geometric decay in the sufficiently smooth case, its convergence rate depending only on the analyticity of the solution and on the configuration of the source points $\gamma$, and not on any mesh size.
    In particular, by the denseness of the solution space of elliptic equations (a Runge-type approximation theorem), the space spanned by the family of fundamental solutions with source points distributed outside the domain is dense in the space of analytic solutions of the homogeneous equation, and the approximation theory of the MFS guarantees the spectral accuracy of the interpolation approximation, which provides theoretical support for "obtaining high accuracy with the fewest degrees of freedom";
    moreover, once $\gamma$ departs from 1 the family of fundamental solutions $\Phi(|\mathbf{x}-\gamma\mathbf{x}_b|)$ tends to become linearly dependent and its condition number rises as the source distance increases, which is exactly why Section~2.4 makes the source-point scaling a learnable scalar adapted $\gamma$, so as to compromise between avoiding the singularity and maintaining well-posedness.
    In summary, the core advantage of KernelOnet-PIKF is that its kernel space $\mathcal{V}_{B}$ is constrained within an approximation subspace of the homogeneous solution space, so that the approximation space matches the physical structure of the solution naturally and the approximation error can be made arbitrarily small; this is the root of its ability to achieve high accuracy with very few parameters and in an unsupervised manner.

\paragraph{(3) Decomposition error of the source-term correction---bias analysis of KernelOnet-HK} KernelOnet-HK takes the approximation space to be the direct sum of the analytic fundamental-solution subspace and the low-rank correction subspace, $\mathcal{V}_{B}\oplus\mathcal{V}_{K_c}$, and its total error is correspondingly divided into two parts according to the decomposition of Eq.~\eqref{eq:hk_decomp}:
\begin{equation}
u-\mathcal{G}_\theta(a)
=\underbrace{\bigl(u_{h}-u_{h,\theta}\bigr)}_{\text{approximation error of the homogeneous branch}}
+\underbrace{\bigl(u_{p}-u_{p,\theta}\bigr)}_{\text{approximation error of the source branch}},
\label{eq:hk_err_dec}
\end{equation}
where $u_{h,\theta}=\sum_{j=1}^{B}b_j(a)\,\Phi(|\mathbf{x}-\gamma\mathbf{x}_b^{(j)}|)$ is spanned by the analytic fundamental solutions and $u_{p,\theta}=\sum_{k=1}^{K_c}c_k(a)\,\varphi_\theta(|\mathbf{x}-\mathbf{t}_k|)$ is spanned by the low-rank correction kernels.
    Note that the decomposition of Eq.~\eqref{eq:hk_decomp} is not unique---$u_h$ and $u_p$ may differ by a $\mathcal{L}$-homogeneous solution; this degree of freedom is precisely absorbed by the homogeneous branch (it suffices to take $u_h$ to be the homogeneous part matching the boundary conditions), so that the error analysis below is insensitive to it.
    The error of the homogeneous branch is exactly the same as in the case of KernelOnet-PIKF in item~(2): since every fundamental solution satisfies $\mathcal{L}=0$ inside the domain, its approximation error is determined only by the boundary fitting residual and converges spectrally as $B$ increases.
    The error of the source branch, on the other hand, is determined by the "compressibility" of $u_p$ in the correction subspace $\mathcal{V}_{K_c}$: by the best approximation theorem,
\begin{equation}
\|u_{p}-u_{p,\theta}\|_{\Omega}\le \bigl(1+\Lambda\bigr)\inf_{v\in\mathcal{V}_{K_c}}\|u_{p}-v\|_{\Omega},
\label{eq:hk_src_bound}
\end{equation}
where $\Lambda$ is a stability constant related to sampling and optimization and the second factor on the right-hand side is the best approximation error of $u_p$ in $\mathcal{V}_{K_c}$. Therefore, as long as the nonlinear source term $u_p$ can be compressed efficiently in the $K_c$-dimensional subspace---for example when its energy is concentrated in a few modes---the error of the source branch decays rapidly with $K_c$; otherwise it decays slowly.

    Applying the linear principal part to Eq.~\eqref{eq:hk_expand} gives the PDE residual of the whole expansion. Since the homogeneous branch satisfies $\mathcal{L}$ exactly and contributes nothing, we obtain
\begin{equation}
\mathcal{R}_{\theta}(\mathbf{x})
=\mathcal{L}\mathcal{G}_\theta(\mathbf{x})+N[\mathcal{G}_\theta](\mathbf{x})
=\sum_{k=1}^{K_c}c_k(a)\,\mathcal{L}\varphi_\theta\!\left(\left|\mathbf{x}-\mathbf{t}_k\right|\right)+N[\mathcal{G}_\theta](\mathbf{x}).
\label{eq:hk_residual}
\end{equation}
When the $\mathcal{L}$-image of $\varphi_\theta$ is made to approximate a set of source-term basis functions $\chi_k$, the sum $\sum_k c_k\chi_k$ is the low-rank expansion of the source term $-N[u]$, and Eq.~\eqref{eq:hk_residual} is precisely the remainder of that expansion. The residual is therefore determined entirely by the approximation capability of the correction branch and is independent of the homogeneous branch, with a clear physical origin: the low-rank source-term subspace carries the error and can be reduced adaptively with $K_c$ and the compressibility of the source term , representing $-N[u]$ spatially adaptively as a function of $\mathbf{x}$.

    The $K_c$ in Eq.~\eqref{eq:hk_expand} is therefore an explicit "bias--capacity" knob: when $K_c=0$ we have $\mathcal{V}_{K_c}=\{0\}$, KernelOnet-HK degenerates to KernelOnet-PIKF and its error reduces to a purely homogeneous approximation error, which for nonlinear problems manifests itself as an irreducible model bias; as $K_c$ increases, the approximation capability of the source branch is enhanced and the approximation error of Eq.~\eqref{eq:hk_src_bound} decreases, but the number of coefficients and network parameters to be calibrated grows and the dependence on interior labels increases, so that the estimation error rises accordingly, and the model approaches the fully data-driven KernelOnet-RBF as $K_c\to B$ with the centres and basis-function form of the correction kernel agreeing with those of KernelOnet-RBF (an idealized limit). This is the theoretical basis on which KernelOnet-HK exchanges a controllable physical bias for the ability to handle nonlinearity: its accuracy ceiling is determined by the compressibility of the source term in a low-dimensional subspace.

Taken together, the errors of the three variants can be decomposed uniformly into an approximation error and an estimation error, where the approximation error is determined by the choice of the kernel space (for KernelOnet-HK by the direct sum $\mathcal{V}_{B}\oplus\mathcal{V}_{K_c}$) and the estimation error by the amount of training data and the optimization process.
    Table~\ref{tab:theory_compare} compares the three at the theoretical level in terms of the properties of the kernel space, the origin of the error and the convergence.
    It can be seen that the differences among the three essentially reduce to a compromise between "the agreement between the kernel space and the physical structure of the true solution" and "the flexibility of the approximation space": KernelOnet-PIKF exchanges a highly restricted but physically consistent kernel space for a minimal approximation error and unsupervised training;
    KernelOnet-RBF exchanges a flexible kernel space for adaptability to problems that differ from the fundamental solution;
    and KernelOnet-HK superimposes a low-rank source-term correction on the spectral approximation of the homogeneous branch, with $K_c$ as a knob providing a continuous transition between the two ends.

\begin{table}[htbp]
\centering
\small
\caption{Comparison of KernelOnet-RBF, KernelOnet-PIKF and KernelOnet-HK from the perspective of kernel function theory.}
\label{tab:theory_compare}
\setlength{\tabcolsep}{4pt}
\begin{tabular}{>{\raggedright\arraybackslash\hyphenpenalty=10000}p{2.5cm}>{\centering\arraybackslash}p{4.9cm}>{\centering\arraybackslash}p{5.3cm}>{\centering\arraybackslash}p{2.6cm}}
\toprule
 & \textbf{KernelOnet-PIKF} & \textbf{KernelOnet-HK} & \textbf{KernelOnet-RBF} \\
\midrule
Kernel space & An approximation subspace of the homogeneous solution space (spanned by fundamental solutions) & $\mathcal{V}_{B}\oplus\mathcal{V}_{K_c}$: fundamental-solution subspace $+$ low-rank correction subspace & Neural-network-parameterized radial basis function \\
Approximation error & Can be made arbitrarily small; the space is dense in the solution space & Homogeneous branch minimal (spectral convergence) $+$ source branch decaying with $K_c$ according to compressibility & Can be large; depends on the learned kernel shape \\
Estimation error & Boundary fitting residual only & Boundary fitting of the homogeneous branch $+$ interior/residual fitting of the correction branch & Data fitting $+$ kernel shape learning \\
Convergence & Spectral or nearly exponential & Spectral for the homogeneous branch; total error improves with $K_c$ and the compressibility of the source term & Spectral or algebraic, depending on kernel smoothness \\
\bottomrule
\end{tabular}
\end{table}

It should be noted that the above analysis provides, at the level of approximation theory, a "qualitative or semi-quantitative description of the convergence order and the origin of the error"; the constants and the specific convergence rates also depend on practical factors such as the smoothness of the problem solution, the regularization parameters of the kernel function, the capacity of the branch network and the degree of convergence of the optimization, which are usually difficult to identify analytically in advance. This is exactly why Section~3 systematically verifies the accuracy of each method through numerical experiments.
    In addition, universal approximation at the operator level---that is, the ability of the branch network to approximate the coefficient functions---is guaranteed by the classical universal approximation theorem for neural networks \cite{chen1995universal,park1991universal}, so that KernelOnet possesses universal approximation capability for the target operator when the number of kernel functions $B\to\infty$ and the capacity of the branch network is sufficient (for KernelOnet-PIKF this conclusion presupposes that the target solution belongs to, or can be approximated by, the homogeneous solution space);
    the actual performance for finite $B$ and finite data is then determined jointly by the approximation error and the estimation error discussed above.

\section{Numerical examples and discussions}

In this section, we evaluate the performance of the proposed KernelOnet framework through four representative numerical examples: the first three are benchmark cases that verify the accuracy, interpretability, and physical consistency of the framework on idealized mathematical problems from different perspectives, while the last one is an engineering-oriented example of underwater acoustic radiation in a shallow-water waveguide that tests the framework in a realistic complex physical scenario:
(1) the Laplace equation on a circular domain, which is used to examine whether the data-driven kernel can recover the fundamental solution from data when the expansion strictly satisfies the governing equation;
(2) the nonlinear modified Helmholtz equation on a star-shaped domain, which is used to examine the applicability of the hybrid kernel when no analytic fundamental solution exists;
(3) the complex Helmholtz equation in an unbounded exterior domain, which is used to examine how the analytic fundamental-solution kernel handles traveling waves and the radiation condition, and to compare two implementations, namely boundary collocation and boundary integral;
(4) underwater acoustic radiation and propagation induced by spherical-shell vibration in a shallow-water waveguide, which tests the ability of the framework to solve problems in a realistic complex physical scenario.
The model performance is measured by the mean of the sample-level relative $L_2$ error:
\begin{equation}
\mathcal{E}=\frac{1}{N}\sum_{i=1}^{N}\frac{\sqrt{\sum_{k=1}^{n_t}\left(u_{\mathrm{pred},k}^{(i)}-u_{\mathrm{true},k}^{(i)}\right)^{2}}}{\sqrt{\sum_{k=1}^{n_t}\left(u_{\mathrm{true},k}^{(i)}\right)^{2}}},
\end{equation}
where $u_{\mathrm{pred},k}^{(i)}$ and $u_{\mathrm{true},k}^{(i)}$ are the predicted and reference values, respectively, of the $i$-th sample at the $k$-th interior evaluation point; for the complex-valued problems of Case~3 and Case~4, the error is computed separately for the real and imaginary parts and then averaged.

All experiments were carried out on a Linux server configured with an AMD Ryzen 9 9950X3D processor, 96 GB of memory, an NVIDIA GeForce RTX 5090 D v2 graphics card with 24 GB of GPU memory, and the Ubuntu 24.04 operating system. All models are implemented in PyTorch: the branch network of every method is a three-layer fully connected $\mathrm{BranchNet}[160,160,160]$ with $\tanh$ activation, trained with the Adam optimizer \cite{kingma2015adam} at a learning rate of $10^{-4}$. The input dimension of the branch network equals the number of boundary points $n_b$ of each case: 160 for Case~1 and Case~3, 200 for Case~2, and 50 for Case~4; for Case~4, the complex acoustic pressure is encoded separately in terms of its real and imaginary parts, so the actual input dimension is $2\times50=100$. The training and test sets of all four cases contain 2000 samples, and the number of training epochs is uniformly set to $5\times10^5$.

The construction of the trunk network of each method is detailed in Section~2.4: DeepONet uses a standard fully connected trunk $\mathrm{TrunkNet}[160,160,160]$; KernelOnet-RBF uses a data-driven radial basis function network $\mathrm{RBFTrunk}[160,160]$; KernelOnet-PIKF uses an analytic fundamental-solution kernel trunk, whose collocation form is the $\gamma$-shifted fundamental solution and whose boundary integral form is the single-layer potential kernel on the true boundary; and KernelOnet-HK appends a low-rank radial correction branch to the analytic fundamental-solution branch, this shallow network taking the radial distance as input, with hidden layers $[160,160]$ and $\tanh$ activation, and with fixed sampling points as the correction kernel centers.

As for data generation, in Case~1 and Case~3 the boundary conditions are sampled from a Gaussian random field and the reference solutions are generated by the method of fundamental solutions, while in Case~2 the boundary conditions are sampled from a Gaussian random field and the reference solution is generated by the finite element method; in Case~4 the acoustic pressure on the shell surface is synthesized by randomly superposing several vibration modes, and the reference solutions are generated separately for the near and far fields, with the Pekeris waveguide Green's function taken as the kernel in the near field and the normal-mode Green's function as the kernel in the far field, the fields being synthesized after the virtual source strengths are determined by least squares from the Dirichlet data on the shell surface, and with a separate dataset generated for each of the three sound speed profiles in the far field. It should be noted that not all methods require the interior reference solutions provided by the training set: the unsupervised configurations (PI-DeepONet and KernelOnet-PIKF) take only the boundary data from it, and their training involves no interior solution labels at all. The reference solutions generated in the dataset are therefore used only to compute the test error and to train the three supervised baselines, namely DeepONet, KernelOnet-RBF, and KernelOnet-HK; Case~3 and Case~4 adopt unsupervised configurations throughout, and the datasets of these two cases serve only for error evaluation and take no part in training. As for the choice of kernel form, Case~1 uses KernelOnet-PIKF and KernelOnet-HK, whose analytic fundamental-solution branches are both of collocation form; Case~2 uses KernelOnet-HK, whose homogeneous branch is likewise an analytic fundamental solution of collocation form; Case~3 compares both the collocation and the boundary integral forms together with their SVD-truncated variants; and Case~4 uses the collocation form with SVD truncation applied. In addition, Case~3 and Case~4 are both unbounded exterior problems, for which only KernelOnet-PIKF, with the analytic fundamental solution as its kernel, is directly applicable, whereas DeepONet and KernelOnet-RBF do not apply and KernelOnet-HK would also introduce an additional estimation error; both cases are therefore solved with KernelOnet-PIKF. More detailed computational costs, accuracy comparisons, and additional analyses are given in Appendix~\ref{app:supp}.

\subsection{Case~1: Laplace equation on a circular domain}

The first example considers the two-dimensional Laplace equation on a circular domain of radius $R=0.5$:
\begin{equation}
\nabla^2 u(x,y)=0,\quad (x,y)\in\Omega=\{x^2+y^2<0.25\};\qquad u=g,\quad \mathbf{x}\in\Gamma=\partial\Omega,
\end{equation}

Table~\ref{tab:case1_results} summarizes the training results of the five methods on Case~1, including the number of learnable parameters, the relative $L_2$ error, and the average running time per 100 training epochs.
Among them, neither KernelOnet-PIKF nor PI-DeepONet uses interior solution labels: the former takes the boundary residual alone as its loss, whereas the latter takes the boundary residual together with the PDE residual at interior collocation points; all the remaining methods are trained in a supervised manner using interior solution labels.
It can be seen that KernelOnet-PIKF attains the highest accuracy among the five methods with $8.89\times10^{-4}$, and has the fewest total learnable parameters, only 103,041, of which the kernel parameter is a single scalar $\gamma$.
The supervised KernelOnet-RBF ($1.29\times10^{-3}$) also outperforms DeepONet ($1.89\times10^{-3}$) while using fewer parameters.
It is worth noting that this case is a linear homogeneous problem, so the low-rank correction branch of HK is not theoretically necessary: with the correction kernel centers taken by in-domain LHS, the trained correction branch is almost zero, its $L_2$ energy ratio to the homogeneous branch being only about $7\times10^{-6}$. At this point HK and KernelOnet-PIKF are virtually indistinguishable in terms of function space, and the only substantial difference between them lies in the loss function---the loss of HK contains only the interior data term and no boundary residual, so it does not directly constrain the boundary condition; its accuracy is therefore $2.04\times10^{-3}$, lower than the $8.89\times10^{-4}$ of KernelOnet-PIKF, which fits the boundary residual directly, and also slightly lower than the $1.29\times10^{-3}$ of KernelOnet-RBF, which adaptively learns the kernel shape from data; the gap thus arises from the type of loss rather than from the correction branch itself.
PI-DeepONet converges to $3.53\times10^{-3}$, an accuracy about $1/4$ that of KernelOnet-PIKF, and takes 0.53 s per 100 epochs, about 9 times the 0.06 s of the latter.


\begin{table}[htbp]
\centering
\caption{Case~1: comparison of the training results of DeepONet, PI-DeepONet, KernelOnet-RBF, KernelOnet-PIKF, and KernelOnet-HK.}
\label{tab:case1_results}
\begin{tabular}{lccc}
\toprule
Method & Learnable parameters & Relative $L_2$ & Time per 100 epochs (s) \\
\midrule
DeepONet & 180,801 & $1.89\times 10^{-3}$ & 0.08 \\
PI-DeepONet & 180,801 & $3.53\times 10^{-3}$ & 0.53 \\
KernelOnet-RBF & 129,281 & $1.29\times 10^{-3}$ & 0.43 \\
KernelOnet-PIKF & 103,041 & $8.89\times 10^{-4}$ & 0.06 \\
KernelOnet-HK & 132,503 & $2.04\times 10^{-3}$ & 0.10 \\
\bottomrule
\end{tabular}
\end{table}

\begin{figure}[htbp]
\centering
\includegraphics[width=\textwidth]{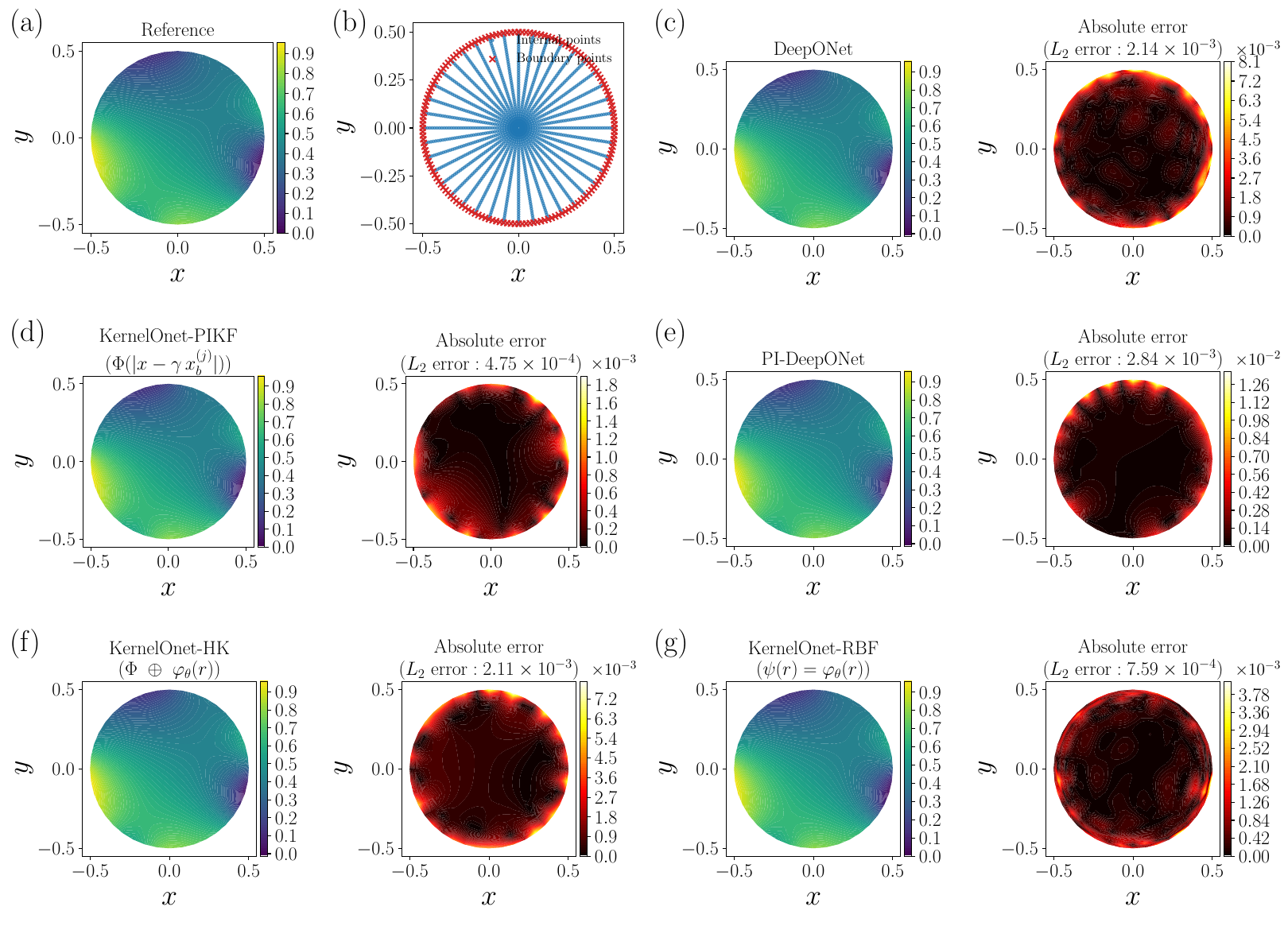}
\caption{Case~1: predicted solutions and absolute error fields for a representative test sample. (a) MFS reference solution. (b) Distribution of the boundary points (red crosses, used to encode the Dirichlet data and generate the kernel centers) and the interior evaluation points (blue dots). (c)--(g) Predicted solutions and absolute errors of DeepONet, KernelOnet-PIKF, PI-DeepONet, KernelOnet-HK, and KernelOnet-RBF, with the corresponding relative $L_2$ errors annotated. Note that KernelOnet-PIKF and PI-DeepONet did not use any interior solution labels during training.}
\label{fig:case1_fields}
\end{figure}

\begin{figure}[htbp]
\centering
\includegraphics[width=\textwidth]{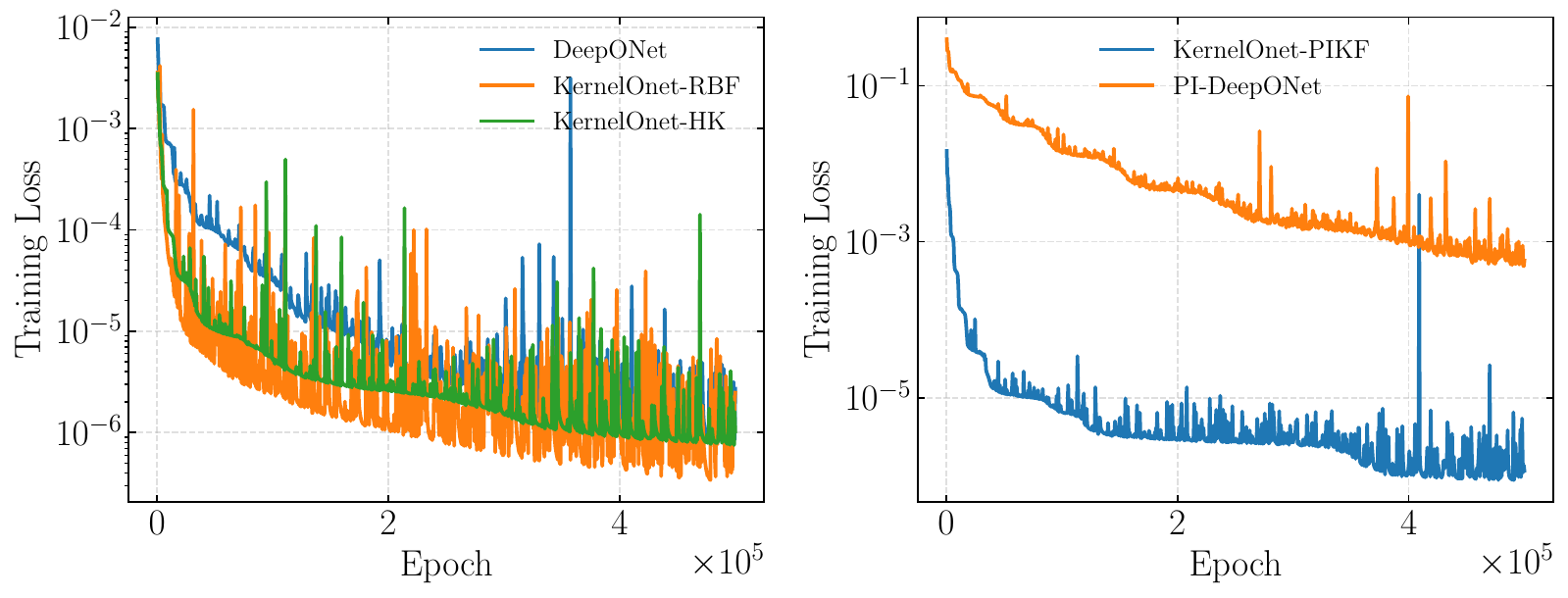}
\caption{Case~1: convergence history of the training loss. The left panel shows the supervised methods (DeepONet, KernelOnet-RBF, KernelOnet-HK), whose loss is the mean squared error relative to the interior reference solution; the right panel shows the unsupervised methods (KernelOnet-PIKF and PI-DeepONet), whose losses are the boundary residual and the weighted sum of the PDE residual and the boundary residual, respectively.}
\label{fig:case1_loss}
\end{figure}

Figure~\ref{fig:case1_fields} shows the predicted solutions and absolute error fields of the five methods on the same randomly selected test sample, where Figure~\ref{fig:case1_fields}(a) is the MFS reference solution and Figure~\ref{fig:case1_fields}(b) displays the distribution of the training collocation points: the boundary points (red crosses) are uniformly distributed on the circumference, serving both to encode the Dirichlet boundary condition and, through the radial scaling $\gamma\,\mathbf{x}_b^{(j)}$, to generate the kernel centers;
    the interior evaluation points (blue dots) are arranged in the radial-angular direction and are naturally denser near the center of the circle.
In terms of the predicted fields, all five methods qualitatively reproduce the overall morphology of the reference solution, but the differences among the absolute error fields lie mainly in their magnitudes: apart from PI-DeepONet, whose error is spread widely over the domain, the errors of the remaining methods are all concentrated near the boundary.
The error of DeepONet (Figure~\ref{fig:case1_fields}(c), $L_2=2.14\times10^{-3}$) is concentrated in a ring near the boundary, indicating that a free trunk network without physical priors struggles to capture the rapid variation of the solution in the vicinity of the boundary.
KernelOnet-PIKF (Figure~\ref{fig:case1_fields}(d), $L_2=4.75\times10^{-4}$) achieves the smallest error, which is strictly confined to a thin boundary layer and is almost zero in the interior region.
    Since the PDE residual is identically zero, the error is a harmonic function whose amplitude is controlled by the boundary residual according to the maximum principle; the boundary data of this problem contain high-frequency components, whose harmonic extension decays rapidly toward the interior, so the error exhibits a thin boundary-layer structure.
PI-DeepONet (Figure~\ref{fig:case1_fields}(e), $L_2=2.84\times10^{-3}$) is comparable in magnitude to DeepONet, and its error is more widely distributed over the domain.
    The difference between the two lies in the way physical information is embedded: PI-DeepONet penalizes the PDE residual at interior collocation points as a soft constraint, whereas KernelOnet-PIKF hard-codes the physical information into the kernel structure, the consequences of which will be further reflected in the loss convergence in Figure~\ref{fig:case1_loss}.
The errors of KernelOnet-HK (Figure~\ref{fig:case1_fields}(f), $L_2=2.11\times10^{-3}$) and KernelOnet-RBF (Figure~\ref{fig:case1_fields}(g), $L_2=7.59\times10^{-4}$) are likewise mainly distributed near the boundary, with magnitudes entering the range $10^{-3}\sim10^{-4}$;
    the error structure of HK is similar to that of KernelOnet-PIKF but somewhat larger in magnitude, for the reason given above.

Figure~\ref{fig:case1_loss} shows the convergence history of the training loss. Since the loss definitions of the supervised and unsupervised groups of methods differ (the former being the mean squared error of the interior solution and the latter the boundary residual), they cannot be compared directly and are therefore plotted in two separate subfigures.
Figure~\ref{fig:case1_loss} (left) compares the three supervised methods: the losses of all three keep decreasing during training and eventually fall to about the $1\times10^{-6}$ level---with final-stage averages of $1.3\times10^{-6}$ (KernelOnet-HK), $1.7\times10^{-6}$ (KernelOnet-RBF), and $1.9\times10^{-6}$ (DeepONet), respectively.
    This indicates that with sufficient training all three supervised methods can suppress the interior fitting error to a very low level, and that the differences in their final prediction accuracy (Table~\ref{tab:case1_results}) stem mainly from their ability to resolve the error in the boundary neighborhood rather than from the loss values themselves.
Figure~\ref{fig:case1_loss} (right) compares the two unsupervised methods.
    The boundary residual of KernelOnet-PIKF decreases monotonically from about $1.5\times10^{-2}$ to $1.2\times10^{-6}$ and converges smoothly, corresponding to the highest prediction accuracy in Table~\ref{tab:case1_results};
    the weighted loss of PI-DeepONet decreases from $3.9\times10^{-1}$ to about $5.7\times10^{-4}$ (of which the PDE residual is about $1.9\times10^{-4}$ and the boundary residual about $3.8\times10^{-5}$); although it does not stagnate, its final accuracy is still lower than that of KernelOnet-PIKF.
    The reason is that PI-DeepONet must simultaneously optimize two competing objectives, the PDE residual and the boundary residual, and the balance between the weights $\lambda_{\mathrm{PDE}}$ and $\lambda_{\mathrm{BC}}$ requires manual tuning;
    whereas for KernelOnet-PIKF the PDE residual is identically zero by virtue of the kernel structure, so the optimization objective degenerates into a single boundary fitting problem, formally equivalent to a well-conditioned least-squares problem that requires no weight tuning and suffers no gradient conflict, and therefore converges rapidly and smoothly.
    This is mutually corroborated by the fact in Table~\ref{tab:case1_results} that PI-DeepONet takes about nine times as long per epoch as KernelOnet-PIKF: the hard constraint is superior to the soft constraint in accuracy, stability, and efficiency simultaneously.

\begin{figure}[htbp]
\centering
\includegraphics[width=\textwidth]{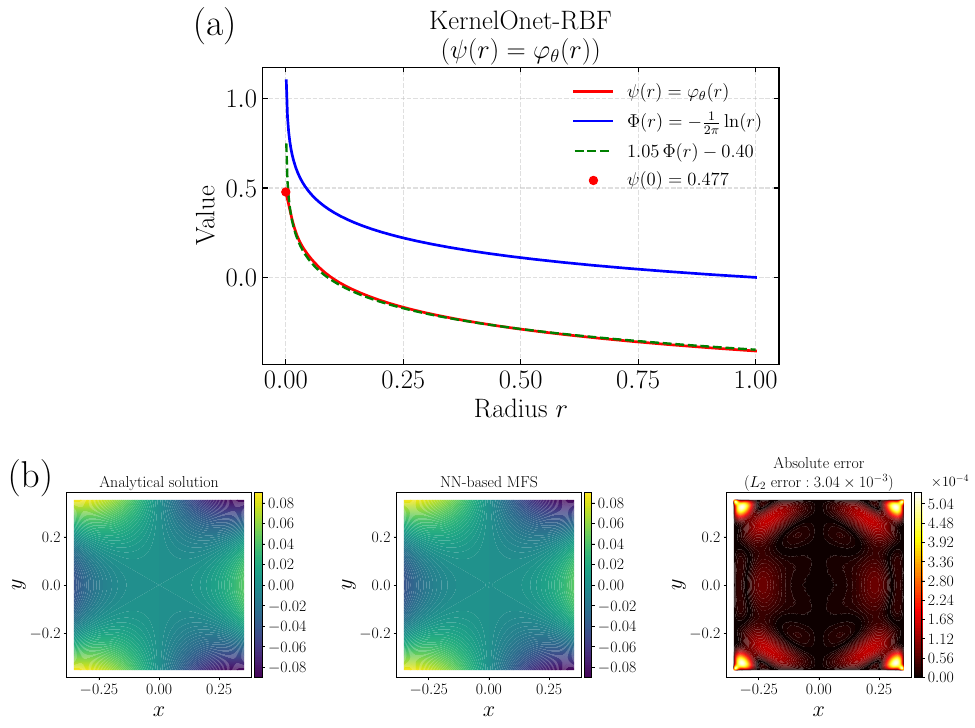}
\caption{Case~1: kernel diagnostics and functional verification of the learned kernel. (a) The data-driven kernel $\psi(r)=\varphi_\theta(r)$ of KernelOnet-RBF (red), which is well described by the affine fit $1.05\,\Phi(r)-0.40$ (green dashed line), i.e., the network recovers the fundamental solution $\Phi(r)=-\frac{1}{2\pi}\ln r$ up to a scale and an offset, while remaining finite at the origin with $\psi(0)=0.477$. (b) The NN-based MFS obtained by taking this learned kernel as the fundamental solution and collocating on the boundary of the inscribed square: left, the analytical solution $u^*=x^3-3xy^2$; middle, the predicted solution; right, the absolute error field (the relative $L_2$ error is given in the title); the source points are $160$ boundary points on the circle of radius $0.5$, and the solution domain is the square inscribed in this circle.}
\label{fig:case1_kernel_mfs}
\end{figure}

Unlike the fully connected trunk of DeepONet, the kernel of KernelOnet is a univariate radial function that can be plotted directly and compared pointwise with the analytic fundamental solution, thereby allowing one to "read out" the physics learned by the network. Figure~\ref{fig:case1_kernel_mfs}(a) is the kernel $\psi(r)=\varphi_\theta(r)$ learned by KernelOnet-RBF after training (red solid line).
    Although this kernel is learned entirely from data without being told anything about the Laplace operator during training, its shape is highly similar to that of the analytic fundamental solution $\Phi(r)=-\frac{1}{2\pi}\ln r$ (blue solid line): it likewise rises steeply as $r\to0$ and decays monotonically as $r$ increases.
    Quantitatively, fitting it with a two-parameter affine transformation yields $\psi(r)\approx1.05\,\Phi(r)-0.40$ (green dashed line), and this fitted curve almost completely coincides with the learned kernel.
    This result shows that the network autonomously "discovers" the fundamental solution of the Laplace operator from the data, differing only by a scale factor and a constant shift---the scale factor can be exactly absorbed by the coefficients $b_j$ output by the branch network, while the constant shift corresponds to a common component of the coefficients and can likewise be adjusted by the network, so neither affects the expressive capacity.
    The only essential difference between the two lies in the behavior at the origin: the analytic fundamental solution diverges at $r=0$, whereas the learned kernel takes the finite value $\psi(0)=0.477$, meaning that the network automatically learns a regularized, non-singular fundamental solution.
    This provides direct numerical evidence for the claim in Section~2.4 that "in constant-coefficient linear problems, the learned kernel can be viewed as a regularized approximation of the fundamental solution," and also explains why KernelOnet-RBF can achieve higher accuracy with significantly fewer parameters than DeepONet: its trunk is constrained to a rotation-invariant function that depends only on the radial distance, so the search space is greatly reduced compared with that of a free trunk, while the function family containing the true solution remains within it.

To further verify this claim from a functional point of view, we directly freeze the trained network and perform a boundary-type solve with this kernel as the fundamental solution: the source points are taken as the boundary points of Case~1 ($160$ points on the circle of radius $0.5$), the solution domain is the square inscribed in this circle, the target solution is the harmonic function $u^*=x^3-3xy^2$, and the coefficients are obtained by least squares from collocation on the boundary of the square according to $u(\mathbf{x})=\sum_j c_j\psi(|\mathbf{x}-\mathbf{s}_j|)$. Figure~\ref{fig:case1_kernel_mfs}(b) gives the result of this solve, whose left, middle, and right panels are the analytical solution, the NN-based MFS solution, and the absolute error field, respectively: the prediction is visually almost indistinguishable from the analytical solution, the relative $L_2$ error on the grid is $3.04\times10^{-3}$ (maximum absolute error $5.3\times10^{-4}$), and the error is concentrated near the four vertices that coincide with the source points, i.e., exactly where the learned kernel deviates from the fundamental solution. It can be seen that the learned kernel not only approximates the fundamental solution in shape but can also be used directly for boundary-type solving just like the analytic fundamental solution.

\subsection{\texorpdfstring{Case~2: nonlinear modified Helmholtz equation on a star-shaped domain}{Case~2: nonlinear modified Helmholtz equation on a star-shaped domain}}
The previous example considered a linear problem on a simple geometry. As a further test of the applicability of the framework, this example combines "complex geometry" with "nonlinearity": we solve the modified Helmholtz equation with a cubic nonlinear term on a five-pointed star domain
\begin{equation}
\nabla^{2}u-k^{2}u+\varepsilon u^{3}=0,\quad \mathbf{x}\in\Omega;\qquad u=g,\quad \mathbf{x}\in\Gamma,
\label{eq:case2_pde}
\end{equation}
where $\Omega=\{\rho R(\theta)(\cos\theta,\sin\theta):\rho\in[0,1]\}$ and $R(\theta)=1+0.2\cos(5\theta)$, so that the boundary $\Gamma$ is a $C^{\infty}$ curve containing no corners; we fix $k=2$ and sweep $\varepsilon=0.1,0.5,1,2,3,4$ to examine the influence of the nonlinearity strength, with $\varepsilon=4$ taken as the main case. The cubic nonlinearity is chosen both to distinguish the problem from the linear one and to make the nonlinear term depend only on $u$ itself. Here the nonlinear and linear terms are of the same order: writing $\rho_{\mathrm{nl}}=\varepsilon u^{2}/k^{2}$, its typical value over $u\in[0,1]$ is about $0.5$, reaching $1.0$ when $u=1$; in the main case $\varepsilon=4=k^{2}$, the equation can be rewritten as $\nabla^{2}u=k^{2}u\,(1-u^{2})$, and the solution tends to the saturation value $u\to1$ in the inner region, which is a rather strong nonlinearity. The reference solution is generated by the finite element method implemented in \texttt{scikit-fem} together with Newton iteration; its discretization accuracy is verified against a manufactured solution and lies far below the operator accuracy being compared.

\begin{figure}[htbp]
\centering
\includegraphics[width=0.62\textwidth]{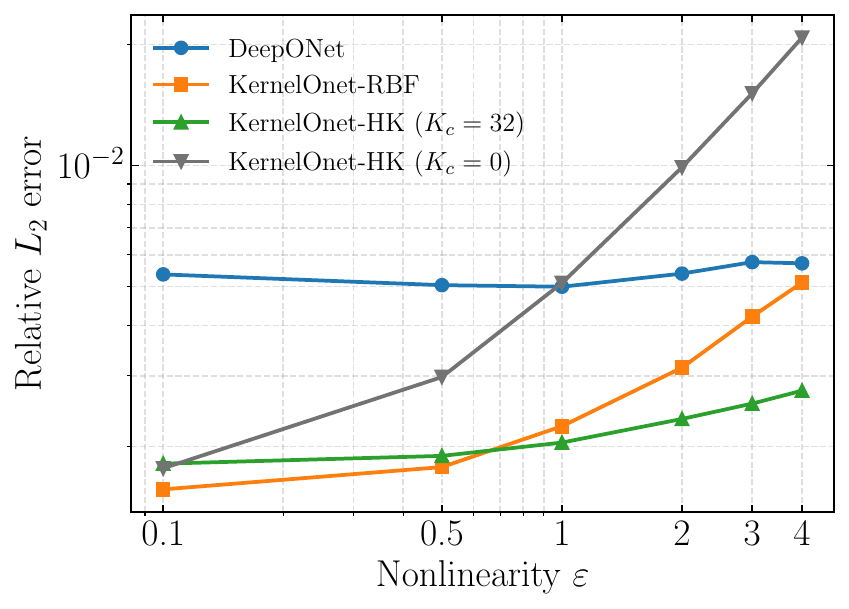}
\caption{Case~2: relative $L_2$ error as a function of the nonlinearity strength $\varepsilon$. HK with the correction branch switched off ($K_c=0$) deteriorates rapidly with $\varepsilon$, followed by KernelOnet-RBF, whereas HK with the low-rank correction branch retained hardly degrades and surpasses RBF for $\varepsilon\ge1$.}
\label{fig:case2_eps}
\end{figure}

\begin{figure}[htbp]
\centering
\includegraphics[width=\textwidth]{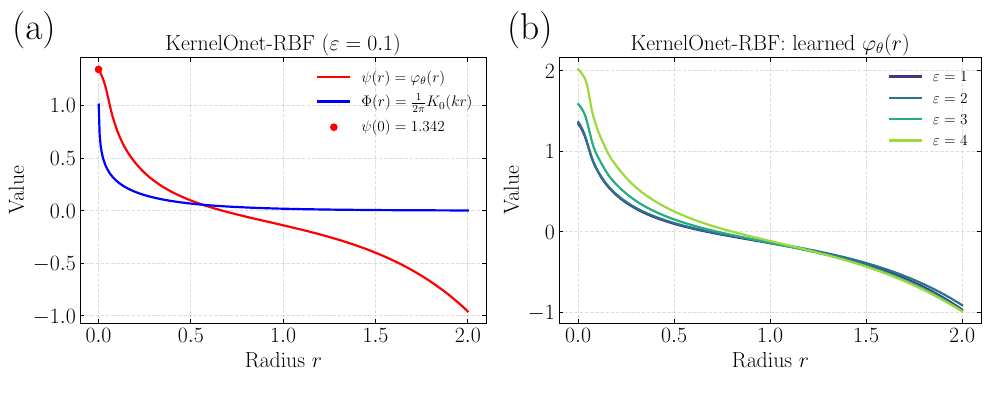}
\caption{Case~2: (a) comparison of the kernel $\psi(r)=\varphi_\theta(r)$ learned by KernelOnet-RBF (red) with the analytic fundamental solution $\Phi(r)=\frac{1}{2\pi}K_0(kr)$ (blue) at $\varepsilon=0.1$, where the red dot is $\psi(0)$; (b) the learned kernels $\varphi_\theta(r)$ at $\varepsilon=1,2,3,4$, whose shapes are similar while their amplitudes increase with the nonlinearity strength.}
\label{fig:case2_star_hk}
\end{figure}

\begin{figure}[htbp]
\centering
\includegraphics[width=\textwidth]{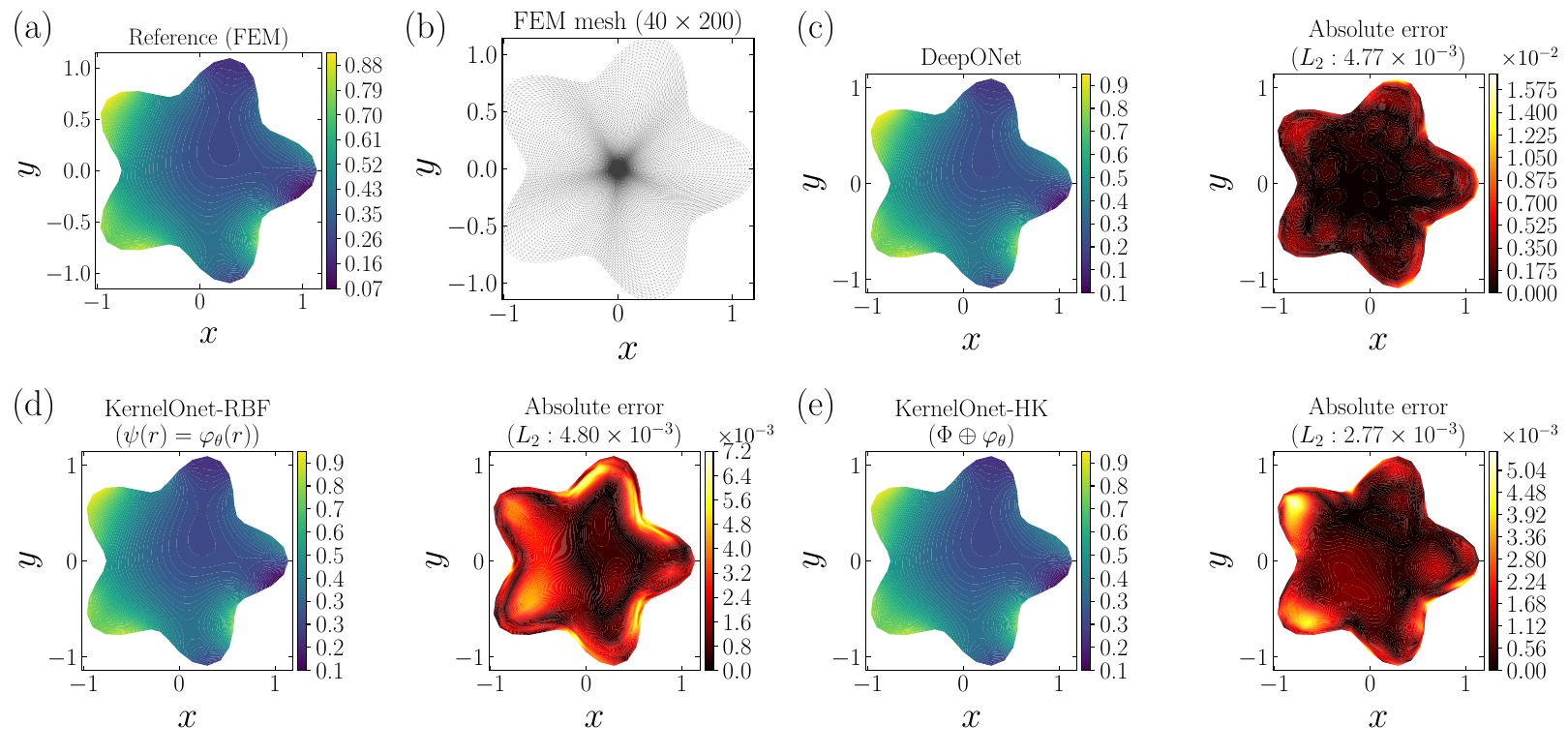}
\caption{Case~2: (a) FEM reference solution; (b) finite element mesh ($40\times200$, central fan triangulation); (c)--(e) predicted solutions (left) and absolute error fields (right) of DeepONet, KernelOnet-RBF, and KernelOnet-HK, where the titles of the error plots give the relative $L_2$ error for this sample.}
\label{fig:case2_star_fields}
\end{figure}

Figure~\ref{fig:case2_eps} gives the variation of the relative $L_2$ error of each method with the nonlinearity strength $\varepsilon$, where "HK ($K_c=0$)" denotes the baseline with the low-rank correction branch switched off and only the analytic fundamental-solution branch retained; to use interior solution labels as the other methods do, this baseline is trained in a supervised manner, and its error is precisely the "irreducible model bias" described in Section~2.5. As $\varepsilon$ increases, HK without the correction branch ($K_c=0$) deteriorates rapidly, its error growing from $1.77\times10^{-3}$ to $2.08\times10^{-2}$ ($\varepsilon=4$, about $12$ times); the fully data-driven KernelOnet-RBF also degrades markedly ($1.57\to5.11\times10^{-3}$, about $3.3$ times); whereas KernelOnet-HK, which retains the low-rank correction branch, increases only slowly from $1.81\times10^{-3}$ to $2.75\times10^{-3}$ (about $1.5$ times). Fitting the error as an approximate power law in $\varepsilon$ ($\mathrm{err}\sim\varepsilon^{p}$) gives $p\approx0.68$ ($K_c=0$), $0.32$ (RBF), and $0.11$ (HK), i.e., the error of HK grows most slowly with the nonlinearity. Correspondingly, the advantage of HK over RBF widens monotonically with $\varepsilon$: RBF is slightly better for $\varepsilon\le0.5$, the two are close at $\varepsilon=1$, after which HK overtakes it and pulls ahead, and at $\varepsilon=4$ HK is about $1.9$ times better than RBF and about $7.6$ times better than $K_c=0$; by contrast, the error of DeepONet stays near $5\times10^{-3}$ throughout the sweep and does not improve with $\varepsilon$. The mechanism behind this trend is that the analytic fundamental-solution branch $\Phi=K_0(kr)/(2\pi)$ provides the correct structure of the homogeneous field and the near-boundary behavior for free, so the network only needs to learn a low-dimensional source-term correction, whereas RBF must learn both the homogeneous field and the nonlinear source term from data and struggles increasingly as the nonlinearity grows. The amplitude of the correction branch also grows in step with $\varepsilon$, quantitatively confirming that it carries the nonlinear component: the fraction of the prediction's $L_2$ energy that it accounts for increases from about $0.3\%$ at $\varepsilon=0.1$ to about $26.6\%$ at $\varepsilon=4$, and the ratio of coefficient norms $\|c\|/\|b\|$ increases from $0.11$ to $0.20$; both consistently indicate that the correction branch accommodates the anharmonic component introduced by the cubic nonlinear term $\varepsilon u^{3}$. It should be pointed out that $\varepsilon\ge2$ already exceeds the sufficient uniqueness condition given by the maximum principle ($\varepsilon<k^{2}/(3u_{\max}^{2})\approx1.33$); however, using Newton iteration from two different initial guesses, we obtained the same solution for $\varepsilon=2,3,4$ (maximum difference $\le10^{-13}$), indicating that the solution is unique and can be solved robustly over the range considered.

Figure~\ref{fig:case2_star_hk} examines the relationship between the learned kernel and the fundamental solution of this linear principal part. Figure~\ref{fig:case2_star_hk}(a) compares, at $\varepsilon=0.1$ (weak nonlinearity), the kernel $\psi(r)=\varphi_\theta(r)$ learned by KernelOnet-RBF pointwise with the fundamental solution $\Phi(r)=\frac{1}{2\pi}K_0(kr)$ of the modified Helmholtz equation: the two are qualitatively similar only near the origin---$\psi$ takes a finite value at the origin ($\psi(0)=1.342$, whereas $\Phi$ diverges there) and decreases monotonically with $r$; but as $r$ increases the two separate noticeably, $\psi$ decaying more slowly than $\Phi$, crossing zero at $r\approx0.6$ and becoming negative, reaching about $-0.96$ at $r=2$, where $r=2$ already exceeds the maximum radius of the star-shaped domain and corresponds to the value of the kernel function itself, whereas $\Phi$ is always positive and tends rapidly to zero as $r$ increases. It can thus be seen that in this problem the learned kernel does not coincide with an affine transformation of $\Phi$, in contrast to Case~1 (the Laplace equation), where the kernel almost recovers an affine image of the fundamental solution: for problems with a nonlinear source term, the single data-driven radial kernel carries information about both the homogeneous field and the source term in its fitting. This contrast also delineates the scope of applicability of the statement that "the learned kernel is the fundamental solution": it stems from the property that in constant-coefficient linear problems the solution can be spanned by an expansion in fundamental solutions, and is therefore a conclusion for linear problems; for nonlinear problems, no usable fundamental solution exists for the equation, and the learned kernel can no longer be interpreted as a fundamental solution, see Section~2.4.1. Figure~\ref{fig:case2_star_hk}(b) further gives the learned kernels at $\varepsilon=1,2,3,4$: the four have similar shapes, but their amplitudes increase systematically with the nonlinearity strength, $\psi(0)$ being about $1.34$ at $\varepsilon=1$ and increasing to $2.01$ at $\varepsilon=4$, i.e., the network adapts the scale of the kernel to cope with a stronger nonlinear source term. The network evaluation part of HK only needs to be computed for $K_c=32$ correction centers, whereas RBF requires computation over all $B=200$ centers, which is a direct manifestation of the cost difference between the two.

Figure~\ref{fig:case2_star_fields} gives the field distribution of the main case ($k=2$, $\varepsilon=4$) on a representative test sample: Figure~\ref{fig:case2_star_fields}(a) is the FEM reference solution, (b) is the finite element mesh ($40\times200$, central fan triangulation), and (c)--(e) are in turn the predicted solutions (left) and absolute error fields (right) of DeepONet, KernelOnet-RBF, and KernelOnet-HK. All three methods reproduce the overall morphology of the reference solution, and the differences in accuracy are mainly reflected in the resolution of the boundary neighborhood: the absolute error of DeepONet is distributed fairly uniformly over the domain, with a magnitude comparable to that of RBF ($L_2=4.77\times10^{-3}$ on this sample), indicating that a fully connected trunk lacking physical priors struggles to capture the rapid variation near the boundary caused by the strong nonlinearity; KernelOnet-RBF has $L_2=4.80\times10^{-3}$ on this sample, but its error is clearly concentrated near the five convex extrema of the star-shaped domain and is more confined to the thin boundary layer; the error field of KernelOnet-HK has the lowest overall magnitude ($2.77\times10^{-3}$), and the peaks at the five convex extrema are also lower than those of RBF, being more uniformly distributed over the whole domain. In the test-set average sense, the relative $L_2$ error of HK is $2.75\times10^{-3}$, better than $5.11\times10^{-3}$ for RBF and $5.71\times10^{-3}$ for DeepONet.


\subsection{Case~3: complex Helmholtz equation in an unbounded exterior domain}

The third case study considers the two-dimensional complex Helmholtz equation in an unbounded exterior domain, which is used to test the ability of KernelOnet to solve problems on infinite domains and to capture traveling-wave propagation:
\begin{equation}
\nabla^2 u + k^2 u = 0,\quad \mathbf{x}\in\Omega=\{\mathbf{x}\in\mathbb{R}^2:\ |\mathbf{x}|>R\};\qquad u=g,\quad \mathbf{x}\in\Gamma=\partial\Omega,
\end{equation}
where $R=0.5$, the wavenumber $k$ is adjustable, and the boundary condition is of Dirichlet type; the solution $u(\mathbf{x})=u^{\mathrm{re}}(\mathbf{x})+i\,u^{\mathrm{im}}(\mathbf{x})$ is complex-valued and satisfies the Sommerfeld radiation condition at infinity (i.e., only outward-propagating waves exist). This section scans $k\in\{5,10,15,20\}$ and takes $k=20$ as the main case of interest. Here only KernelOnet-PIKF, whose kernel is the analytic fundamental solution, and its boundary integral form KernelOnet-PIKF-SL are applicable: its kernel $G(r)=\frac{i}{4}H_0^{(1)}(kr)$ inherently satisfies the Helmholtz equation and the Sommerfeld radiation condition, whereas DeepONet, PI-DeepONet, and KernelOnet-RBF all lack a mechanism to enforce the radiation condition (see Section~2.4.2).

Table~\ref{tab:case3_results} gives the average relative $L_2$ error over $k\in\{5,10,15,20\}$ for the two kernel-basis forms, collocation and boundary integral, together with their SVD-truncated variants. All four configurations are trained without supervision, with only the boundary residual imposed: the collocation form moves the source points radially inward, i.e., $x_s=\gamma\,x_b$ with a learnable $\gamma<1$; the boundary integral form adopts the single-layer potential kernel basis of Section~2.4.2; and the SVD variants apply the orthogonal kernel-basis preprocessing of Appendix~\ref{app:svd} to the boundary kernel matrix. Without truncation, the error of the collocation form varies slowly with the wavenumber at the $10^{-3}$ level, whereas the boundary integral form degrades markedly as the wavenumber increases; after SVD truncation, not only does the magnitude of both drop across the board, but their growth with the wavenumber also becomes the gentlest, indicating that the truncated orthogonal kernel basis is more robust as the wavenumber rises. The truncation order is selected offline according to the reconstructability of the boundary data: for the collocation kernel, $q=40$ suffices at all wavenumbers, whereas the singular values of the single-layer potential kernel decay slowly and the required order grows with the wavenumber, so that if $q=40$ were taken uniformly, the integral form would truncate away the high-order modes carrying the boundary data for $k\ge15$, degrading the error to the $10^{-1}$ level. This shows that SVD preprocessing is exactly the key to eliminating the numerical ill-conditioning of the single-layer potential kernel and bringing the integral form to the same accuracy level as the collocation form. After training, the source-point scaling coefficient of the collocation form converges to $\gamma\approx0.50$ at $k=20$, lying inside the obstacle, so that the fundamental solution strictly satisfies the governing equation and the radiation condition in the exterior domain.

\begin{table}[htbp]
\centering
\small
\caption{Case~3: average relative $L_2$ error (averaged over the real/imaginary parts) of the two kernel-basis forms, collocation and boundary integral, and of their SVD-truncated variants, on the complex Helmholtz problem in an unbounded exterior domain, as a function of the wavenumber $k$. The SVD truncation order is selected offline according to the boundary reconstructability criterion: $q=40$ for the collocation form at all wavenumbers, and $q=40,40,75,110$ for the boundary integral form at $k=5,10,15,20$, respectively.}
\label{tab:case3_results}
\begin{tabular}{lcccc}
\toprule
Wavenumber $k$ & $5$ & $10$ & $15$ & $20$ \\
\midrule
KernelOnet-PIKF (collocation) & $1.14\times10^{-3}$ & $1.37\times10^{-3}$ & $2.64\times10^{-3}$ & $1.90\times10^{-3}$ \\
KernelOnet-PIKF-SVD (collocation $+$ SVD) & $1.12\times10^{-3}$ & $1.25\times10^{-3}$ & $1.30\times10^{-3}$ & $1.36\times10^{-3}$ \\
KernelOnet-PIKF-SL (boundary integral) & $2.62\times10^{-3}$ & $2.49\times10^{-3}$ & $3.25\times10^{-3}$ & $5.15\times10^{-3}$ \\
KernelOnet-PIKF-SL-SVD (boundary integral $+$ SVD) & $1.31\times10^{-3}$ & $1.40\times10^{-3}$ & $1.56\times10^{-3}$ & $1.85\times10^{-3}$ \\
\bottomrule
\end{tabular}
\end{table}

\begin{figure}[htbp]
\centering
\includegraphics[width=\textwidth]{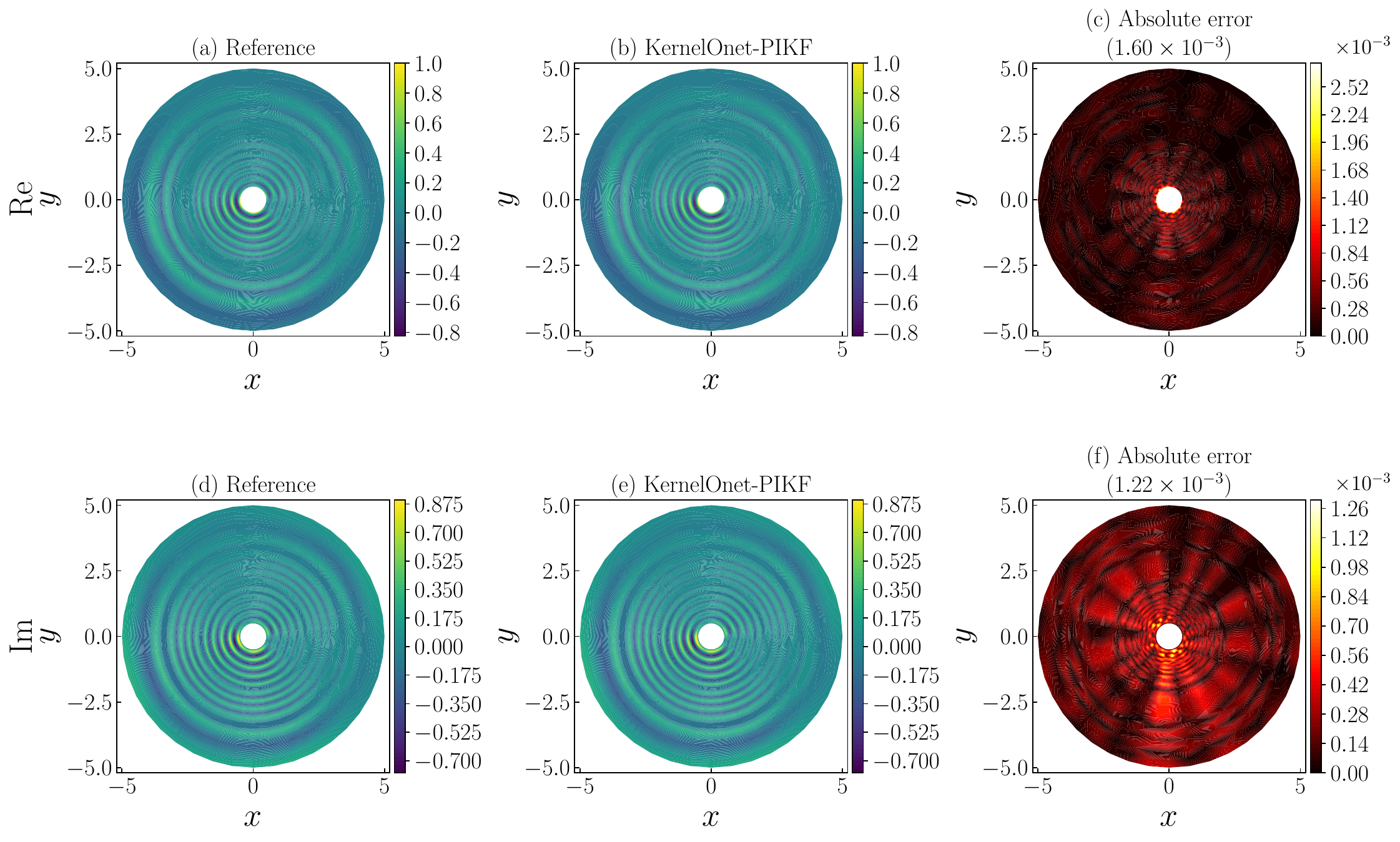}
\caption{Case~3: predicted solution and absolute error field for a representative test sample at $k=20$ (KernelOnet-PIKF, collocation form). The top and bottom rows show the real and imaginary parts of the solution, respectively, and the three columns give in turn the MFS reference solution, the KernelOnet-PIKF predicted solution, and the absolute error field annotated with the relative $L_2$ error.}
\label{fig:case3_fields}
\end{figure}

Figure~\ref{fig:case3_fields} gives the predicted solution and absolute error field of KernelOnet-PIKF (collocation form) on a representative test sample at $k=20$: the top and bottom rows show the real and imaginary parts of the solution, respectively, and the three columns give in turn the MFS reference solution, the predicted solution, and the absolute error field. The method accurately captures the oscillation and outward decay of the solution in the unbounded exterior domain: the predicted fields of the real and imaginary parts are visually almost indistinguishable from the reference solution, and the error is mainly concentrated in the near-field region around the circular boundary and decays rapidly outward. Since the expansion strictly satisfies the governing equation and the Sommerfeld radiation condition, the error field is likewise an outward-propagating solution satisfying that equation and radiation condition, and its distribution is determined by the boundary fitting residual; it is therefore concentrated in the boundary neighborhood and decays toward the far field.

\subsection{Case~4: underwater acoustic radiation and propagation caused by spherical-shell vibration in a shallow-water waveguide}

As an extension from benchmark problems to engineering applications, this case applies KernelOnet to a practically oriented acoustic problem: underwater acoustic radiation and propagation caused by the vibration of a shell structure in a shallow-water environment.
    This problem is widespread in application scenarios such as structural acoustics of underwater vehicles and ocean engineering / ocean ambient noise, and the difficulty of solving it lies in the fact that, as acoustic waves propagate in the shallow-water waveguide, they undergo multiple reflections from the sea surface and the seafloor, forming a complex waveguide interference structure; therefore, a dedicated Green's function capable of characterizing the waveguide reflection effects must be used, rather than the free-space fundamental solution;
    at the same time, the computational domain is usually unbounded, and both the complex near-field acoustic field of the structure and the traveling-wave propagation in the far field must be captured simultaneously.
    This scenario is a direct continuation of the unbounded-exterior-domain traveling-wave problem of Case~3, but the physical model is closer to engineering practice.
    The problem background and model setup of this case follow the study of Fu et al. on shell acoustic radiation in a shallow-water waveguide \cite{fu2020hybridfemsbm}.

\begin{figure}[htbp]
\centering
\includegraphics[width=0.6\textwidth]{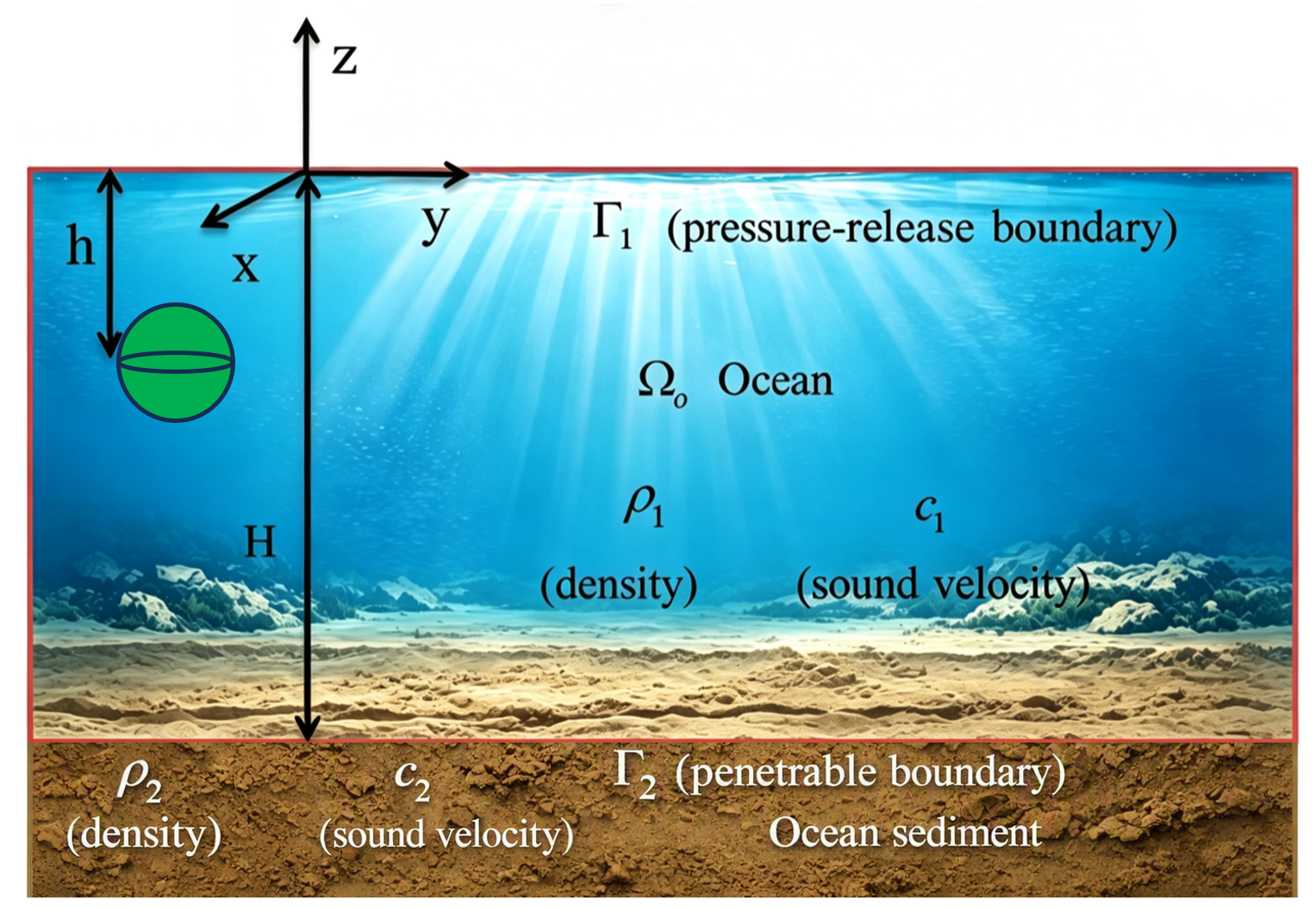}
\caption{Case~4: schematic of the numerical model for shell structural acoustic radiation in a shallow-water waveguide (not to scale). The origin of the rectangular coordinates $(x,y,z)$ lies at the sea surface, with the $z$ axis pointing vertically upward; $\Omega_o$ is the seawater domain (density $\rho_1$, sound speed $c_1$); the sea surface is a pressure-release boundary $\Gamma_1$, and at a depth $H$ below it lies a penetrable boundary $\Gamma_2$, beneath which is the seafloor sediment layer (density $\rho_2$, sound speed $c_2$). This case takes a spherical-shell radius $R=0.5$~m, sphere-center depth $h=15$~m, sea depth $H=25$~m, and frequency $f=240$~Hz; the sea-surface reflection coefficient is $a_2=-1$ and the seafloor reflection coefficient is $a_1=0.4626$.}
\label{fig:case4_geom}
\end{figure}

As shown in Figure~\ref{fig:case4_geom}, consider a thin-walled spherical shell fully immersed in a shallow sea, with radius $R=0.5$~m, sphere center at depth $h=15$~m, and sea depth $H=25$~m, seawater density $\rho_1=1025$~kg/m$^3$, seafloor sediment density $\rho_2=2600$~kg/m$^3$, and sound speed $c_2=1620$~m/s.
    Rectangular coordinates $(x,y,z)$ are adopted, with the origin at the sea surface and the $z$ axis pointing vertically upward; the sea surface $z=0$ is a pressure-release boundary and the seafloor $z=-H$ is a penetrable boundary; the center of the spherical shell is located at $z=-h$.
Since the spherical-shell structure and its surface vibration are axisymmetric about the $z$ axis, the acoustic field $p$ is independent of the azimuthal angle $\eta$, and the three-dimensional model can be reduced to a two-dimensional axisymmetric problem (consistent with the axisymmetric treatment of Fu et al.~\cite{fu2020hybridfemsbm}). Collocation points $\{\theta_j\}$ are taken uniformly along the two-dimensional axisymmetric shell boundary ($\theta_j$ being the polar angle relative to the $z$ axis), and each collocation point corresponds to one azimuthal ring on the three-dimensional shell surface; the source points are arranged along the same meridian.
    The acoustic propagation frequency is set to $f=240$~Hz, corresponding to the angular frequency $\omega=2\pi f$.
It should be noted that the sound speed in a real ocean is a depth-dependent sound speed profile; however, since the depth variation of the sound speed has little influence on near-field underwater acoustic propagation (negligible compared with the far field), an isospeed approximation is adopted in the near-field computation of this case, with a reference sound speed $c_1=1510$~m/s and the corresponding wavenumber $k=\omega/c_1$.
In the far field, the depth-dependent sound speed profile $c_1(z)$ is instead incorporated, and three profiles are examined together, $c_1(z)=1507-0.24z$, $c_1(z)=1510$, and $c_1(z)=1513+0.24z$, corresponding respectively to near-bottom-accelerating, uniform, and near-bottom-decelerating shallow-water waveguides;
    under different profiles, the modes and eigenvalues of the normal modes differ, and the required propagating modes and the waveguide interference structure change accordingly.
    The acoustic field satisfies the Helmholtz equation in the frequency domain
\begin{equation}
\nabla^2 p + k^2 p = 0,
\end{equation}
where $p$ is the complex acoustic pressure: its modulus $|p|$ represents the amplitude of the acoustic-pressure oscillation at that frequency, and its argument $\arg p$ represents the phase of the oscillation. Since, in a linear frequency-domain acoustic field, $p$ differs from the velocity potential only by a constant factor, $p$ thereby also carries the meaning of the velocity potential. For ease of presentation, the sound pressure level (SPL) is taken as $\mathrm{SPL}=20\log_{10}(|p|/10^{-6})$~dB, i.e., with $10^{-6}$~Pa as the reference pressure and the decibel as the unit for the acoustic-pressure amplitude; the acoustic-pressure distributions in the figures of this section are displayed in this quantity.

\begin{figure}[htbp]
\centering
\includegraphics[width=\textwidth]{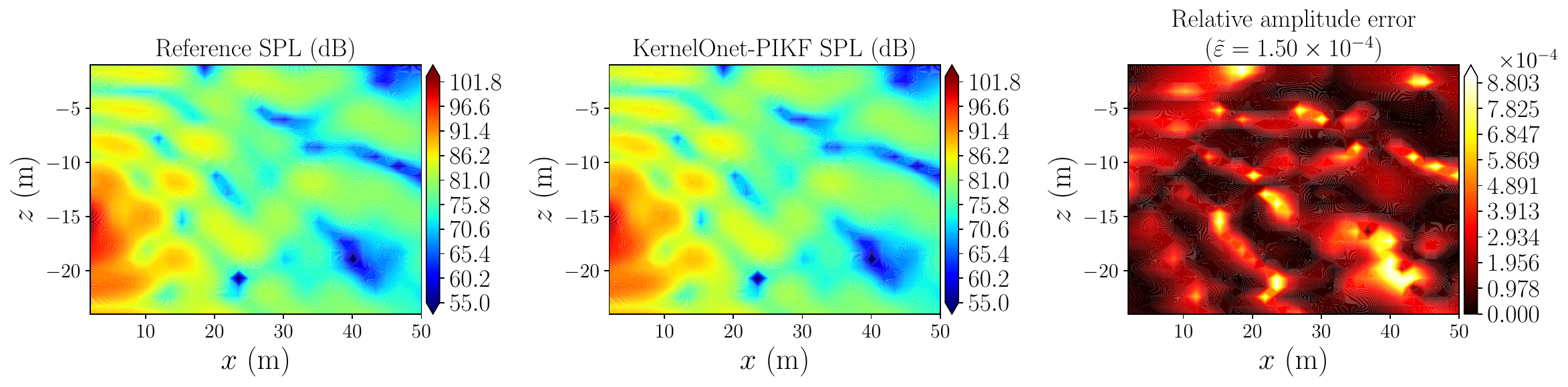}
\caption{Case~4 (near-field acoustic radiation in a shallow-water waveguide, $x\in[2,50]$~m, $z\in[-24,-1]$~m): reference solution (left), unsupervised KernelOnet-PIKF prediction (middle), and relative amplitude error (right) for a representative test sample. The left and middle columns show the sound pressure level SPL $=20\log_{10}(|p|/10^{-6})$ (dB), and the right column shows the relative amplitude error $\tilde{\varepsilon}=\bigl||p_{\mathrm{pred}}|-|p_{\mathrm{ref}}|\bigr|/|p_{\mathrm{ref}}|$. The shallow-water waveguide interference structure formed by multiple reflections at the sea surface and the seafloor is clearly visible, and the prediction is visually almost indistinguishable from the reference solution.}
\label{fig:case4_near}
\end{figure}

The kernel functions of the near and far fields are taken as the shallow-water waveguide Green's function $G^{n}$ and the normal-mode Green's function $G^{f}$, respectively: $G^{n}$ explicitly incorporates the waveguide reflection effects through an infinite mirror-image superposition of the sea surface and the seafloor and is applicable to the near field for incidence angles below $40^\circ$; $G^{f}$ is formed by superposing the normal modes determined by the sound speed profile and is applicable to the far field. Their definitions, the incidence-angle dependence of the reflection coefficients, and the normal-mode eigenvalue problem are given in Appendix~\ref{app:wggreen}. This case constructs the kernels separately for the near and far fields and solves them separately: the near-field kernel is summed over $N_\eta$ virtual nodes on the ring in the form of a ring source, whereas the far-field kernel is evaluated as a point source since the field distance is far larger than the shell size; the reference acoustic fields use $G^{n}$ and $G^{f}$ as kernels, respectively, and are synthesized after the virtual source strengths are obtained by least squares from the same Dirichlet data on the shell-surface collocation points. The operator takes the complex acoustic pressure $p(\theta_j)$ on the shell-surface collocation points as input and the acoustic field as output, i.e., it is the solution operator from the shell-surface complex acoustic pressure to the acoustic field. Since the far-field kernel is determined by the sound speed profile $c_1(z)$, a dataset is generated for each of the three profiles and one operator model is trained for each, so as to examine the adaptability of the operator to different waveguides. The source points of both the near- and far-field kernels are taken on concentric virtual spheres obtained by contracting the shell-surface collocation points radially toward the sphere center; the kernel matrices are numerically ill-conditioned under this source-point arrangement, so the boundary kernel matrices are preprocessed with the SVD orthogonal kernel basis according to Appendix~\ref{app:svd}, with $q=40$ for the near field; for the far field, because the magnitudes of the kernel values on the evaluation points and of the boundary kernel values differ drastically, the retained directions are instead selected according to the input subspace (taking the orthogonal basis of the subspace spanned by the shell-surface vibration modes, of dimension $J=4$; the criterion is given in Appendix~\ref{app:svd}), and the complex coefficients are output by the branch network.

Table~\ref{tab:case4_results} summarizes the relative $L_2$ errors of the various configurations on the test set together with three acoustic quality metrics. All three metrics are defined in terms of the complex acoustic pressure and are computed sample by sample and then averaged over the $2000$ test samples: the relative amplitude error is $\bigl||p_{\mathrm{pred}}|-|p_{\mathrm{ref}}|\bigr|/|p_{\mathrm{ref}}|$, the phase error is $\bigl|\arg(p_{\mathrm{pred}}/p_{\mathrm{ref}})\bigr|$, and the transmission loss deviation is the root mean square of $\mathrm{TL}_{\mathrm{pred}}-\mathrm{TL}_{\mathrm{ref}}$, where $\mathrm{TL}=-20\log_{10}\bigl(|p(r,z)|/|p(r_0,z_0)|\bigr)$ and the reference point $(r_0,z_0)$ is taken at the source depth at the closest distance within the window concerned; the first two are averaged over the evaluation points. The statistical windows are consistent with the corresponding acoustic-field figures, i.e., $x\in[2,50]$~m for the near field and $x\in[950,1000]$~m for the far field, and only the interior of the water column is counted, excluding one row each at the sea surface and the seafloor, so as to avoid meaningless relative errors where $p\equiv0$ on the pressure-release boundary. For the near field, the unsupervised KernelOnet-PIKF (collocation form, $q=40$) attains an accuracy of $8.57\times10^{-4}$; as in Cases 1 and 3, this shows that when the kernel function is physically consistent, high accuracy can be obtained without any interior-solution labels. For the far field, the normal-mode kernel is evaluated as a point source and contains only $3$ propagating modes, and its shell-surface boundary matrix is more ill-conditioned than the near-field ring-source kernel (with a condition number of $10^{17}$); if truncated according to the first $q$ singular directions of the boundary kernel matrix, the projection error on the shell surface is only $2.4\times10^{-4}$, but the amplification factor of the discarded directions on the far field, $w_k=\lVert G_i\mathbf{v}_k\rVert_2/\sigma_k$, is as high as $10^{3}\sim10^{8}$, and the far-field truncation error reaches $0.61$, consistent with the truncation floor, indicating that its root cause is the representational capability of the kernel basis rather than insufficient optimization, and that no amount of training can get past it (measured value $0.609$). The far field therefore instead selects the retained directions according to the input subspace: the input is a complex linear combination of $J=4$ shell-surface vibration modes, whose spanning subspace has orthogonal basis $\Psi_d$; the relative amplification factor on this subspace is only about $2$, and the boundary residual again becomes a stable surrogate for the interior field. In this setting, the unsupervised KernelOnet-PIKF attains $1.18\times10^{-3}$, $1.31\times10^{-3}$, and $1.18\times10^{-3}$ under the three sound speed profiles, respectively, without using any interior acoustic-field labels during training, showing that the operator with the normal-mode Green's function as its kernel is applicable to different sound speed profiles. The three quality metrics corroborate the relative $L_2$ error: the relative amplitude error is $7.2\times10^{-4}\sim1.0\times10^{-3}$, comparable in magnitude to the relative $L_2$ error, indicating that the prediction error comes mainly from the amplitude rather than the phase; the phase error does not exceed $4.1\times10^{-2}$ degrees, which, converted with the near-field wavenumber $k=\omega/c_1\approx1.0$~rad/m, corresponds to an equivalent acoustic path error of less than $1$~mm, far smaller than the wavelength of about $6.3$~m; and the transmission loss deviations are all below $10^{-2}$~dB, far smaller than the variation scale of the transmission loss of the reference solution itself, which decays by $23.0$~dB within the near-field window and fluctuates by about $2.5$~dB within the far-field window, see Table~\ref{tab:app_tl} (see Appendix~\ref{app:supp}). In addition, under the three sound speed profiles, the maximum relative differences in the relative $L_2$ error, the relative amplitude error, and the phase error all do not exceed $20\%$, and although the transmission loss deviation varies relatively more, its absolute value is likewise below $10^{-2}$~dB, further indicating that the operator is insensitive to changes in the sound speed profile.

\begin{table}[htbp]
\centering
\small
\caption{Case~4: training results with the unsupervised KernelOnet-PIKF for both the near field and the far field under the various sound speed profiles. The near field uses the collocation form with truncation order $q=40$, and the far field selects the retained directions of the kernel basis according to the input subspace; the relative $L_2$ error is reported separately for the real and imaginary parts and then averaged, and the definitions and statistical conventions of the relative amplitude error, the phase error, and the transmission loss deviation are given in the main text.}
\label{tab:case4_results}
\resizebox{\textwidth}{!}{%
\begin{tabular}{llcccc}
\toprule
Solution configuration & Kernel function & Relative $L_2$ & Relative amplitude error & Phase error ($^\circ$) & Transmission loss deviation (dB) \\
\midrule
Near field \ KernelOnet-PIKF & Waveguide Pekeris $G^{n}$ (ring source) & $8.57\times 10^{-4}$ & $7.17\times 10^{-4}$ & $3.2\times 10^{-2}$ & $6.7\times 10^{-3}$ \\
Far field \ KernelOnet-PIKF ($c_1(z)=1507-0.24z$) & Normal mode $G^{f}$ & $1.18\times 10^{-3}$ & $9.04\times 10^{-4}$ & $3.9\times 10^{-2}$ & $9.1\times 10^{-3}$ \\
Far field \ KernelOnet-PIKF ($c_1(z)=1510$) & Normal mode $G^{f}$ & $1.31\times 10^{-3}$ & $1.02\times 10^{-3}$ & $4.1\times 10^{-2}$ & $7.5\times 10^{-3}$ \\
Far field \ KernelOnet-PIKF ($c_1(z)=1513+0.24z$) & Normal mode $G^{f}$ & $1.18\times 10^{-3}$ & $8.80\times 10^{-4}$ & $3.9\times 10^{-2}$ & $4.4\times 10^{-3}$ \\
\bottomrule
\end{tabular}}
\end{table}

Figure~\ref{fig:case4_near} gives, for a random test sample in the near field, the reference solution, the predicted solution, and the relative amplitude error of the unsupervised KernelOnet-PIKF;
    the left and middle columns use the sound pressure level SPL (dB) color scale, so as to reveal simultaneously the strong field near the shell surface and the weak field in the far region.
    It can be seen that the predicted solution and the reference solution are visually almost indistinguishable: the acoustic field is strongest near the spherical shell, decays rapidly outward along the radial direction, and exhibits oblique modal interference fringes formed by multiple reflections at the sea surface and the seafloor, which is exactly the typical feature distinguishing shallow-water waveguide propagation from free-space propagation.
    The error field is mainly concentrated in the near-field region around the spherical shell and decays rapidly outward.
    In the error field, the relative deviation is most pronounced at the waveguide interference nodes, i.e., where $|p|$ is small: a node is where the multiple paths of sea-surface and seafloor reflections cancel each other coherently, and its residual amplitude is determined by the tiny imbalance among the complex phases of the various paths, making it extremely sensitive to the complex-phase error of the kernel function $G^{n}$---with even a slight deviation in the complex phase, the waves that should cancel cannot cancel strictly, and the node positions shift accordingly.

\begin{figure}[htbp]
\centering
\includegraphics[width=\textwidth]{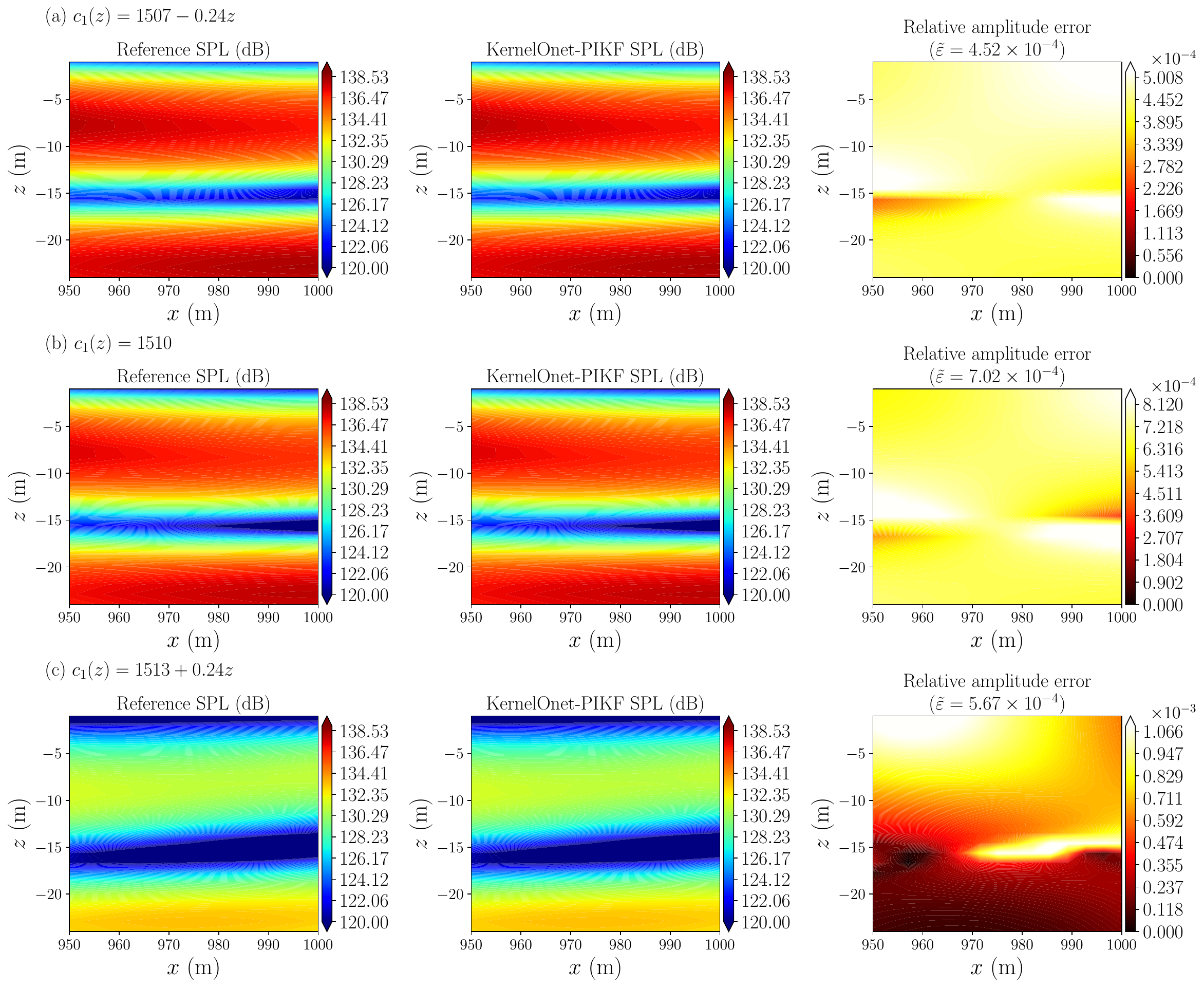}
\caption{Case~4 (far-field acoustic propagation using the normal-mode kernel, $x\in[950,1000]$~m, $z\in[-24,-1]$~m): reference solution (left column), unsupervised KernelOnet-PIKF prediction (middle column), and relative amplitude error (right column) for a representative test sample, corresponding to three shallow-water sound speed profiles $c_1(z)$. Here $z$ is the vertical coordinate (sea surface $z=0$, seafloor $z=-H$); the left and middle columns show the sound pressure level SPL (dB), and the right column shows the relative amplitude error $\tilde{\varepsilon}=\bigl||p_{\mathrm{pred}}|-|p_{\mathrm{ref}}|\bigr|/|p_{\mathrm{ref}}|$. The three profiles: (a) $c_1(z)=1507-0.24z$, (b) $c_1(z)=1510$, and (c) $c_1(z)=1513+0.24z$. Under all three profiles, the operator accurately reproduces the far-field waveguide modal interference structure (nodes and antinodes along the depth direction); the figure shows a single test sample, and the average errors over the test set are given in Table~\ref{tab:case4_results}.}
\label{fig:case4_far}
\end{figure}

Figure~\ref{fig:case4_far} shows the reference solution, the unsupervised KernelOnet-PIKF predicted solution, and the relative amplitude error in the far field under the three sound speed profiles, at distances $x\in[950,1000]$~m.
    Each row corresponds to one sound speed profile $c_1(z)$, where $z$ is the vertical coordinate (sea surface $z=0$, seafloor $z=-H$), in turn $c_1(z)=1507-0.24z$, $c_1(z)=1510$, and $c_1(z)=1513+0.24z$.
    The far-field acoustic pressure exhibits a clear normal-mode standing-wave structure along the depth direction, and the reference solution and the predicted solution agree closely, indicating that the operator has successfully learned the coherent-superposition law of the normal modes. Similar to the near field, the relative amplitude deviation in the error field is most pronounced near the waveguide interference nodes (where $|p|$ is small), which is caused by the coherent cancellation of the multiple paths at a node and by the fact that the residual amplitude is determined by the imbalance among the complex phases of the various paths.
    At the same time, comparing the positions of the modal standing waves in the three rows, one can observe that the node/antinode positions shift as the sound speed profile changes, indicating that the far-field waveguide modal structure is sensitive to the sound speed profile; this observation is consistent with the conclusions of Fu et al.
    Although the modal structure changes markedly with the profile, the operator can still stably characterize the far-field propagation in the three waveguides, further verifying the ability of KernelOnet-PIKF with the normal-mode Green's function as its kernel to characterize the modal propagation properties of the shallow-water far-field waveguide.
As a reference, Table~\ref{tab:app_tl} (see Appendix~\ref{app:supp}) gives the transmission loss of the reference solution itself: in the near field it decreases by $23.0$~dB from $2$~m outward at $26.8$~m, close to the $22.5$~dB of spherical spreading; within the far-field window it fluctuates by only about $2.5$~dB, whereas cylindrical spreading over the same distance brings only about $0.2$~dB of loss, showing that the far-field structure is dominated by modal interference.



\section{Conclusion and Future Work}

This paper has proposed and systematically verified the Kernel Operator Network (KernelOnet), an interpretable operator learning framework that embeds kernel functions explicitly into a neural operator.
    By replacing the implicit trunk network of DeepONet with explicit kernel functions, KernelOnet is mathematically consistent with the kernel expansion of meshless collocation methods, and three complementary kernel construction strategies are given, namely the data-driven learnable kernel (KernelOnet-RBF), the physics-informed kernel (KernelOnet-PIKF) and the hybrid kernel (KernelOnet-HK), which together form a spectrum from a strong physical prior to fully data-driven learning; the accuracy basis of the three can also be given respectively by reproducing kernel space theory and by the spectral convergence of the fundamental-solution expansion.
    In the four examples---the Laplace equation on a circular domain, the nonlinear modified Helmholtz equation on a star-shaped domain, the complex Helmholtz equation in an unbounded exterior domain, and the underwater acoustic radiation and propagation caused by spherical-shell vibration in a shallow-water waveguide---KernelOnet achieved high-accuracy solutions in all cases; on the two examples that can be compared directly with DeepONet it attains higher accuracy with fewer learnable parameters. Its advantages can be summarized along three main lines.
    The first is physical consistency. The physics-informed kernel hard-codes the governing equation into the network structure, so that the expansion satisfies the governing equation exactly, can therefore be trained without supervision and without interior solution labels, and dispenses with the cost of repeatedly evaluating the PDE residual by automatic differentiation; this is confirmed by all configurations of Examples~3 and~4 as well as by KernelOnet-PIKF in Example~1. For unbounded exterior domains and travelling-wave propagation problems, KernelOnet-PIKF, which takes the analytic fundamental solution as its kernel, is also the only one of the operator learning methods compared in this paper that is directly applicable.
    The second is interpretability. The analytic fundamental-solution kernel allows the homogeneous part of the expansion to be compared pointwise with the analytic fundamental solution, while the few coefficients and kernel parameters of the correction branch give physical insight into the nonlinear source term: in Example~1 the data-driven kernel function is almost recovered as an affine image of the fundamental solution, whereas in Example~2 the $L_2$ energy proportion of the correction branch grows from $0.3\%$ to $26.6\%$ with the nonlinearity, directly quantifying the weight of the nonlinear component in the solution.
    The third is computational efficiency and engineering applicability. For problems that admit an analytic fundamental solution, the physics-informed kernel and the hybrid kernel attain high accuracy without interior solution labels; for strongly nonlinear and variable-coefficient problems that admit none, the data-driven KernelOnet-RBF compensates for the breadth of applicability by adaptively learning the kernel shape, at the price of a weaker physical prior. In the shallow-water waveguide example, both the near-field ring-source kernel and the far-field normal-mode kernel are numerically ill-conditioned, and once the retained directions of the kernel basis are selected from the input subspace both the near field and the far field can be trained without supervision and remain robust under three sound speed profiles. In addition, the inference cost of the operator is far below that of per-instance solvers, and in nonlinear problems and scenarios with expensive boundary integrals the training cost is recovered after a few hundred to two thousand queries; once training is complete, the operator can be evaluated directly on evaluation grids of arbitrary resolution without retraining.
    These results show that combining explicit kernel functions with physical priors achieves a good compromise among accuracy, interpretability, computational efficiency and breadth of applicability.

Several limitations remain in this work, and they indicate corresponding directions for improvement. First, the construction of the physics-informed kernel has not yet been automated: the analytic fundamental solution or Green's function must be derived by hand for each specific physical setting---Example~4, for instance, required the separate construction of a near-field mirror-image superposition kernel and a far-field normal-mode kernel for the shallow-water waveguide, together with a separate delimitation of their respective ranges of validity. A feasible improvement is to build a library of kernel functions for common operators and media and to automate derivations such as the method of images and separation of variables; for media that cannot be solved analytically, one may precompute numerical Green's functions, or retain the structure of the hybrid kernel and let the low-rank learned branch absorb the bias of the analytic kernel. Second, when extending to high frequencies and large scales the size of the kernel basis grows accordingly: the order of the SVD truncation introduced to suppress ill-conditioning rises markedly with the wavenumber---in Example~3 the single-layer potential kernel grows from $40$ at $k=5$ to $110$ at $k=20$, and the far field of Example~4 even requires a switch to a criterion that selects the directions from the input subspace---which shows that the selection of the retained directions still relies on case-by-case analysis. The corresponding improvement is to incorporate the amplification factor $w_k$ of the evaluation block into an automatic order-selection criterion, and to combine preconditioning, block low-rank factorization and multiscale kernel bases layered by frequency band, so as to decouple the size of the kernel basis from the wavenumber. Third, the theoretical guarantee of the overall approximation is still incomplete: the error bounds of Section~2.5 presuppose the positive definiteness of the learned kernel, and the spectral convergence of the fundamental-solution expansion is an asymptotic conclusion; the two have not yet been combined into a convergence rate and generalization error bound that covers both the kernel expansion and the learning of the coefficients by the branch network. One may, within the native space framework, regard the operator approximation as a restricted approximation problem, decompose the error according to the size of the kernel basis and the number of training samples, and give an explicit rate for the correction branch of the hybrid kernel under a compressibility assumption on the source term.

\pdfbookmark[1]{CRediT authorship contribution statement}{bmark-credits}
\section*{CRediT authorship contribution statement}
Yuan Guo: Writing -- review \& editing, Writing -- original draft, Visualization, Validation, Methodology, Investigation, Formal analysis, Data curation, Conceptualization.
Hanshu Chen: Writing -- review \& editing, Software, Validation.
Qiang Xi: Writing -- review \& editing, Software, Validation.
Timon Rabczuk: Supervision, Writing -- review \& editing.
Zhuojia Fu: Writing -- review \& editing, Methodology, Supervision, Project administration, Funding acquisition.

\pdfbookmark[1]{Declaration of competing interest}{bmark-decl}
\section*{Declaration of competing interest}
The authors declare that they have no known competing financial interests or personal relationships that could have appeared to influence the work reported in this paper.

\pdfbookmark[1]{Data availability}{bmark-data}
\section*{Data availability}
All data and code will be made available upon acceptance of the manuscript at \url{https://github.com/YuanGuo-hhu/KernelOnet}.

\pdfbookmark[1]{Acknowledgments}{bmark-ack}
\section*{Acknowledgments}
The research was supported by the National Natural Science Foundation of China (12122205, 12372196).

\appendix

\renewcommand{\thetable}{A\arabic{table}}
\setcounter{table}{0}

\section{Common fundamental solutions of differential operators}
\label{app:fund}

The KernelOnet-PIKF presented in Section 2.4 of this paper uses the analytic fundamental solution of the governing equation as the kernel function, whose applicability presupposes that the governing equation admits an explicit fundamental solution.
To facilitate the direct construction of KernelOnet-PIKF for different problems, this appendix collects the fundamental solutions of several common linear differential operators in two dimensions (2D) and three dimensions (3D);
the complete listing of the corresponding harmonic functions and radial Trefftz functions can be found in the appendix of the physics-informed kernel function neural network (PIKFNN) \cite{fu2024pikfnn}.

Let $\mathcal{L}$ be a constant-coefficient linear partial differential operator, $\mathbf{x}\in\mathbb{R}^{d}$, the radial distance between the field point $\mathbf{x}$ and the source point $\mathbf{x}_s$ be $r=|\mathbf{x}-\mathbf{x}_s|$, and the fundamental solution $\Phi$ satisfy
\begin{equation}
\mathcal{L}\,\Phi(\mathbf{x},\mathbf{x}_s)=-\delta(\mathbf{x}-\mathbf{x}_s).
\end{equation}
Table~\ref{tab:fundamental_solutions} lists the fundamental solutions of several common operators $\mathcal{L}$ in 2D and 3D, where $\Delta$ is the Laplace operator, $k$ is the wavenumber (or decay coefficient), $D$ is the diffusion coefficient, $\mathbf{v}$ is the convection velocity, $\mu=\sqrt{k^2/D+|\mathbf{v}|^2/(4D^2)}$, $H_0^{(1)}$ is the zero-order Hankel function of the first kind, and $K_0$, $J_0$, $I_0$ are the zero-order modified Bessel function of the second kind, the zero-order Bessel function of the first kind, and the zero-order modified Bessel function of the first kind, respectively.
These fundamental solutions satisfy the corresponding homogeneous governing equation ($\mathcal{L}\Phi=0$, $r\neq0$), and can therefore be directly used as the kernel backbone of KernelOnet-PIKF.

\begin{table}[htbp]
\centering
\small
\caption{Fundamental solutions $\Phi$ of common differential operators (2D and 3D).}
\label{tab:fundamental_solutions}
\begin{tabular}{lcc}
\toprule
Operator $\mathcal{L}$ & 2D & 3D \\
\midrule
$\Delta$ (Laplace) & $-\dfrac{1}{2\pi}\ln r$ & $\dfrac{1}{4\pi r}$ \\[2pt]
$\Delta+k^{2}$ (Helmholtz) & $\dfrac{i}{4}H_0^{(1)}(kr)$ & $\dfrac{e^{-ikr}}{4\pi r}$ \\[2pt]
$\Delta-k^{2}$ (modified Helmholtz) & $\dfrac{1}{2\pi}K_0(kr)$ & $\dfrac{e^{-kr}}{4\pi r}$ \\[2pt]
$D\Delta+\mathbf{v}\cdot\nabla-k^{2}$ (convection--diffusion) & $\dfrac{1}{2\pi}K_0(\mu r)\,e^{-\frac{\mathbf{v}\cdot\mathbf{r}}{2D}}$ & $\dfrac{e^{-\mu r-\frac{\mathbf{v}\cdot\mathbf{r}}{2D}}}{4\pi r}$ \\[2pt]
$\Delta^{2}$ (biharmonic) & $-\dfrac{r^{2}\ln r}{8\pi}$ & $\dfrac{r}{8\pi}$ \\[2pt]
\bottomrule
\end{tabular}
\end{table}

It should be emphasized that the above fundamental solutions (together with their corresponding higher-order variants and the fundamental solutions of time-dependent operators, such as those of the heat operator $\partial/\partial t-k\Delta$ and the wave operator $\partial^{2}/\partial t^{2}-c_{1}^{2}\Delta$) can all be constructed as PIKF in the manner described in Section 2.3 and Section 2.4.
If the governing equation of a certain class of problems itself admits no analytic fundamental solution (such as nonlinear equations with nonlinear source terms), then the strict version of KernelOnet-PIKF is no longer applicable;
in this case, one may construct the hybrid kernel KernelOnet-HK using the fundamental solution of the linear principal part of the governing equation, or instead fully adopt the data-driven KernelOnet-RBF; the specific trade-off is discussed in Section 2.4.

\section{SVD Orthogonalization of the Kernel Basis}
\label{app:svd}

When the kernel function family is numerically ill-conditioned under a given discretization of source points and evaluation points, using it directly as the kernel severely pollutes the gradients. To this end, this paper applies an orthogonalization preprocessing of the kernel basis, based on the singular value decomposition (SVD), to the kernel matrix $G(\mathbf{x})$: for the collocation form (Eq.~\eqref{eq:pikf_colloc}) and the boundary integral form (Eq.~\eqref{eq:pikf_bie}), the basis function family $\psi_j(\mathbf{x})$ consists of linear combinations (pointwise or boundary-integral) of fundamental solutions, so the preprocessing steps are identical for the two, and only the construction of $G(\mathbf{x})$ differs.

For the collocation form (Eq.~\eqref{eq:pikf_colloc}), the basis functions are $\psi_j(\mathbf{x})=\Phi\bigl(\left|\mathbf{x}-\gamma\,\mathbf{x}_b^{(j)}\right|\bigr)$. The kernel matrix $G(\mathbf{x})$ is constructed from the $B$ basis functions as follows: at a single evaluation point $\mathbf{x}$, $\mathbf{g}(\mathbf{x})=\bigl[\psi_1(\mathbf{x}),\ldots,\psi_B(\mathbf{x})\bigr]\in\mathbb{C}^{1\times B}$ is the row vector of the kernel matrix at that point, so that the $j$-th column of the kernel matrix is the basis function $\psi_j(\mathbf{x})$; stacking these rows at the $n_t$ evaluation points $\{\mathbf{x}_m\}_{m=1}^{n_t}$ yields the discrete kernel matrix $G(\mathbf{x})\in\mathbb{C}^{n_t\times B}$ with entries $G_{mj}=\psi_j(\mathbf{x}_m)$. A singular value decomposition is then performed on this kernel matrix

\begin{equation}
G(\mathbf{x})=U(\mathbf{x})\,\Sigma(\mathbf{x})\,V^{H}(\mathbf{x}),\qquad \Sigma(\mathbf{x})=\mathrm{diag}(\sigma_1(\mathbf{x}),\sigma_2(\mathbf{x}),\ldots,\sigma_r(\mathbf{x})),\qquad \sigma_1(\mathbf{x})\ge\sigma_2(\mathbf{x})\ge\cdots\ge\sigma_r(\mathbf{x})>0,
\end{equation}

where $r\le\min(n_t,B)$ is the rank of the matrix, and $U(\mathbf{x})=[\mathbf{u}_1(\mathbf{x}),\ldots,\mathbf{u}_r(\mathbf{x})]$ and $V(\mathbf{x})=[\mathbf{v}_1(\mathbf{x}),\ldots,\mathbf{v}_r(\mathbf{x})]$ are the left and right singular vectors, respectively: the left singular vector $\mathbf{u}_k(\mathbf{x})$ is defined on the evaluation points and is a function of spatial position, while the right singular vector $\mathbf{v}_k(\mathbf{x})$ is likewise a function of spatial position and varies with the evaluation point set; $(\cdot)^{H}$ denotes the conjugate transpose. When the kernel function family is approximately linearly dependent under the source-point arrangement, the singular values decay sharply from some point onward, the numerical rank is far smaller than the number of collocation points, and the condition number $\mathrm{cond}(G(\mathbf{x}))=\sigma_1(\mathbf{x})/\sigma_r(\mathbf{x})$ reaches as high as $10^{16}$, whereupon the small singular directions amplify tiny perturbations of the boundary fitting residual into huge fluctuations of the expansion coefficients and make the gradient computation severely ill-conditioned. Taking the first $q$ principal singular directions forms a well-conditioned orthogonal kernel basis

\begin{equation}
U_q(\mathbf{x})=U(\mathbf{x})\,[:,1:q],
\end{equation}

The selection of retained directions cannot rest solely on the magnitude of the singular values of the boundary kernel matrix; the amplification of the evaluation block must also be taken into account. Let $G^{(b)}(\mathbf{x})$ denote the boundary kernel matrix and $G(\mathbf{x})$ the kernel matrix at the evaluation points, with $\mathbf{v}_k$ the right singular vector of $G^{(b)}$; then the amplification factor of the $k$-th direction over the evaluation block is $w_k=\lVert G(\mathbf{x})\mathbf{v}_k\rVert_2/\sigma_k^{(b)}$. Since the interior field error caused by the boundary fitting residual $\boldsymbol{\delta}$ is exactly $\lVert G(\mathbf{x})\boldsymbol{\delta}\rVert_2$, directions for which $\sigma_k^{(b)}$ is very small while $w_k$ is very large must be retained or discarded as a whole: truncating merely by the magnitude of the singular values discards their contributions to the interior field along with them. When the boundary block and the evaluation block are of comparable magnitude (Cases 1--3), $w_k$ is bounded and taking the first $q$ principal singular directions suffices; when the two differ greatly (the normal-mode kernel of the far field in Case~4, where $w_k$ reaches $10^{3}\sim10^{8}$), the retained directions should instead be selected according to the input subspace: let the input data span a $J$-dimensional subspace with complex orthonormal basis $\Psi_d$, and take

\begin{equation}
\Psi_b=\Psi_d,\qquad \Psi_i=G(\mathbf{x})\,\mathrm{pinv}\bigl(G^{(b)}(\mathbf{x})\bigr)\,\Psi_d,
\end{equation}

that is, $\Psi_b$ and $\Psi_i$ serve as the kernel bases of the boundary block and the evaluation block, respectively. In this case the boundary residual always lies within the input subspace, its amplification to the interior field is characterized by $\Psi_i\Psi_b^{+}$ and is bounded in norm, and the boundary residual again becomes a stable surrogate for the interior field. When the input is a linear combination of known modes, $\Psi_d$ can be given analytically; otherwise it can be estimated offline by the proper orthogonal decomposition (POD) of the training boundary data matrix. Neither relies on interior solution labels.

that is, the columns $\mathbf{u}_1(\mathbf{x}),\ldots,\mathbf{u}_q(\mathbf{x})$ of $U_q(\mathbf{x})$ are taken as the kernel basis, and this kernel basis is well-conditioned. The operator output is taken as

\begin{equation}
\mathcal{G}_\theta(a)(\mathbf{x})=U_q(\mathbf{x})\,\mathbf{a},\qquad \mathbf{a}=\mathcal{B}_\theta(a)\in\mathbb{C}^{q},
\end{equation}

where the complex coefficients $\mathbf{a}$ are learned by the branch network: since the network outputs real numbers only, each complex coefficient is formed by two outputs, its real part and its imaginary part, so the output dimension of the branch network is $2q$, and the real and imaginary parts of the solution are given by the cross inner products of the real and imaginary parts of the coefficients with the real and imaginary parts of $U_q(\mathbf{x})$.

For the boundary integral form (Eq.~\eqref{eq:pikf_bie}), the basis functions are integrals of the fundamental solution over boundary patches, $\psi_j(\mathbf{x})=\int_{\Gamma_j}\Phi\bigl(\left|\mathbf{x}-\mathbf{y}\right|\bigr)\,\mathrm{d}\Gamma_{\mathbf{y}}$, i.e., flattened boundary-element kernels. The kernel matrix $G(\mathbf{x})$ is constructed in the same way and subjected to the same singular value decomposition and truncation, yielding the kernel basis $U_q(\mathbf{x})$ and the operator output in the same form. Hence, under the SVD preprocessing the collocation form and the boundary integral form are unified into the operator structure $\mathcal{G}_\theta(a)(\mathbf{x})=U_q(\mathbf{x})\,\mathbf{a}$, the only difference being the construction of the kernel matrix $G(\mathbf{x})$ (pointwise fundamental solutions or boundary integral kernels).

The accuracy of this truncation can be quantified by the singular spectrum. Suppose that the reference solution lies in the reachable space of the kernel function, i.e., there exists a coefficient vector $\boldsymbol{\alpha}$ such that $p(\mathbf{x})=G(\mathbf{x})\,\boldsymbol{\alpha}$; writing $G_q(\mathbf{x})=U_q(\mathbf{x})\,\Sigma_q(\mathbf{x})\,V_q^{H}(\mathbf{x})$, the projection error of $p$ onto the truncated subspace $\mathrm{span}(U_q(\mathbf{x}))$ satisfies
\begin{equation}
\left\|p-U_q(\mathbf{x})U_q^{H}(\mathbf{x})p\right\|_2=\left\|(I-U_q(\mathbf{x})U_q^{H}(\mathbf{x}))(G(\mathbf{x})-G_q(\mathbf{x}))\boldsymbol{\alpha}\right\|_2\le\sigma_{q+1}(\mathbf{x})\left\|\boldsymbol{\alpha}\right\|_2,
\end{equation}
that is, the truncation error is controlled jointly by the largest discarded singular value $\sigma_{q+1}(\mathbf{x})$ and the norm of the expansion coefficients. It should be emphasized that this estimate is in the boundary metric: the truncation error of the evaluation block (the interior field) is $\lVert G(\mathbf{x})(I-U_q(\mathbf{x})U_q^{H}(\mathbf{x}))\boldsymbol{\alpha}\rVert_2$, which is controlled by the amplification factor $w_k$ rather than by $\sigma_k$; a small $\sigma_{q+1}/\sigma_1$ therefore does not imply that the interior field remains accurate after truncation---the far field of Case~4 is a counterexample, where the retained directions must instead be selected according to the input subspace. The ratio of the sum of squares of the first $q$ singular values to the total sum of squares, $\sum_{k=1}^{q}\sigma_k^{2}(\mathbf{x})\big/\sum_{k=1}^{r}\sigma_k^{2}(\mathbf{x})\approx100\%$, is by itself insufficient to assert that the truncation is lossless; the premise that ``the reference solution is indeed a reachable vector of this kernel expansion'' is also required. By the Eckart--Young theorem, the error of the optimal $q$-rank approximation of $G(\mathbf{x})$ in the spectral norm is exactly the largest discarded singular value
\begin{equation}
\left\|G(\mathbf{x})-U_q(\mathbf{x})\,\Sigma_q(\mathbf{x})\,V_q^{H}(\mathbf{x})\right\|_2=\sigma_{q+1}(\mathbf{x}),
\end{equation}
and when $\sigma_{q+1}(\mathbf{x})$ has fallen to machine precision relative to $\sigma_1(\mathbf{x})$, the truncation is numerically exact.

It should be emphasized that the improvement in the condition number comes from switching to the orthogonal basis of left singular vectors rather than from the truncation itself: the columns of $U(\mathbf{x})$ are themselves mutually orthogonal and of unit norm, so $\mathrm{cond}(U(\mathbf{x}))=1$; the additional role of the truncation is to constrain the approximation space to the physically meaningful principal singular subspace, removing the spurious modes associated with small singular values that are numerically unreliable upon extrapolation, and thereby eliminating their ill-conditioned amplification of the gradients.

When the kernel matrix is fixed during training (as in the boundary integral form, or when the source-point positions are fixed), this preprocessing depends only on the kernel function and the discretization layout and is independent of the sample data; it can be completed offline once before training and introduces no learnable parameters. If the source-point positions take part in learning, then $G(\mathbf{x})$ changes accordingly and the singular value decomposition must be recomputed, so that it is no longer a one-off offline operation.

It should be emphasized that the SVD truncation does not destroy the physical consistency of the kernel. From $U_q(\mathbf{x})=G(\mathbf{x})\,V_q(\mathbf{x})\,\Sigma_q(\mathbf{x})^{-1}$ it follows that every column of $U_q(\mathbf{x})$ is a column vector of the kernel matrix $G(\mathbf{x})$, i.e., a linear combination of the basis functions (fundamental solutions or their boundary integrals); since the fundamental solutions satisfy the governing equation exactly and linear superposition does not change the nature of the solution, each column still satisfies the homogeneous governing equation and the corresponding boundary conditions, so that the expansion $U_q(\mathbf{x})\,\mathbf{a}$ automatically satisfies the governing equation for arbitrary coefficients and the property that the PDE residual vanishes identically is preserved. Therefore, training remains an unsupervised boundary residual fit, enforcing the boundary conditions only at the boundary collocation points

\begin{equation}
\mathcal{J}(\theta)=\frac{1}{N\,B}\sum_{i=1}^{N}\sum_{k=1}^{B}\left|\bigl(U_q^{(b)}(\mathbf{x})\,\mathbf{a}_i\bigr)_k-y_{b,k}^{(i)}\right|^{2},
\end{equation}

where $U_q^{(b)}(\mathbf{x})$ is the value of the kernel basis $U_q(\mathbf{x})$ at the boundary collocation points, i.e., the rows of the kernel matrix $G(\mathbf{x})$ corresponding to the boundary collocation points after truncation, and $\mathbf{a}_i=\mathcal{B}_\theta\bigl(\mathbf{y}_b^{(i)}\bigr)$ is the expansion coefficient output by the branch network for the $i$-th sample.

Finally, the coefficient representation $U_q(\mathbf{x})=G(\mathbf{x})\,V_q(\mathbf{x})\,\Sigma_q(\mathbf{x})^{-1}$ also reveals the functional nature of the truncated basis and provides the theoretical basis for the discrete invariance discussed above: since $U_q(\mathbf{x})\subset\mathrm{col}(G(\mathbf{x}))$, the truncated basis can be extended to a continuous basis through the combination coefficients $C(\mathbf{x})=V_q(\mathbf{x})\,\Sigma_q(\mathbf{x})^{-1}\in\mathbb{C}^{B\times q}$---the $k$-th column of the kernel basis, $\mathbf{u}_k(\mathbf{x})=\sum_{j=1}^{B}C_{jk}(\mathbf{x})\,\psi_j(\mathbf{x})$, is a linear combination of the basis functions defined at an arbitrary spatial point, and on the training grid $G(\mathbf{x})\,C(\mathbf{x})=U_q(\mathbf{x})$ holds exactly elementwise, so that the extension is not an approximation but an exact recovery of the functional nature of $U_q(\mathbf{x})$. Therefore the basis can be evaluated at any new evaluation point via $U_q(\mathbf{x})=G(\mathbf{x})\,C(\mathbf{x})$, and changing the resolution of the evaluation grid requires no retraining, so that the discrete invariance of the operator is preserved. In Case~4, both the near-field Pekeris kernel and the far-field normal-mode kernel are numerically ill-conditioned under the source-point arrangement, and the above preprocessing is therefore adopted: for the near field $q=40$ is taken (determined by the criterion $\sigma_{q+1}(\mathbf{x})/\sigma_1(\mathbf{x})\le5\times10^{-4}$); although the far-field normal-mode kernel is likewise ill-conditioned, its small singular directions carry the dominant energy of the far field ($\sigma_{50}/\sigma_1\sim10^{-16}$ while $w_{50}\sim10^{8}$), and truncating to the first $q$ singular directions gives a far-field error of $0.61$, so the retained directions are instead selected according to the input subspace (the orthonormal basis of the subspace spanned by the shell-surface vibration modes, of dimension $J=4$). See Section~3.4 for details.

\section{Green's Functions for the Shallow-Water Waveguide}
\label{app:wggreen}

This section gives the complete construction of the near-field and far-field kernel functions in Case~4.

\subsection{Near field: Pekeris waveguide Green's function}

Since the field points in the near field are not far from the spherical shell, the acoustic field is determined jointly by the superposition of the reflected waves and the direct wave; the kernel function is therefore taken to be the shallow-water waveguide Green's function $G^{n}$, which is generalized from the fundamental solution of Case~3: the standard Helmholtz fundamental solution is replaced by the waveguide Green's function obtained from the infinite image superposition over the sea surface and the seafloor, so that the reflection effects are explicitly incorporated into the kernel function.
    Here the sea surface is a pressure-release boundary with reflection coefficient $a_2=-1$;
    the seafloor is a penetrable boundary with reflection coefficient $a_1=0.4626$. This reflection coefficient is determined by the incidence angle $\theta$, and is computed as
\begin{equation}
a_1=\begin{cases}
\dfrac{a\cos\theta-\sqrt{b^{2}-\sin^{2}\theta}}{a\cos\theta+\sqrt{b^{2}-\sin^{2}\theta}}, & |\sin\theta|<b,\\[6pt]
1, & |\sin\theta|\ge b,
\end{cases}
\end{equation}
where $a=\rho_2/\rho_1$, $b=c_1/c_2$, and $\theta$ is the incidence angle. Substituting the parameters of this case gives $a=2.537$ and $b=0.932$, with $a_1=0.4626$ at $\theta=0$; this coefficient varies very little as long as the incidence angle is below $40^\circ$, so it can be approximated as the constant $a_1=0.4626$ in the near field. This also delineates the range of validity of the simplified Pekeris waveguide Green's function $G^{n}$: incidence angles below $40^\circ$, i.e., the horizontal distance between the field point and the source point must not be too large, so that $G^{n}$ applies only to the near field.
    In this way, $G^{n}$ inherently satisfies the Helmholtz equation and the reflecting boundaries of the shallow-water waveguide. The source point is mapped about the $z$ axis into $N_\eta$ circumferential virtual nodes and the kernel function is summed along the circumferential direction; its explicit expression is
\begin{equation}
G^{n}(x,y,z;x_0,y_0,z_0)=\sum_{\varepsilon=0}^{N_\eta-1}\sum_{\lambda=0}^{\infty}(a_1 a_2)^{\lambda}\left(\frac{e^{-ikR_1^{(\varepsilon)}}}{R_1^{(\varepsilon)}}+a_1\frac{e^{-ikR_2^{(\varepsilon)}}}{R_2^{(\varepsilon)}}+a_2\frac{e^{-ikR_3^{(\varepsilon)}}}{R_3^{(\varepsilon)}}+a_1 a_2\frac{e^{-ikR_4^{(\varepsilon)}}}{R_4^{(\varepsilon)}}\right),
\label{eq:pikf_gn}
\end{equation}
where the four propagation paths and the circumferential virtual nodes are
\begin{equation}
\left\{\begin{aligned}
R_1^{(\varepsilon)}&=\sqrt{(x-x_\varepsilon)^{2}+(y-y_\varepsilon)^{2}+(2\lambda H+z-z_\varepsilon)^{2}},\\
R_2^{(\varepsilon)}&=\sqrt{(x-x_\varepsilon)^{2}+(y-y_\varepsilon)^{2}+(2\lambda H+2(H-h)+z+z_\varepsilon)^{2}},\\
R_3^{(\varepsilon)}&=\sqrt{(x-x_\varepsilon)^{2}+(y-y_\varepsilon)^{2}+(2\lambda H+2h-z-z_\varepsilon)^{2}},\\
R_4^{(\varepsilon)}&=\sqrt{(x-x_\varepsilon)^{2}+(y-y_\varepsilon)^{2}+(2(\lambda+1)H-z+z_\varepsilon)^{2}},\\
x_\varepsilon&=\sqrt{x_0^{2}+y_0^{2}}\cos\eta_\varepsilon,\quad y_\varepsilon=\sqrt{x_0^{2}+y_0^{2}}\sin\eta_\varepsilon,\quad z_\varepsilon=z_0,\quad \eta_\varepsilon=\frac{2\pi\varepsilon}{N_\eta},
\end{aligned}\right.
\label{eq:pikf_R}
\end{equation}
where $(x,y,z)$ and $(x_0,y_0,z_0)$ are the coordinates of the field point and the source point, respectively, $(x_\varepsilon,y_\varepsilon,z_\varepsilon)$ is the circumferential virtual node obtained by rotating the source point about the $z$ axis through the azimuthal angle $2\pi\varepsilon/N_\eta$, $h$ and $H$ are the immersion depth of the spherical shell and the sea depth, respectively, $k$ is the near-field wavenumber, and $N_\eta$ is the number of circumferential nodes; $R_1$ is the direct wave, $R_2$ the single seafloor image, $R_3$ the single sea-surface image, and $R_4$ the double seafloor--sea-surface image, whose coefficients are $1$, $a_1$, $a_2$, and $a_1 a_2$, respectively.
As $\lambda$ increases, the images move farther and farther from the field point and their magnitudes decay according to $(a_1 a_2)^{\lambda}$, and $\lambda\to\infty$ corresponds to the exact infinite-image solution; since $|a_1 a_2|=0.4626<1$, the series converges rapidly in $\lambda$, and taking the highest order of the image superposition to be $M=200$ already makes the truncation error negligible. KernelOnet-PIKF uses $G^{n}$ as its kernel, so the expansion automatically satisfies the governing equation and the waveguide boundaries, and training can therefore be completed unsupervised using only the boundary residuals on the shell-surface collocation points.

\subsection{Far field: normal-mode Green's function}

In the far field the horizontal distance becomes larger, the incidence angle at the seafloor accordingly exceeds $40^\circ$, and the above approximation of treating $a_1$ as a constant no longer holds, so the normal-mode Green's function is used instead \cite{fu2020hybridfemsbm}
\begin{equation}
G^{f}(x,y,z;x_0,y_0,z_0)=\frac{i\pi}{\rho_1}\sum_{d=1}^{N_m}\phi_d(z_0)\,\phi_d(z)\,H_0^{(1)}\!\Big(\mu_d\sqrt{(x-x_0)^{2}+(y-y_0)^{2}}\Big),
\label{eq:pikf_gf}
\end{equation}
where $H_0^{(1)}$ is the zeroth-order Hankel function of the first kind and $N_m$ is the number of propagating modes; under the strictly penetrable (Robin) seafloor condition, all three sound speed profiles of this case have $N_m=3$ propagating modes (because $c_2>c_1$, trapped modes exist only in the narrow band $k_2<\mu<k_1$), and $\mu_d$ and $\phi_d$ are the eigenvalue and the normalized mode of the $d$-th normal mode, respectively. The modes $\phi_d$ and the eigenvalues $\mu_d$ are obtained by solving the Sturm-Liouville eigenvalue problem determined by the shallow-water sound speed profile $c_1(z)$:
\begin{equation}
\frac{d^{2}\phi_d(z)}{dz^{2}}+\left[\frac{\omega^{2}}{c_1^{2}(z)}-\mu_d^{2}\right]\phi_d(z)=0,\qquad -H<z<0,
\end{equation}
and satisfy the pressure-release boundary condition at the sea surface and the penetrable boundary condition at the seafloor
\begin{equation}
\phi_d(0)=0,\qquad \phi_d(-H)+\frac{\rho_2}{\rho_1}\frac{1}{\sqrt{\mu_d^{2}-(\omega/c_2)^{2}}}\,\frac{d\phi_d(-H)}{dz}=0.
\end{equation}
where $\kappa=\sqrt{\mu_d^{2}-(\omega/c_2)^{2}}$ is the vertical decay wavenumber of the normal mode in the sediment layer; this condition is derived from the continuity of pressure and of normal displacement at the seafloor interface, and $\sqrt{\mu_d^{2}-(\omega/c_2)^{2}}$ should appear in the denominator to keep the dimensions consistent.

\renewcommand{\thetable}{D\arabic{table}}
\setcounter{table}{0}
\renewcommand{\thetable}{D\arabic{table}}

\section{Supplementary Results for the Numerical Examples}
\label{app:supp}

This appendix reports supplementary results that are not expanded in the main text: the accuracy levels and H1 semi-norm errors of the various methods, the computational cost of training and inference, the break-even comparison with classical per-instance solvers, the batch-size ablation, the resolution invariance with respect to the evaluation grid, and the transmission loss of the reference solution. All times are measured on the machine used in this paper, and the training and inference times are synchronized by \texttt{torch.cuda.synchronize()} to reflect the true GPU wall-clock time; except in the batch-size ablation subsection, all methods are trained with the full batch, i.e., every epoch traverses all $2000$ training samples.

\subsection{Relative H1 semi-norm error and comparison at equal accuracy}
The relative $H^1$ semi-norm error measures the gradient difference between the predicted and the reference solution, and is computed at the sample level:
\begin{equation}
\mathcal{E}_{H^1}=\frac{1}{N}\sum_{i=1}^{N}\frac{\bigl\lVert\nabla\bigl(u_{\mathrm{pred}}^{(i)}-u_{\mathrm{true}}^{(i)}\bigr)\bigr\rVert}{\lVert\nabla u_{\mathrm{true}}^{(i)}\rVert},
\end{equation}
where $\lVert\cdot\rVert$ is the discrete $L_2$ norm on the evaluation grid: the gradient is approximated by second-order central differences, the star-shaped-domain grid of Case~2 uses body-fitted curvilinear coordinates and the gradient is transformed to physical space through the metric tensor of that coordinate system; the discrete summation is weighted by the area element of each evaluation grid, namely $r\,{\rm d}r\,{\rm d}\theta$ for the polar grids of Cases~1 and~3, the analytic area element $\rho R(\theta)^{2}\,{\rm d}\rho\,{\rm d}\theta$ for Case~2, and a constant area element for the uniform meridional-plane grid of Case~4, which is equivalent to equal-weight summation. For the complex-valued fields of Cases~3 and~4, the gradient modulus is given by the sum of the squares of the real and imaginary parts, consistent with the treatment of the relative $L_2$ error.

The equal-accuracy comparison in Table~\ref{tab:acc} shows that the kernel-based methods converge markedly faster than DeepONet: on Cases~1 and~2 they reach $10^{-2}$ in less than one third of the epochs required by the latter, neither DeepONet nor PI-DeepONet ever drops to $10^{-3}$ within the whole budget, whereas KernelOnet-PIKF reaches that accuracy on Case~1 at $3.8\times10^{5}$ epochs. The H1 semi-norm errors give exactly the same ordering as the $L_2$ errors, which shows that the high accuracy of the kernel-based methods is not limited to the $L_2$ norm. The ratio of the two also varies with the type of problem: the solutions of Cases~1 and~2 are smooth and the H1 error is about $6\sim24$ times the $L_2$ error, indicating that the gradient is harder to fit than the function values themselves; the acoustic fields of Cases~3 and~4 are oscillatory, and the ratio of the gradient to the function value is governed by the wavenumber content of the field itself, so the two are close, namely $0.93\sim1.04$ for Case~3 and $0.999\sim1.001$ for Case~4. The near equality in Case~4 arises because its far-field reference solution and the network kernel are both normal-mode expansions and the error likewise lies in the space spanned by the same set of propagating modes, so that the gradient norm in the numerator and the denominator are scaled by the same wavenumber factor and cancel. Moreover, several configurations in the table reach $10^{-2}$ already at the $10^4$-epoch level, with the corresponding training wall-clock times given in Table~\ref{tab:cost}, only seconds to tens of seconds, which shows that the $5\times10^{5}$-epoch budget mainly serves to give the comparison methods ample opportunity to converge rather than being necessary for the method of this paper.
\begin{table}[htbp]
\centering\footnotesize\rmfamily
\caption{Relative $L_2$ and relative $H^1$ semi-norm errors of each case and each method, together with the numbers of training epochs required for the validation error to first fall to given accuracy levels, all computed over $2000$ test samples; ``$\text{--}$'' means that the level was not reached within the $5\times10^{5}$-epoch budget.Case~2 takes $\varepsilon=4$ and also lists the $K_c=0$ baseline with the correction branch disabled; Case~3 takes the main case $k=20$; for Case~4 the three sound speed profiles are listed for the far field.}
\label{tab:acc}
\resizebox{\textwidth}{!}{%
\begin{tabular}{llcccccc}
\toprule
Case & Method & rel. $L_2$ & rel. H1 & to $5\times10^{-2}$ & to $10^{-2}$ & to $5\times10^{-3}$ & to $10^{-3}$ \\
\midrule
1 & DeepONet & $1.89\times10^{-3}$ & $3.47\times10^{-2}$ & 9,500 & 105,000 & 182,500 & -- \\
1 & PI-DeepONet & $3.53\times10^{-3}$ & $4.17\times10^{-2}$ & 21,500 & 225,000 & 389,000 & -- \\
1 & KernelOnet-RBF & $1.29\times10^{-3}$ & $3.10\times10^{-2}$ & 3,000 & 14,500 & 40,000 & -- \\
1 & KernelOnet-PIKF & $8.89\times10^{-4}$ & $6.30\times10^{-3}$ & 1,500 & 10,500 & 20,000 & 379,500 \\
1 & KernelOnet-HK & $2.04\times10^{-3}$ & $2.19\times10^{-2}$ & 3,000 & 29,000 & 93,000 & -- \\
\midrule
2 ($\varepsilon=4$) & DeepONet & $5.71\times10^{-3}$ & $7.06\times10^{-2}$ & 19,500 & 275,000 & -- & -- \\
2 ($\varepsilon=4$) & KernelOnet-RBF & $5.11\times10^{-3}$ & $4.76\times10^{-2}$ & 7,500 & 44,000 & -- & -- \\
2 ($\varepsilon=4$) & KernelOnet-HK & $2.75\times10^{-3}$ & $2.75\times10^{-2}$ & 6,000 & 56,000 & 159,500 & -- \\
2 ($\varepsilon=4$) & KernelOnet-HK ($K_c=0$ baseline) & $2.08\times10^{-2}$ & $1.28\times10^{-1}$ & 8,000 & -- & -- & -- \\
\midrule
3 ($k=20$) & KernelOnet-PIKF (collocation) & $1.90\times10^{-3}$ & $1.77\times10^{-3}$ & 2,000 & 14,500 & 45,000 & -- \\
3 ($k=20$) & KernelOnet-PIKF-SVD (collocation $+$ SVD) & $1.36\times10^{-3}$ & $1.37\times10^{-3}$ & 2,500 & 10,500 & 18,500 & -- \\
3 ($k=20$) & KernelOnet-PIKF-SL (boundary integral) & $5.15\times10^{-3}$ & $5.20\times10^{-3}$ & 19,000 & 106,000 & -- & -- \\
3 ($k=20$) & KernelOnet-PIKF-SL-SVD (boundary integral $+$ SVD) & $1.85\times10^{-3}$ & $1.92\times10^{-3}$ & 2,000 & 9,500 & 17,500 & -- \\
\midrule
4 (near field) & KernelOnet-PIKF & $8.57\times10^{-4}$ & $8.56\times10^{-4}$ & 3,500 & 13,000 & 21,000 & 284,000 \\
4 (far field) & KernelOnet-PIKF ($c_1{=}1507{-}0.24z$) & $1.18\times10^{-3}$ & $1.18\times10^{-3}$ & 3,500 & 6,500 & 12,500 & -- \\
4 (far field) & KernelOnet-PIKF ($c_1{=}1510$) & $1.31\times10^{-3}$ & $1.31\times10^{-3}$ & 3,500 & 6,500 & 12,500 & -- \\
4 (far field) & KernelOnet-PIKF ($c_1{=}1513{+}0.24z$) & $1.18\times10^{-3}$ & $1.18\times10^{-3}$ & 3,500 & 6,500 & 13,000 & -- \\
\bottomrule
\end{tabular}}
\end{table}

\subsection{Model size, training cost, and GPU memory usage}
Table~\ref{tab:cost} summarizes the learnable parameters, the training cost, and the GPU memory usage of the various methods. As for the learnable parameters, the two DeepONets consist of two fully connected networks, a branch network and a trunk network, and have the largest parameter counts, about $1.8\times10^{5}$; the kernel of KernelOnet-PIKF has only $1$ scalar parameter, giving the smallest parameter count, $1.03\times10^{5}$ in Case~1; KernelOnet-RBF and KernelOnet-HK additionally contain a shallow kernel network and a low-rank correction network, respectively, and their parameter counts lie in between, about $1.3\sim1.5\times10^{5}$. As for the training speed, the physics-informed kernel, which needs to be evaluated only at the boundary collocation points, costs the least per epoch, namely $0.061$~s/100 epochs for KernelOnet-PIKF in Case~1, slightly below the $0.076$~s/100 epochs of DeepONet; KernelOnet-RBF needs to be evaluated at all $160$ boundary source points and reaches $0.431$~s/100 epochs, about seven times the former; PI-DeepONet must evaluate the PDE residual at the interior collocation points and back-propagate it in every epoch, costing $0.534$~s/100 epochs, comparable to RBF. The peak training GPU memory is consistent with the scale of the trunk evaluation: in Case~1 KernelOnet-PIKF uses only $48$~MiB, whereas KernelOnet-RBF and PI-DeepONet reach $689$ and $687$~MiB, respectively; apart from the $332$~MiB of KernelOnet-RBF, the peak inference memory does not exceed $71$~MiB, the latter again arising from the need to evaluate all source points. The differences in Case~3 come from the size of the kernel basis: the untruncated collocation form costs the most per epoch, $0.105$~s/100 epochs, while its boundary integral form and the SVD-truncated variants drop to $0.063\sim0.069$~s/100 epochs. The complete training wall-clock time and per-sample inference time of each configuration are given in Table~\ref{tab:app_be}.
\begin{table}[htbp]
\centering\footnotesize\rmfamily
\caption{Learnable parameters, training cost, and GPU memory usage of each case and each method. The per-sample training time is obtained by dividing the time per $100$ epochs by $100\times2000$. Case~2 takes the nonlinear main case $\varepsilon=4$ and also lists the $K_c=0$ baseline with the correction branch disabled; Case~3 takes the main case $k=20$; both the near field and the far field of Case~4 are trained unsupervised, and the three sound speed profiles are listed for the far field.}
\label{tab:cost}
\resizebox{\textwidth}{!}{%
\begin{tabular}{llccccc}
\toprule
Case & Method & Learnable parameters & s/100 ep. & Training per sample ($\mu$s) & Memory, training (MiB) & Memory, inference (MiB) \\
\midrule
1 & DeepONet & 180,801 & 0.076 & 0.38 & 128 & 21 \\
1 & PI-DeepONet & 180,801 & 0.534 & 2.67 & 687 & 21 \\
1 & KernelOnet-RBF & 129,281 & 0.431 & 2.15 & 689 & 332 \\
1 & KernelOnet-PIKF & 103,041 & 0.061 & 0.30 & 48 & 20 \\
1 & KernelOnet-HK & 132,503 & 0.102 & 0.51 & 169 & 59 \\
\midrule
2 ($\varepsilon=4$) & DeepONet & 187,201 & 0.074 & 0.37 & 103 & 20 \\
2 ($\varepsilon=4$) & KernelOnet-RBF & 142,121 & 0.407 & 2.04 & 641 & 312 \\
2 ($\varepsilon=4$) & KernelOnet-HK & 147,275 & 0.119 & 0.59 & 184 & 71 \\
2 ($\varepsilon=4$) & KernelOnet-HK ($K_c=0$ baseline) & 142,123 & 0.098 & 0.49 & 112 & 34 \\
\midrule
3 ($k=20$) & KernelOnet-PIKF (collocation) & 128,801 & 0.105 & 0.52 & 104 & 56 \\
3 ($k=20$) & KernelOnet-PIKF-SVD (collocation $+$ SVD) & 90,160 & 0.065 & 0.33 & 97 & 43 \\
3 ($k=20$) & KernelOnet-PIKF-SL (boundary integral) & 128,800 & 0.069 & 0.34 & 158 & 54 \\
3 ($k=20$) & KernelOnet-PIKF-SL-SVD (boundary integral $+$ SVD) & 112,700 & 0.063 & 0.32 & 99 & 44 \\
\midrule
4 (near field) & KernelOnet-PIKF & 80,560 & 0.063 & 0.31 & 62 & 32 \\
4 (far field) & KernelOnet-PIKF ($c_1{=}1507{-}0.24z$) & 68,968 & 0.062 & 0.31 & 37 & 24 \\
4 (far field) & KernelOnet-PIKF ($c_1{=}1510$) & 68,968 & 0.062 & 0.31 & 37 & 24 \\
4 (far field) & KernelOnet-PIKF ($c_1{=}1513{+}0.24z$) & 68,968 & 0.062 & 0.31 & 37 & 24 \\
\bottomrule
\end{tabular}}
\end{table}

\subsection{Reference-solution generation cost and break-even comparison}
This subsection compares the cost of a classical per-instance solver with the inference cost of the neural operator on a common basis. The per-instance solve is timed in two ways: \emph{pre-factorized} means that the instance-independent matrix and its factorization are constructed only once, so that each instance needs only a coefficient solve and an evaluation, which is feasible for linear problems with fixed geometry; \emph{from scratch} means that the matrix is reconstructed and factorized for every instance, in which case the geometry, the coefficients, and the source term may all vary from instance to instance.

Table~\ref{tab:app_be} shows that the benefit of break-even depends on how expensive the classical solver is. Case~2 is a nonlinear problem in which every instance must be reassembled and solved through Newton iterations, costing about $1.0$~s per instance, so that the operator needs only a few hundred to two thousand queries to amortize the training cost, namely $3.6\times10^{2}$ queries for DeepONet and $5.8\times10^{2}$ for KernelOnet-HK. The near field of Case~4 requires the solution of a ring-source boundary integral for the source strengths, and if the matrix is allowed to be reconstructed for every instance, the break-even point is about $6.3\times10^{2}$ queries. Conversely, Cases~1 and~3 and the far field of Case~4 are linear problems with fixed geometry, for which the pre-factorized method of fundamental solutions costs only $0.005\sim0.3$~ms per instance, below the operator inference cost, so that no break-even point exists in the table; here the value of the operator lies not in replacing a single solve but in obtaining the solution for arbitrary new boundary data in a single forward pass. The stratification of the inference times is precisely what causes the differences in the break-even points: the configurations trained directly on the boundary residual all cost $0.063\sim0.252$~ms/sample, whereas KernelOnet-RBF must evaluate all source points and reaches $1.4\sim1.5$~ms/sample, and hence has the highest break-even point on Case~2. The reference-solution generation cost differs enormously among the cases: the finite-element solution of Case~2 requires about $41$~minutes per dataset, whereas the method of fundamental solutions of Cases~1 and~3 requires only $0.15$ and $0.34$~s, respectively; since the unsupervised configurations need only boundary data during training, this cost does not enter their training cost but is used for evaluation and for the supervised baselines. As an order-of-magnitude reference for GPU-native solvers, the GPU-based Galerkin finite-element solver TensorMesh~\cite{wen2026tensorgalerkin} takes about $10\sim77$~s to solve a Helmholtz problem with on the order of one million nodes on an A100, and the cost grows with the wavenumber; such solvers are far faster than the CPU implementation of this paper for a single large-scale solve, but the system must still be reassembled and re-solved whenever the boundary conditions or the source term change.

\begin{table}[htbp]
\centering\footnotesize\rmfamily
\caption{Reference-solution generation cost, the inference cost of the classical per-instance solver and of the neural operator, and the break-even number of queries $N^{\star}$ for each case. All times are measured on the machine used in this paper, averaged over $100$ instances. Solvers: the method of fundamental solutions (MFS) for Cases~1 and~3, finite elements $+$ Newton for Case~2, a ring-source boundary integral for the near field and normal modes for the far field of Case~4. The per-instance solve and $N^{\star}$ are both given in the pre-factorized and the from-scratch settings; Case~2 is a nonlinear problem whose Jacobian changes with the solution and which has no pre-factorized setting, so only a single value is listed. ``$\text{--}$'' means that the single-instance cost of the classical solver is below the operator inference cost, in which case no break-even point exists;$^{\dagger}$ denotes unsupervised training that uses no interior solution labels, whose training does not depend on the reference-solution generation cost.}
\label{tab:app_be}
\resizebox{\textwidth}{!}{%
\begin{tabular}{llcccccc}
\toprule
Case & Method & Ref. soln. generation (s/2000) & Per-instance solve (ms) & Inference (ms/sample) & Training wall clock (h) & $N^{\star}$ \\
\midrule
1 & DeepONet & $0.15$ & $0.032/1.02$ & $0.086$ & $0.106$ & $--/4.1\times10^{5}$ \\
1 & PI-DeepONet$^{\dagger}$ & $0.15$ & $0.032/1.02$ & $0.085$ & $0.741$ & $--/2.9\times10^{6}$ \\
1 & KernelOnet-RBF & $0.15$ & $0.032/1.02$ & $1.515$ & $0.599$ & $--/--$ \\
1 & KernelOnet-PIKF$^{\dagger}$ & $0.15$ & $0.032/1.02$ & $0.067$ & $0.084$ & $--/3.2\times10^{5}$ \\
1 & KernelOnet-HK & $0.15$ & $0.032/1.02$ & $0.183$ & $0.142$ & $--/6.1\times10^{5}$ \\
\midrule
2 ($\varepsilon=4$) & DeepONet & $2445$ & $1022$ & $0.083$ & $0.103$ & $3.6\times10^{2}$ \\
2 ($\varepsilon=4$) & KernelOnet-RBF & $2445$ & $1022$ & $1.439$ & $0.566$ & $2.0\times10^{3}$ \\
2 ($\varepsilon=4$) & KernelOnet-HK & $2445$ & $1022$ & $0.252$ & $0.165$ & $5.8\times10^{2}$ \\
2 ($\varepsilon=4$) & KernelOnet-HK ($K_c=0$) & $2445$ & $1022$ & $0.166$ & $0.135$ & $4.8\times10^{2}$ \\
\midrule
3 ($k=20$) & KernelOnet-PIKF$^{\dagger}$ & $0.34$ & $0.029/103.98$ & $0.194$ & $0.146$ & $--/5.1\times10^{3}$ \\
3 ($k=20$) & KernelOnet-PIKF-SVD$^{\dagger}$ & $0.34$ & $0.029/103.98$ & $0.063$ & $0.091$ & $--/3.1\times10^{3}$ \\
3 ($k=20$) & KernelOnet-PIKF-SL$^{\dagger}$ & $0.34$ & $0.029/103.98$ & $0.066$ & $0.096$ & $--/3.3\times10^{3}$ \\
3 ($k=20$) & KernelOnet-PIKF-SL-SVD$^{\dagger}$ & $0.34$ & $0.029/103.98$ & $0.063$ & $0.088$ & $--/3.1\times10^{3}$ \\
\midrule
4 (near field) & KernelOnet-PIKF$^{\dagger}$ & $9.10$ & $0.295/494.70$ & $0.064$ & $0.087$ & $1.3\times10^{6}/6.3\times10^{2}$ \\
4 (far field, 1507$-$0.24$z$) & KernelOnet-PIKF$^{\dagger}$ & $1.30$ & $0.005/1298$ & $0.064$ & $0.087$ & $--/2.4\times10^{2}$ \\
4 (far field, 1510) & KernelOnet-PIKF$^{\dagger}$ & $1.37$ & $0.005/1298$ & $0.065$ & $0.085$ & $--/2.4\times10^{2}$ \\
4 (far field, 1513$+$0.24$z$) & KernelOnet-PIKF$^{\dagger}$ & $1.32$ & $0.005/1298$ & $0.064$ & $0.087$ & $--/2.4\times10^{2}$ \\
\bottomrule
\end{tabular}}
\end{table}

\subsection{Batch-size ablation}
Table~\ref{tab:app_bs} examines the influence of the batch size on the three data-driven models in Case~2. Four batch sizes are used, $500$, $1000$, $1500$, and the full batch, and all training settings other than the batch size are identical to those of this case in Table~\ref{tab:acc}; when the batch size is smaller than the training-set size, every epoch traverses all training samples batch by batch, and the time per epoch increases as the batch size decreases. All three models benefit from mini-batches, but to different degrees: the relative $L_2$ error of DeepONet drops from $5.71\times10^{-3}$ with the full batch to $2.15\times10^{-3}$ at a batch size of $500$, an improvement of a factor of $2.66$, making it the most sensitive to the batch size; KernelOnet-RBF drops from $5.11\times10^{-3}$ to $3.03\times10^{-3}$, an improvement of $1.69$ times; and KernelOnet-HK drops from $2.75\times10^{-3}$ to $1.75\times10^{-3}$, an improvement of $1.57$ times, with the ordering of the H1 semi-norm errors exactly the same. Consequently, the relative merits of DeepONet and KernelOnet-RBF change with the batch size: with the full batch RBF is slightly better, while with mini-batches DeepONet overtakes it; KernelOnet-HK, by contrast, remains the best at all four batch sizes. The gain in accuracy is paid for in training time: the time per $100$ epochs increases by a factor of about $3.5\sim3.9$, and the total wall-clock time accordingly grows from $0.10\sim0.57$~h to $0.36\sim2.20$~h; the peak GPU memory, in contrast, decreases slightly as the batch size decreases, from $103$ to $61$~MiB for DeepONet, whereas the two kernel models change only marginally, their memory being dominated by the evaluation of the kernel function at all source points. It can be seen that the full batch is not the training setting with the best accuracy, but since the main text adopts the full batch uniformly for the three models in every case, this does not affect the comparison among the methods; if accuracy is the sole objective, a batch size of $500$ can further reduce the error at about four times the training wall-clock time.
\begin{table}[htbp]
\centering\footnotesize\rmfamily
\caption{Batch-size ablation for the three data-driven models on Case~2 ($\varepsilon=4$): relative $L_2$/H1 semi-norm errors, training time per $100$ epochs, total training wall-clock time, and peak training GPU memory. The full batch means that the batch size equals the training-set size, and its values agree with the corresponding rows of this case in Table~\ref{tab:cost}.}
\label{tab:app_bs}
\resizebox{\textwidth}{!}{%
\begin{tabular}{llccccc}
\toprule
Method & Batch size & rel. $L_2$ & rel. H1 & s/100 ep. & Total wall clock (h) & Peak memory (MiB) \\
\midrule
DeepONet & 500 & $2.15\times10^{-3}$ & $2.18\times10^{-2}$ & $0.260$ & $0.362$ & $61$ \\
DeepONet & 1000 & $2.82\times10^{-3}$ & $3.47\times10^{-2}$ & $0.136$ & $0.188$ & $73$ \\
DeepONet & 1500 & $3.03\times10^{-3}$ & $3.63\times10^{-2}$ & $0.137$ & $0.190$ & $89$ \\
DeepONet & Full batch (2000) & $5.71\times10^{-3}$ & $7.06\times10^{-2}$ & $0.074$ & $0.103$ & $103$ \\
\midrule
KernelOnet-RBF & 500 & $3.03\times10^{-3}$ & $2.98\times10^{-2}$ & $1.587$ & $2.204$ & $630$ \\
KernelOnet-RBF & 1000 & $3.81\times10^{-3}$ & $3.87\times10^{-2}$ & $0.812$ & $1.127$ & $633$ \\
KernelOnet-RBF & 1500 & $3.84\times10^{-3}$ & $3.84\times10^{-2}$ & $0.802$ & $1.114$ & $638$ \\
KernelOnet-RBF & Full batch (2000) & $5.11\times10^{-3}$ & $4.76\times10^{-2}$ & $0.407$ & $0.566$ & $641$ \\
\midrule
KernelOnet-HK & 500 & $1.75\times10^{-3}$ & $1.62\times10^{-2}$ & $0.427$ & $0.594$ & $174$ \\
KernelOnet-HK & 1000 & $2.10\times10^{-3}$ & $2.03\times10^{-2}$ & $0.223$ & $0.310$ & $178$ \\
KernelOnet-HK & 1500 & $2.23\times10^{-3}$ & $2.11\times10^{-2}$ & $0.216$ & $0.300$ & $180$ \\
KernelOnet-HK & Full batch (2000) & $2.75\times10^{-3}$ & $2.75\times10^{-2}$ & $0.119$ & $0.165$ & $184$ \\
\bottomrule
\end{tabular}}
\end{table}

\subsection{Resolution invariance}
Table~\ref{tab:app_res} shows that the relative $L_2$ errors of the three models all remain at the $10^{-3}$ level throughout the refinement, and that those of DeepONet and KernelOnet-HK vary by less than $10\%$: the solution of the proposed operator is determined by the data on the training grid, so that once training is complete it can be evaluated directly on an evaluation grid of arbitrary resolution without retraining. The error of KernelOnet-RBF rises to $4.3\times10^{-3}$ when the evaluation grid is refined to twice the training grid or more, and stays within $1.5\times10^{-3}$ on all the other grids.
\begin{table}[htbp]
\centering\footnotesize\rmfamily
\caption{Resolution invariance of the three data-driven models with respect to the evaluation grid on Case~1: relative $L_2$ error as a function of the evaluation grid $n_r\times n_\theta$. All models are trained on the native $40\times40$ grid and then evaluated directly on each grid with frozen weights; the reference solution is sampled from the same fundamental solution on each grid, and the average over $2000$ test samples is reported. The number of evaluation-grid points increases from $100$ to $25{,}600$, a span of $256$ times.}
\label{tab:app_res}
\resizebox{\textwidth}{!}{%
\begin{tabular}{lccccc}
\toprule
Evaluation grid $n_r\times n_\theta$ & $10\times10$ & $20\times20$ & $40\times40$ (training) & $80\times80$ & $160\times160$ \\
\midrule
DeepONet & $2.35\times10^{-3}$ & $2.07\times10^{-3}$ & $1.89\times10^{-3}$ & $1.80\times10^{-3}$ & $1.76\times10^{-3}$ \\
KernelOnet-RBF & $1.50\times10^{-3}$ & $1.35\times10^{-3}$ & $1.29\times10^{-3}$ & $4.36\times10^{-3}$ & $4.31\times10^{-3}$ \\
KernelOnet-HK & $2.50\times10^{-3}$ & $2.19\times10^{-3}$ & $2.04\times10^{-3}$ & $1.96\times10^{-3}$ & $1.92\times10^{-3}$ \\
\bottomrule
\end{tabular}}
\end{table}


\subsection{Transmission loss of the reference solution in Case~4}
Table~\ref{tab:app_tl} gives the transmission loss of the reference acoustic field. The reference solutions are synthesized waveguide Green's functions in which the modal excitation amplitudes are random for each sample, so that the absolute sound level has no physical meaning, and the transmission loss in the table therefore represents the relative sound-level change within the window. The near field decays rapidly outward from $2$~m, dropping by $23.0$~dB at $26.8$~m, in agreement with the $20\log_{10}(r/r_0)\approx22.5$~dB of spherical spreading, which indicates that the acoustic field at this distance is still dominated by spherical spreading; within the far-field window the fluctuation is only about $2.5$~dB, whereas cylindrical spreading yields only about $0.2$~dB of loss over the same $50$~m distance, so that the far-field structure is dominated by modal interference. The far-field reference point happens to lie at a relative minimum of the modal interference, $9\sim13$~dB below the peak of the window, so that the transmission loss at most points in the window is negative, i.e., stronger than at the reference point.
\begin{table}[htbp]
\centering\footnotesize\rmfamily
\caption{Transmission loss of the reference solution of Case~4 in each evaluation window, over $2000$ test samples. $\mathrm{TL}=-20\log_{10}\bigl(|p(r,z)|/|p(r_0,z_0)|\bigr)$, where the reference point $(r_0,z_0)$ is taken at the source depth at the nearest distance within the window; the TL at the source depth is the sample mean, and the TL within the window is listed as the minimum, median, and maximum over the pooled set of samples and window points.}
\label{tab:app_tl}
\resizebox{\textwidth}{!}{%
\begin{tabular}{llccc}
\toprule
Window & Reference point $(r_0,z_0)$ (m) & TL at source depth (dB) & TL in window (dB) & Reference point relative to peak (dB) \\
\midrule
Near field ($x\in[2,50]$~m) & $(2.0,\,14.7)$ & $2$~m: $0$; $26.8$~m: $+23.0$; $50$~m: $+21.2$ & $-18.0$ / $+20.8$ / $+81.7$ & $-1.1$ \\
Far field ($c_1(z)=1507-0.24z$) & $(950.0,\,14.6)$ & $950$~m: $0$; $975$~m: $-1.8$; $1000$~m: $-2.5$ & $-53.0$ / $-11.3$ / $+43.7$ & $-13.0$ \\
Far field ($c_1(z)=1510$) & $(950.0,\,14.6)$ & $950$~m: $0$; $975$~m: $-1.6$; $1000$~m: $-1.6$ & $-37.4$ / $-9.6$ / $+51.1$ & $-12.7$ \\
Far field ($c_1(z)=1513+0.24z$) & $(950.0,\,14.6)$ & $950$~m: $0$; $975$~m: $-0.1$; $1000$~m: $+0.5$ & $-39.7$ / $-7.4$ / $+52.0$ & $-8.8$ \\
\bottomrule
\end{tabular}}
\end{table}

\pdfbookmark[1]{References}{bmark-references}

\bibliographystyle{unsrtnat}
\bibliography{references}

\end{document}